\documentclass[12pt]{article}
\usepackage{times}
\usepackage[hungarian,british]{babel}
\usepackage{url}
\usepackage{latexsym}
\usepackage[utf8]{inputenc}
\usepackage[hidelinks]{hyperref}
\usepackage{color}
\usepackage[dvipsnames]{xcolor}
\definecolor{OliveGreen}{rgb}{0,0.6,0}
\usepackage{graphicx}
\usepackage{booktabs,amsmath,multicol,bm}
\usepackage{breakcites}
\usepackage{float}
\usepackage[nottoc]{tocbibind}
\usepackage[top=2.5cm, bottom=4cm, left=2.5cm, right=2.5cm]{geometry}
\usepackage{array}
\newcolumntype{L}[1]{>{\raggedright\let\newline\\\arraybackslash\hspace{0pt}}m{#1}}

\DeclareMathOperator{\Attention}{Attention}
\DeclareMathOperator{\softmax}{softmax}
\DeclareMathOperator{\MultiHead}{MultiHead}
\DeclareMathOperator{\Concatenate}{Concatenate}
\DeclareMathOperator{\head}{head}

\DeclareMathOperator{\FFN}{FFN}
\DeclareMathOperator{\MLP}{MLP}
\DeclareMathOperator{\DP}{DP}
\DeclareMathOperator{\SDP}{SDP}
\DeclareMathOperator{\BL}{BL}
\DeclareMathOperator{\Loss}{Loss}

\DeclareMathOperator{\Layer}{Layer}
\DeclareMathOperator{\LayerNormalization}{LayerNormalization}

\begin{document}
	\title{Deep Learning Based Chatbot Models}
	\author{Rich\'ard Kriszti\'an Cs\'aky \\ Budapest University of Technology and Economics}
	\date{\today}
	
\begin{titlepage}
		\centering
		\includegraphics[width=0.6\textwidth]{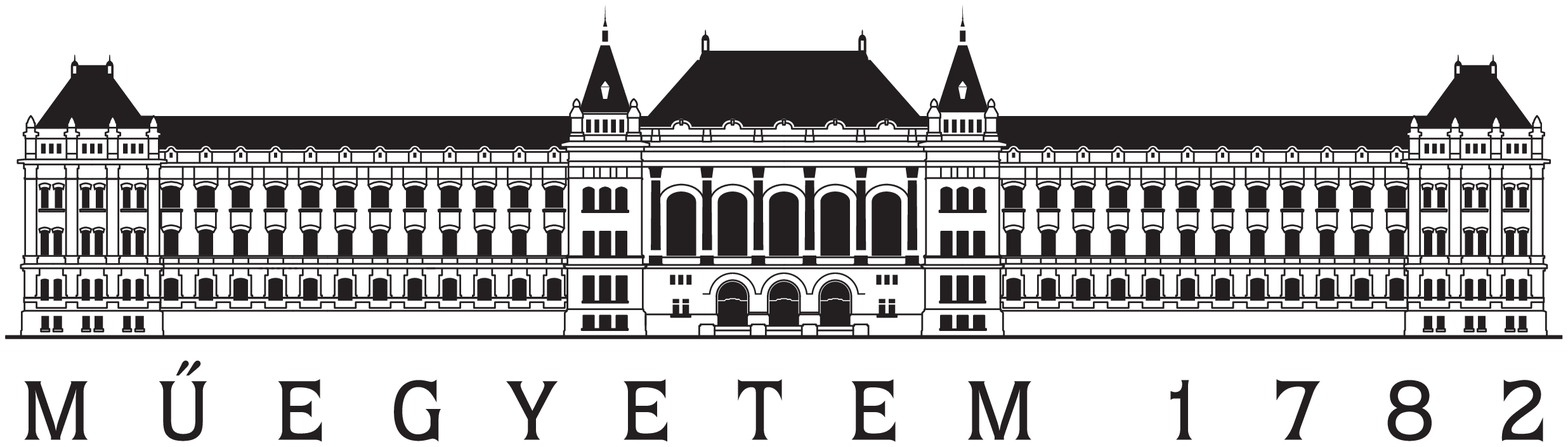}\par\vspace{1cm}
		{\scshape Budapest University of Technology and Economics \\ Faculty of Electrical Engineering and Informatics \\
			Department of Automation and Applied Informatics \par}
		\vspace{2cm}
		{\huge\mdseries Deep Learning Based Chatbot Models\par}
		\vspace{0.5cm}
		{\scshape\Large Scientific Students' Associations Report\par}
		\vspace{1.5cm}
		
		{\Large Author: \\ Richárd Krisztián Csáky\par}
		\vspace{1.5cm}
		{\Large Supervised by \\ G\'abor Recski\par}
		\vfill
		\par
		{\Large 2017\par}

\end{titlepage}

\newpage\begin{abstract}
A conversational agent (chatbot) is a piece of software that is able to communicate with humans using natural language. Modeling conversation is an important task in natural language processing and artificial intelligence (AI). Indeed, ever since the birth of AI, creating a good chatbot remains one of the field’s hardest challenges. While chatbots can be used for various tasks, in general they have to understand users’ utterances and provide responses that are relevant to the problem at hand.

In the past, methods for constructing chatbot architectures have relied on hand-written rules and templates or simple statistical methods. With the rise of deep learning these models were quickly replaced by end-to-end trainable neural networks around 2015. More specifically, the recurrent encoder-decoder model \cite{Cho:2014} dominates the task of conversational modeling. This architecture was adapted from the neural machine translation domain, where it performs extremely well. Since then a multitude of variations \cite{Serban:2015} and features were presented that augment the quality of the conversation that chatbots are capable of.

In my work, I conduct an in-depth survey of recent literature, examining over 70 publications related to chatbots published in the last 3 years. Then I proceed to make the argument that the very nature of the general conversation domain demands approaches that are different from current state-of-the-art architectures. Based on several examples from the literature I show why current chatbot models fail to take into account enough priors when generating responses and how this affects the quality of the conversation. In the case of chatbots these priors can be outside sources of information that the conversation is conditioned on like the persona \cite{Li:2016} or mood of the conversers. In addition to presenting the reasons behind this problem, I propose several ideas on how it could be remedied.

The next section of my paper focuses on adapting the very recent Tranformer \cite{Vaswani:2017} model to the chatbot domain, which is currently the state-of-the-art in neural machine translation. I first present my experiments with the vanilla model, using conversations extracted from the Cornell Movie-Dialog Corpus \cite{Danescu:2011}. Secondly, I augment the model with some of my ideas regarding the issues of encoder-decoder architectures. More specifically, I feed additional features into the model like mood or persona together with the raw conversation data. Finally, I conduct a detailed analysis of how the vanilla model performs on conversational data by comparing it to previous chatbot models and how the additional features, affect the quality of the generated responses.
\end{abstract}

\newpage\tableofcontents
\newpage\section{Introduction} \label{sec:intro}

A conversational agent (chatbot) is a piece of software that is able to communicate with humans using natural language. Ever since the birth of AI, modeling conversations remains one of the field's toughest challenges. Even though they are far from perfect, chatbots are now used in a plethora of applications like Apple's Siri \cite{Siri:2017}, Google's Google Assistant \cite{Google:2017} or Microsoft's Cortana \cite{Cortana:2017}. In order to fully understand the capabilities and limitations of current chatbot architectures and techniques an in-depth survey is conducted, where related literature published over the past 3 years is examined and a recently introduced neural network model is trained using conversational data. 

The paper begins with a brief overview of the history of chatbots in Section~\ref{sec:history}, where the properties and objectives of conversational modeling are discussed. Early approaches are presented in Section~\ref{ssec:22} as well as the current dominating model for building conversational agents, based on neural networks, in Section~\ref{ssec:23}.

In Section~\ref{sec:background} key architectures and techniques are described that were developed over the past 3 years related to chatbots. Publications are grouped into categories on the basis of specific techniques or approaches discussed by the authors. After this, criticism is presented in Section~\ref{ssec:problems} regarding some of the properties of current chatbot models and it is shown how several of the techniques used are inappropriate for the task of modeling conversations.

In the next part (Section~\ref{sec:experiments}) preliminary experiments are conducted by training a novel neural network based model, the Transformer \cite{Vaswani:2017}, using dialog datasets \cite{Danescu:2011,Tiedemann:2009,OpenSubtitles:2016}. Several trainings are run using these datasets, detailed in Section~\ref{ssec:43}. The results of the various training setups are presented in Section~\ref{sec:results} by qualitatively comparing them to previous chatbot models and by using standard evaluation metrics.

Finally, in Section~\ref{sec:future} possible directions for future work are offered. More specifically, several ideas are proposed in order to remedy the problems presented in Section~\ref{ssec:problems} and future research directions are discussed.

\newpage\section{History of Chatbots} \label{sec:history}

\subsection{Modeling Conversations} \label{ssec:21}
Chatbot models usually take as input natural language sentences uttered by a user and output a response. There are two main approaches for generating responses. The traditional approach is to use hard-coded templates and rules to create chatbots, presented in Section~\ref{ssec:22}. The more novel approach, discussed in detail in Section~\ref{ssec:23}, was made possible by the rise of deep learning. Neural network models are trained on large amounts of data to learn the process of generating relevant and grammatically correct responses to input utterances. Models have also been developed to accommodate for spoken or visual inputs. They oftentimes make use of a speech recognition component to transform speech into text \cite{Serban:2017} or convolutional neural networks \cite{Imagenet:2012} that transform the input pictures into useful representations for the chatbot \cite{Havrylov:2017}. The latter models are also called visual dialog agents, where the conversation is grounded on both textual and visual input \cite{Das:2017}.

Conversational agents exist in two main forms. The first one is the more traditional task-oriented dialog system, which is limited in its conversational capabilities, however it is very robust at executing task specific commands and requirements. Task-oriented models are built to accomplish specific tasks like making restaurant reservations \cite{Joshi:2017,Bordes:2016} or promoting movies \cite{Yu:2017}, to name a few. These systems often don't have the ability to respond to arbitrary utterances since they are limited to a specific domain, thus users have to be guided by the dialog system towards the task at hand. Usually they are deployed to tasks where some information has to be retrieved from a knowledge base. They are mainly employed to replace the process of navigating through menus and user interfaces like making the activity of booking flight tickets or finding public transportation routes between locations conversational \cite{Zhao:2017}.  

The second type of dialog agents are the non-task or open-domain chatbots. These conversation systems try to imitate human dialog in all its facets. This means that one should hardly be able to distinguish such a chatbot from a real human, but current models are still far away from such claims. They are usually trained with dialog examples extracted from movie scripts or from Twitter-like post-reply pairs \cite{Vinyals:2015,Shang:2015,Serban:2015,Li:2016}. For these models there isn't a well defined goal, but they are required to have a certain amount of world knowledge and commonsense reasoning capabilities in order to hold conversations about any topic.

Recently an emphasis has been put on integrating the two types of conversational agents. The main idea is to combine the positive aspects of both types, like the robust abilities of goal-oriented dialog systems to perform tasks and the human-like chattyness of open-domain chatbots \cite{Zhao:2017,Yu:2017,Serban:2017}. This is beneficial because the user is more likely to engage with a task-oriented dialog agent if it's more natural, and handles out of domain responses well.

\subsection{Early Approaches} \label{ssec:22}

\begin{figure}[H]
	\centering
	\includegraphics[width=0.4\textwidth]{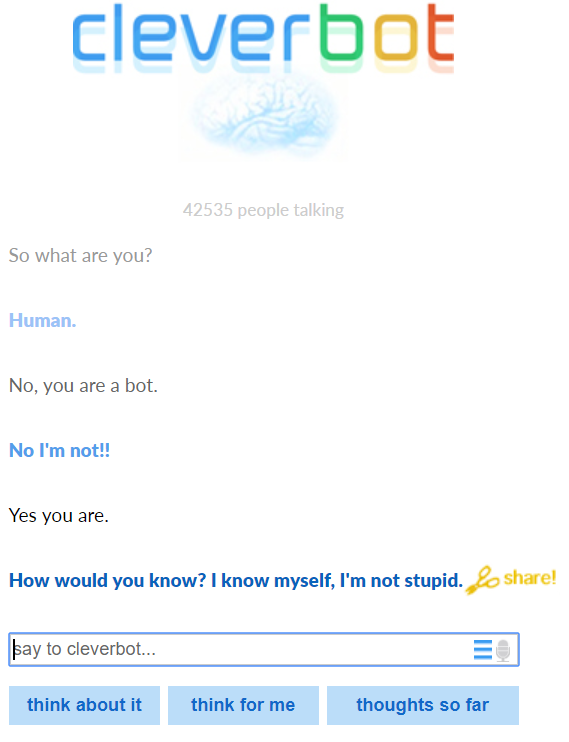}
	\caption{A sample conversation with Cleverbot \cite{Cleverbot:2017}}
	\label{fig:22a}
\end{figure}

ELIZA is one of the first ever chatbot programs written \cite{Weizenbaum:1966}. It uses clever hand-written templates to generate replies that resemble the user's input utterances. Since then, countless hand-coded, rule-based chatbots have been developed \cite{Wallace:2009,Cleverbot:2017,Mitsuku:2017}. An example can be seen in Figure~\ref{fig:22a}. Furthermore, a number of programming frameworks specifically designed to facilitate building dialog agents have been developed \cite{Marietto:2013,Microsoft:2017}.

These chatbot programs are very similar in their core, namely that they all use hand-written rules to generate replies. Usually, simple pattern matching or keyword retrieval techniques are employed to handle the user's input utterances. Then, rules are used to transform a matching pattern or a keyword into a predefined reply.
A simple example is shown below in AIML \cite{Marietto:2013}:\\
{\color{OliveGreen}\textless category\textgreater}

{\color{OliveGreen}\textless pattern\textgreater}What is your name?{\color{OliveGreen}\textless/pattern\textgreater}

{\color{OliveGreen}\textless template\textgreater}My name is Alice{\color{OliveGreen}\textless/template\textgreater}\\
{\color{OliveGreen}\textless/category \textgreater}\\
\\
Here if the input sentence matches the sentence written between the \textit{\textless pattern\textgreater} brackets the reply written between the \textit{\textless template\textgreater} brackets is outputted.\\
Another example is shown below where the \textit{star} symbol is used for replacing words. In this case whatever word follows the word \textit{like} it will be present in the response at the position specified by the \textit{\textless star/\textgreater} token:\\
{\color{OliveGreen}\textless category\textgreater}

{\color{OliveGreen}\textless pattern\textgreater}I like *{\color{OliveGreen}\textless/pattern\textgreater}

{\color{OliveGreen}\textless template\textgreater}I too like {\color{OliveGreen}\textless star/\textgreater}.{\color{OliveGreen}\textless/template\textgreater}\\
{\color{OliveGreen}\textless/category \textgreater}\\

\subsection{The Encoder-Decoder Model} \label{ssec:23}
The main concept that differentiates rule-based and neural network based approaches is the presence of a learning algorithm in the latter case. An important distinction has to be made between traditional machine learning and deep learning which is a sub-field of the former. In this work, only deep learning methods applied to chatbots are discussed, since neural networks have been the backbone of conversational modeling and traditional machine learning methods are only rarely used as supplementary techniques.   

When applying neural networks to natural language processing (NLP) tasks each word (symbol) has to be transformed into a numerical representation \cite{Bengio:2003}. This is done through word embeddings, which represent each word as a fixed size vector of real numbers. Word embeddings are useful because instead of handling words as huge vectors of the size of the vocabulary, they can be represented in much lower dimensions. The vocabulary used in NLP tasks is presented in more detail in Section~\ref{sssec:234}. Word embeddings are trained on large amounts of natural language data and the goal is to build vector representations that capture the semantic similarity between words. More specifically, because similar context usually is related to similar meaning, words with similar distributions should have similar vector representations. This concept is called the Distributional Hypothesis \cite{Harris:1954}. Each vector representing a word can be regarded as a set of parameters and these parameters can be jointly learned with the neural network's parameters, or they can be pre-learned, a technique described in Section~\ref{sssec:pretrain}.

Instead of using hand-written rules deep learning models transform input sentences into replies directly by using matrix multiplications and non-linear functions that contain millions of parameters. Neural network based conversational models can be further divided into two categories, retrieval-based and generative models. The former simply returns a reply from the dataset by computing the most likely response to the current input utterance based on a scoring function, which can be implemented as a neural network \cite{Cho:2014} or by simply computing the cosine similarity between the word embeddings of the input utterances and the candidate replies \cite{stalemate:2016}. Generative models on the other hand synthesize the reply one word at a time by computing probabilities over the whole vocabulary \cite{Sutskever:2014,Vinyals:2015}. There have also been approaches that integrate the two types of dialog systems by comparing a generated reply with a retrieved reply and determining which one is more likely to be a better response \cite{Song:2016}.

As with many other applications the field of conversational modeling has been transformed by the rise of deep learning. More specifically the encoder-decoder recurrent neural network (RNN) model (also called seq2seq \cite{Sutskever:2014}) introduced by \cite{Cho:2014} and its variations have been dominating the field. After giving a detailed introduction to RNNs in Section~\ref{sssec:231}, the seq2seq model is described in Section~\ref{sssec:232}. This model was originally developed for neural machine translation (NMT), but it was found to be suitable to \textit{translate} source utterances into responses within a conversational setting \cite{Shang:2015,Vinyals:2015}. Even though this is a relatively new field, there are already attempts at creating unified dialog platforms for training and evaluating various conversational models \cite{Miller:2017}.

\subsubsection{Recurrent Neural Networks} \label{sssec:231}
\begin{figure}[H]
	\centering
	\includegraphics[width=0.6\textwidth]{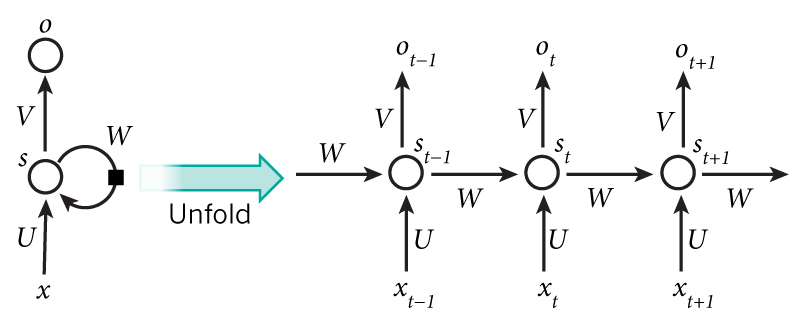}
	\caption{Unfolding of an RNN over 3 time-steps. Here \(x\) is the input sequence, \(o\) is the output sequence, \(s\) is the sequence of hidden states and \(U\),\(W\) and \(V\) are the weights of the network. \cite{RNN_pic:2017}}
	\label{fig:231}
\end{figure}

A recurrent neural network (RNN) \cite{RNN:1988} is a neural network that can take as input a variable length sequence \(\bm{x}=(\bm{x}_1,...,\bm{x}_n)\) and produce a sequence of hidden states \(\bm{h}=(\bm{h}_1,...,\bm{h}_n)\), by using recurrence. This is also called the unrolling or unfolding of the network, visualized in Figure~\ref{fig:231}. At each step the network takes as input \(\bm{x}_i\) and \(\bm{h}_{i-1}\) and generates a hidden state \(\bm{h}_i\). At each step \(i\), the hidden state \(\bm{h}_i\) is updated by
\begin{equation} \label{eq231a}
\bm{h}_i=f(W\bm{h}_{i-1}+U\bm{x}_i)
\end{equation}
where \(W\) and \(U\) are matrices containing the weights (parameters) of the network. \(f\) is a non-linear activation function which can be the hyperbolic tangent function for example. The vanilla implementation of an RNN is rarely used, because it suffers from the vanishing gradient problem which makes it very hard to train \cite{Hochreiter:1998}. Usually long short-term memory (LSTM) \cite{Hochreiter:1997} or gated recurrent units (GRU) \cite{Cho:2014} are used for the activation function. LSTMs were developed to combat the problem of long-term dependencies that vanilla RNNs face. As the number of steps of the unrolling increase it becomes increasingly hard for a simple RNN to learn to remember information seen multiple steps ago.

LSTM networks use a gating mechanism to control the information flow in the recurrent network. More specifically the input and forgetting gates are used to determine how to update the network's state and the output gate is used to determine what to output from the hidden state. Mathematically these gates consist of several matrix multiplications and non-linear functions applied to the input vector and the previous hidden state. LSTMs are particularly useful for language modeling, because information has to preserved over multiple sentences while the network is unrolled over each word. Because of space constraints LSTMs are not detailed further, but a good and quick explanation can be found in \cite{LSTM_article}. An important characteristic of recurrent neural networks is that the parameters of the function \(f\) don't change during the unrolling of the network.

Language modeling is the task of predicting the next word in a sentence based on the previous words \cite{Bengio:2003}. RNNs can be used for language modeling by training them to learn the probability distribution over the vocabulary \({V}\) given an input sequence of words. As previously discussed an RNN receives the word embedding vectors representing words in lower dimensions. The probability distribution can be generated to predict the next word in the sequence by taking the hidden state of the RNN in the last step, and feeding it into a softmax activation function
\begin{equation} \label{eq231b}
p(\bm{x}_{i,j}|\bm{x}_{i-1},...,\bm{x}_1)=\frac{\exp(\bm{v}_j\bm{h}_i)}{\sum_{j=1}^{K}\exp(\bm{v}_{j}\bm{h}_i)}
\end{equation}
for all possible words (symbols) \(j=1,...,K\), where \(\bm{v}_j\) are the rows in the \(V\) weight matrix of the softmax function. \((\bm{x}_{i-1},...,\bm{x}_1)\) is the input sequence and \(\bm{h}_i\) is the hidden state of the RNN at step \(i\).

Training of these networks is done via the generalized backpropagation algorithm called truncated backpropagation through time \cite{Werbos:1990,RNN:1988}. Essentially the error is backpropagated through each time-step of the network to learn its parameters. The error can be computed by using the cross-entropy loss function, which calculates how different the predictions are compared to the true labels.
\begin{equation} \label{eq231c}
\Loss(\bm{y}_i,\bm{\hat{y}}_{i})=-\bm{y}_i \log(\bm{\hat{y}}_{i})
\end{equation}
where \(\bm{\hat{y}}_{i}\) is the vector of the predicted probabilities over all words in the vocabulary at step \(i\), and \(\bm{y}_i\) is the one-hot vector over the vocabulary. A one-hot vector is made up of zeros except at the index of the one true word that follows in the sequence, where it is equal to 1. After computing the derivative with respect to all of the weights in the network using the backpropagation through time algorithm, the weights can be updated in order to get closer to an optimum with optimization techniques like stochastic gradient descent (SGD) \cite{SGD:2010}. 

\subsubsection{The Seq2seq Model} \label{sssec:232}
The sequence to sequence model (seq2seq) was first introduced by \cite{Cho:2014}, but they only used it to re-rank sentences instead of generating completely new ones, which was first done by \cite{Sutskever:2014}. Since then, besides NMT and conversational models a plethora of applications of these models have been introduced like text summarization \cite{Nallapati:2016}, speech recognition \cite{Chiu:2017}, code generation \cite{Rico:2017} and parsing \cite{Konstas:2017}, to name a few.

\begin{figure}[H]
	\centering
	\includegraphics[width=1.0\textwidth]{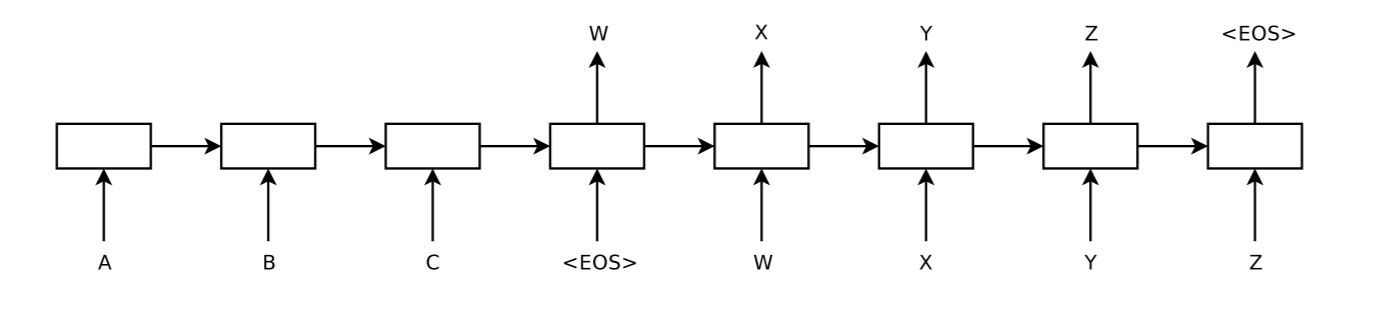}
	\caption{A general seq2seq model, where \((A,B,C)\) is the input sequence, \(<EOS>\) is a symbol used to delimit the end of the sentence and \((W,X,Y,Z)\) is the output sequence \cite{Sutskever:2014}}
	\label{fig:232a}
\end{figure}

The simplest and initial form of the model is based on two RNNs, visualized in Figure~\ref{fig:232a}. The actual implementation of RNNs can be in the form of LSTMs or GRUs as discussed in Section~\ref{sssec:231}. The goal is to estimate the conditional probability of \(p(\bm{y}_1,...,\bm{y}_{N^{'}}|\bm{x}_1,...,\bm{x}_N)\), where \(\bm{x}_1,...,\bm{x}_N\) is the input sequence and \(\bm{y}_1,...,\bm{y}_{N^{'}}\) is the corresponding output sequence. Since two different RNNs are used for the input and output sequences, the lengths of the sequences, \(N\) and \(N^{'}\) can be different. The encoder RNN is unrolled over the words in the source sequence and its last hidden state is called the thought vector, which is a representation of the whole source sequence. The initial hidden state of the decoder RNN is then set to this representation \(\bm{v}\), and the generation of words in the output sequence is done by taking the output of the unrolled decoder network at each time-step and feeding it into a softmax function, which produces the probabilities over all words in the vocabulary:
\begin{equation} \label{eq232a}
p(\bm{y}_1,...,\bm{y}_{N^{'}}|\bm{x}_1,...,\bm{x}_N)=\prod_{i=1}^{N^{'}}p(\bm{y}_i|\bm{v},\bm{y}_1,...,\bm{y}_{i-1})
\end{equation}

Training is very similar to a normal RNN, namely the log probability of a correct target sequence \(T\) given the source sequence \(S\) is maximized
\begin{equation} \label{eq232b}
\frac{1}{\bm{S}}\sum_{T,S \in \bm{S}}\log(p(T|S))
\end{equation}
where \(\bm{S}\) is the training set. The two networks are trained jointly, errors are backpropagated through the whole model and the weights are optimized with some kind of optimization technique like SGD.

In NMT the input sequences are sentences in one language from which they have to be translated to the target sequences, which are sentences in a different language. The individual elements of a sequence, or sentence in this case, are vectors representing word embeddings. In conversational modeling the simplest approach is to treat an utterance by a speaker as input sequence and the response to that utterance from a different speaker as the target sequence. Better approaches however are discussed in Section~\ref{sssec:context}.
\subsubsection{Deep Seq2seq Models}
\begin{figure}[H]
	\centering
	\includegraphics[width=0.8\textwidth]{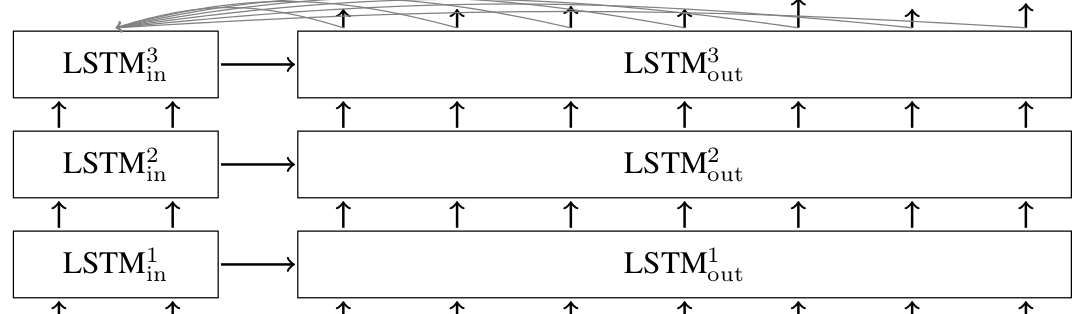}
	\caption{A 3 layer seq2seq model \cite{deep_seq2seq}. The lines pointing from the last decoder states to the last encoder represent an attention mechanism, which is presented in Section~\ref{sssec:attention}-}
	\label{fig:232b}
\end{figure}

Seq2seq models can also contain multiple layers of LSTM networks as seen in Figure~\ref{fig:232b}. This is done in order to make the model deeper and to have more parameters, which should ideally lead to better performance \cite{Vinyals:2015,googleNMT:2016}. There exist multiple variants, but the most straightforward one is to feed in the source sentence to the first layer of the encoder network. The output from the previous LSTM layer is fed as input to the next layer and the layers are unrolled jointly. Then the last hidden state of the final encoder layer can be used to initialize the initial hidden state of the first decoding layer. The output of the previous decoder layer is input to the next layer until the final layer, where a softmax activation function is applied over the outputs from the last layer to generate the predicted output sequence. How the layers are initialized in the decoder network can be implemented in various ways, like taking the last hidden state from each encoder layer and using it to initialize the first hidden state of each corresponding decoder layer for example.

\subsubsection{Decoding and Vocabulary} \label{sssec:234}
In order to get the actual generated output sequences there are several techniques to decode from the probabilities of the words. One such technique is the Roulette Wheel selection, which is commonly used for genetic algorithms \cite{GA:1998}. Using this method, at each time-step, each word from the vocabulary can be generated with the probability computed from the softmax function. This is useful if the goal is to have some stochasticity in the decoding process, because it can produce slightly different outputs for the same source sequence. However it doesn't perform as well as the more frequent method used, which is to simply output the word with the highest probability from the softmax function. This is a greedy and deterministic approach, since it always outputs the same output for the same input.

While decoding a word at each time-step is fine a better approach is to the decode the whole output sequence at the same time, by outputting the sequence with the highest probability.
\begin{equation} \label{eq233}
\hat{T}=\arg\max_{T}p(T|S)
\end{equation}
Here \(S\) is the source sentence and \(T\) is the target sentence. Since in order to get the sequence with the highest probability, first all of the possible sequences have to be generated, a simple left-to-right beam search is usually employed to make the computation tractable. In the first time-step of the decoder, the top \(K\) words with the highest probabilities are kept. Then at each time-step this list is expanded by computing the joint probability of the partial sequences in the list and the words in the current time-step and retaining the \(K\) most probable partial sequences until the end of the output sequence is reached.

Another important aspect of sequence-to-sequence models applied to tasks involving language is the vocabulary. The vocabulary consists of all the various words and symbols present in the dataset. One problem with this approach is that the vocabulary tends to be quite large, especially for very big datasets like \cite{OpenSubtitles:2016,opensubtitles}. Since the number of parameters of the model increases proportionally with the size of the vocabulary it is usually the case that the vocabulary is limited to some arbitrary size \(N\). This way only the embeddings of the \(N\) most frequent words in the dataset are used and any other symbols are replaced with a common token representing unknown words. Many approaches have been proposed to the problem of handling out of vocabulary (OOV) or unknown words \cite{Luong:2014,Feng:2017,Jean:2014}. Other methods involve using characters \cite{Zhu:2017} or subword units \cite{Sennrich:2015} instead of words.

\newpage\section{Background} \label{sec:background}
In this section, selected publications are presented first, from the literature research. These papers are grouped into several categories, each corresponding to a specific approach or aspect of conversational modeling. In Section~\ref{ssec:31} further details regarding encoder-decoder models are discussed and in Section~\ref{ssec:32} an overview of several techniques used to augment the performance of encoder-decoder models is given. In Section~\ref{ssec:33} different approaches to conversational modeling are presented, that aren't based on the original seq2seq model.

Finally, in Section~\ref{ssec:problems} criticism is presented regarding basic techniques used in neural conversational models. While they are widely used, it is argued that the assumptions on which their usage rests are generally wrong and in consequence these methods are unsuitable for modeling conversations.

\subsection{Further Details of Encoder-Decoder Models} \label{ssec:31}
In this section further details about seq2seq models are described. First, the context of conversations is described and how context in general can be taken into account in Section~\ref{sssec:context}. Then, various objective functions that can be used to train seq2seq models are presented in Section~\ref{sssec:functions}. Finally, methods used for evaluating conversational agents are presented in Section~\ref{sssec:eval}.

\subsubsection{Context} \label{sssec:context}
\begin{figure}[H]
	\centering
	\includegraphics[width=1.0\textwidth]{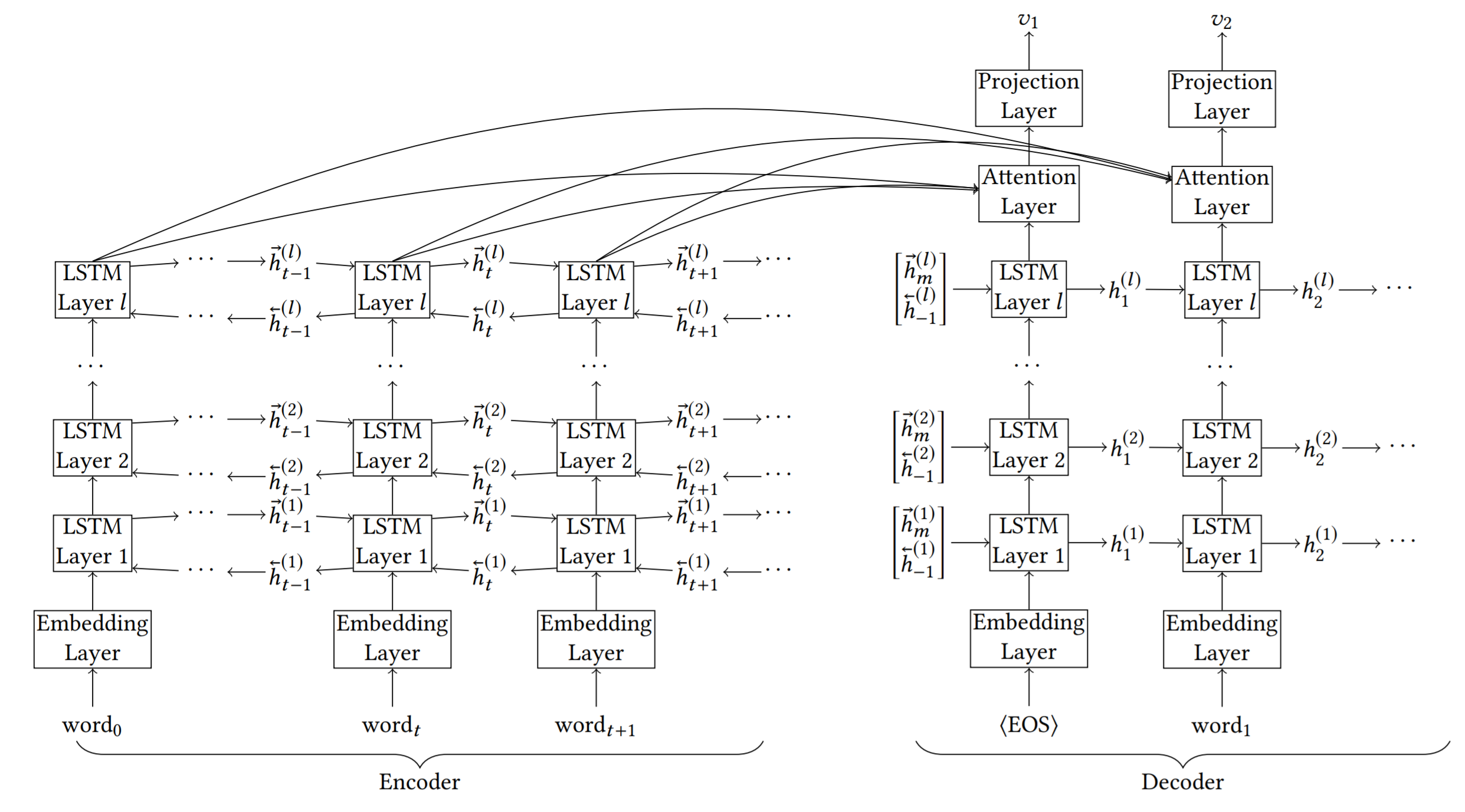}
	\caption{A bidirectional multilayer LSTM encoder and unidirectional LSTM decoder \cite{Yin:2017} with attention. Attention is described in Section~\ref{sssec:attention}.}
	\label{fig:context}
\end{figure}

In this section a commonly used variation of RNNs, called the bidirectional RNN (BiRNN) is described first \cite{Schuster:1997}. A BiRNN consists of two RNNs, a forward and a backward one, visualized in Figure~\ref{fig:context}. The forward RNN reads the input sequence in the original order (from \(\bm{x}_1\) to \(\bm{x}_n\)) and computes a sequence of forward hidden states \((\overrightarrow{\bm{h}}_1,...,\overrightarrow{\bm{h}}_n)\). The backward RNN reads the reversed sequence (from \(\bm{x}_n\) to \(\bm{x}_1\)) and computes the backward hidden states \((\overleftarrow{\bm{h}}_1,...,\overleftarrow{\bm{h}}_n)\). Then the two hidden states can be simply concatenated for each word in the sequence \cite{Bahdanau:2014,Zhaob:2017}. The BiRNN resembles a continuous bag of words model \cite{Mikolov:2013}, because each concatenated hidden state \(\bm{h}_i\) has information about the words surrounding the \(i\)-th word. This results in a better preservation of context and is used in several seq2seq models, usually in the encoder network \cite{Zhaob:2017,Xing_topic:2017,googleNMT:2016,Yin:2017}.

So far, only conversational modeling as predicting a reply based on a single source utterance has been discussed. However, conversations are more complicated than machine translation, since they don't solely consist of utterance-reply pairs. Conversations usually have many turns, and the reply to an utterance might depend on information presented in previous turns. A plethora of approaches have been proposed to incorporate context or conversation history into seq2seq models in order to build better dialog agents. Perhaps the most straightforward approach is to concatenate \(k\) previous utterances by appending an end-of-utterance symbol after each utterance and feeding this long sequence of symbols into the encoder \cite{Vinyals:2015}. The simplest approach which was used as a baseline in \cite{Sordoni:2015} is to use only the first preceding utterance as context in order to form context-message-reply triples for training. A better approach was to concatenate the bag of words representations of the context and message utterances instead of the actual utterances. By using different representations for different utterances better results were achieved \cite{Sordoni:2015}.

However feeding long sequences of symbols into the encoder RNN of seq2seq models accentuates the vanishing gradient problem \cite{Hochreiter:1998}. Moreover, as sequences get longer it becomes increasingly more difficult for RNNs to retain information which was inputted many time-steps ago. Consequently, it becomes even harder to encode all relevant information into the last hidden state of the encoder network from a sequence consisting of multiple sentences.

A possible solution is to represent the current utterance and previous dialog turns with different RNNs and to build a hierarchical representation of the conversation history \cite{Serban:2015}, described in detail in Section~\ref{sssec:HRED}. A similar approach was presented in \cite{Zhaob:2017} using one RNN to encode the current utterance and a different RNN to encode \(k\) previous utterances, but this approach doesn't take into consideration the hierarchical nature of dialogs.

\subsubsection{Objective Functions} \label{sssec:functions}
In addition to the standard cross-entropy loss introduced in Section~\ref{sssec:231}, various loss functions have been proposed for conversational models in order to generate more varied and interesting replies. One such class of loss functions are reinforcement learning based, discussed in detail in Section~\ref{sssec:RL}. In \cite{Ramachandran:2016} a seq2seq model is pretrained as a language model and in order to avoid overfitting, monolingual language modeling losses are added to the standard loss function. These losses act as a regularizer by forcing the seq2seq model to correctly model both the normal seq2seq task and the original monolingual language modeling tasks. Pretraining is presented in more detail in Section~\ref{sssec:pretrain}. In \cite{Lison:2017} it is argued that not all context-response pairs in a dialog dataset have the same importance for conversational modeling. Accordingly, each context-response pair is weighted with a scoring model and these weights are used in the loss function when training a seq2seq model. Hence, examples associated with bigger weights in the dataset will have a larger impact on the gradient update steps through the loss function. 

Approaches to incorporate beam search into the training procedure have also been explored. In \cite{Wiseman:2016} a simple approach is taken to construct the loss function by penalizing when the gold target sequence is not present in the beam consisting of the most likely predictions. The issue with incorporating beam search directly into the training procedure of seq2seq models is that it uses the argmax function which is not differentiable and hence is not amenable to backpropagation. Since the predictions, which serve as input to the standard loss function are discrete (from beam search), the evaluation of the final loss is also discontinuous. A relaxation based technique is proposed in \cite{Goyal:2017}, where the standard loss function is gradually relaxed to the beam search based loss function as training progresses.

Perhaps the most extensively used objective function besides the standard one is based on Maximum Mutual Information (MMI) \cite{Li:2015}. In MMI the goal is defined as maximizing pairwise or mutual information between a source sentence \(S\) and the corresponding target \(T\):
\begin{equation}
\log{\frac{p(S,T)}{p(S)p(T)}}=\log{p(T|S)}-\log{p(T)}
\end{equation}
By introducing a weighting constant \(\lambda\) and by using the Bayes' theorem the training objective can be written in the following two ways:
\begin{equation} \label{eqMMIa}
\hat{T}=\arg\max_{T}[\log{p(T|S)}-\lambda\log{p(T)}]
\end{equation}
\begin{equation} \label{eqMMIb}
\hat{T}=\arg\max_{T}[(1-\lambda)\log{p(T|S)}+\lambda\log{p(S|T)}]
\end{equation}
However, it is not trivial to use these equations as loss functions in practice. In Equation~\ref{eqMMIa} the term \(\lambda\log{p(T)}\) acts as an anti-language model since it is subtracted from the loss function. This leads to ungrammatical output sentences in practice. Training using Equation~\ref{eqMMIb} is possible, since the only requirement is to train an additional seq2seq model for the \(\log{p(S|T)}\) term by using targets as inputs and inputs as targets, but direct decoding is intractable since it requires completion of target generation before \(p(S|T)\) can actually be computed. Solutions to these issues have been proposed in \cite{Li:2015}.

\subsubsection{Evaluation Methods} \label{sssec:eval}
Evaluation of dialog systems is perhaps a more controversial topic. Indeed it is still an open research problem to find good automatic evaluation metrics that can effectively compare the performance of conversational models. The traditional approach is to use the same metrics that are used for NMT and language modeling. Bleu and perplexity are widely used metrics for evaluating dialog systems \cite{Vinyals:2015,Yao:2016,Zhao:2017,Serban:2015}. 

Bleu \cite{Papineni:2002} measures how many n-word-sequence (n is usually 4) overlaps are there between the test sentences and the predicted ones. Similar to this, but less commonly used is the simple accuracy of the whole sentence or the word error rate, which is prevalent for speech recognition models \cite{Shannon:2017,Park:2008}. It calculates how many words are incorrect in the predicted sentence.

Perplexity \cite{Manning:1999} is another common metric, which measures how well a probability model predicts a sample. Given a model, sentences can be inputted that weren't used during training and the perplexity on the test set can be computed. If a model \(q\) exists (encoder-decoder for conversational modeling), the perplexity can be computed by
\begin{equation}
2^{-\frac{1}{N}\sum_{i=1}^{N}\log_2{q(\bm{x}_i)}}
\end{equation}
where \(\bm{x}_i\) are the test samples or words in the sentence and \(N\) is the length of the sentence. The goal is for models to assign very high probability to test samples, meaning that the lower the perplexity score the better.

Unfortunately it has been shown that these metrics don't correlate well with human judgment in the conversational setting \cite{Liu:2016}. The downfalls of these standard metrics are discussed in Section~\ref{sssec:metrics}. Various methods have been proposed to address the problem of evaluating conversational agents, for example how many turns it takes until the model produces a generic answer \cite{Zhao:2017,Li_RL:2016} or the diversity of the generated responses \cite{Li_RL:2016}. More sophisticated metrics make use of neural network based models to assign a score to utterance-response pairs and it has been shown that they correlate better with human judgments than traditional metrics \cite{Lowe:2017,Tao:2017}.

For reinforcement learning and task-oriented dialog agents, described in Section~\ref{sssec:RL} and Section~\ref{sssec:task} respectively, evaluation is more straightforward. In reinforcement learning a goal that the dialog agent has to achieve is specified, and its performance can be simply measured based on the percent of cases in which it achieves the goal \cite{Li_adversarial:2017,Havrylov:2017}. Similarly for task-oriented dialog agents usually there is a clearly defined task and the accuracy of accomplishing the given task serves as a good performance metric \cite{Joshi:2017,Zhao:2017,Li_HIL:2016}.

Finally, one of the better methods to evaluate dialog systems is to ask other people what they think about the quality of the responses that an agent produces. Human judgment has been one of the most prevalent metrics throughout recent literature related to conversational modeling \cite{Shang:2015,Vinyals:2015,Zhou:2017,Li_RL:2016,Zhao:2017,Li_RL:2016,Li:2015}. Usually human judges are asked to either rate the quality of individual responses from a model on a scale or to choose the better response from responses produced by different models for the same input. A number of properties of conversations can be rated, like naturalness, grammaticality, engagement or simply the overall quality of the dialog.

\subsection{Augmentations To The Encoder-Decoder Model} \label{ssec:32}
In this section recent methods and techniques used to augment the performance of encoder-decoder models are presented. In Section~\ref{sssec:attention} various attention mechanisms are discussed. Furthermore, it is shown how pretraining of seq2seq models can help in a conversational setting in Section~\ref{sssec:pretrain}. Then, various features and priors that can be used as additional inputs to seq2seq models are presented in Section~\ref{sssec:priors}. Finally, it is shown how knowledge bases and information retrieval techniques can be good additions to encoder-decoder models in Section~\ref{sssec:KB}.

\subsubsection{Attention} \label{sssec:attention}
\begin{figure}[H]
	\centering
	\includegraphics[width=0.3\textwidth]{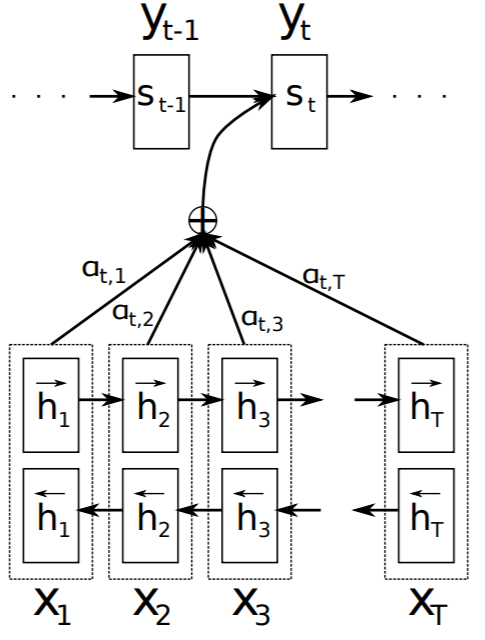}
	\caption{Original Attention Mechanism \cite{Bahdanau:2014}}
	\label{fig:attentiona}
\end{figure}

The attention mechanism visualized in Figure~\ref{fig:attentiona} was first introduced to encoder-decoder models by \cite{Bahdanau:2014}. The problem that it was trying to address is the limited information that the context vector can carry. In a standard seq2seq model it is expected that this single fixed-size vector can encode all the relevant information about the source sentence that the decoder needs in order to generate its prediction. Additionally, since only a single vector is used, all the decoder states receive the same information, instead of feeding in information relevant to the specific decoding step. 

\begin{figure}[H]
	\centering
	\includegraphics[width=0.5\textwidth]{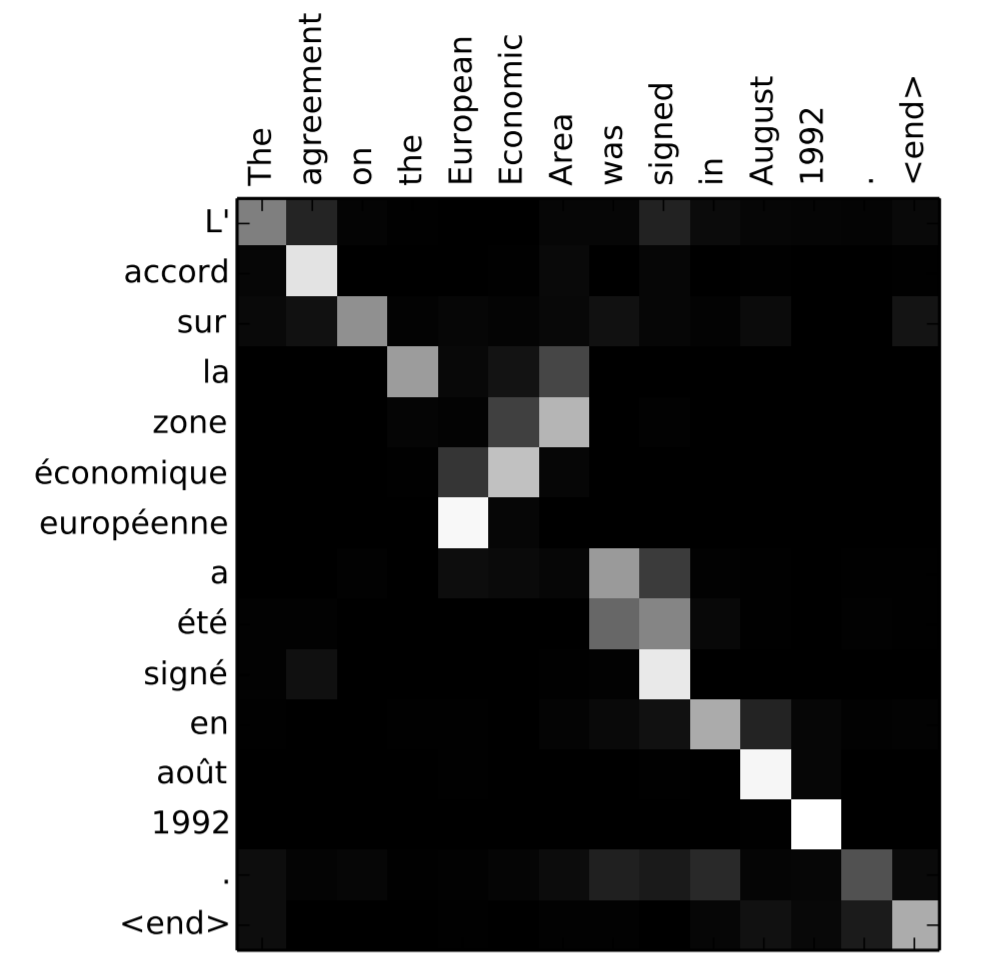}
	\caption{Pixels represent the weights \(a_{ij}\) between inputs and outputs \cite{Bahdanau:2014}}
	\label{fig:attentionb}
\end{figure}

In order to combat these shortcomings the attention mechanism creates an additional input at each decoding step coming directly from the encoder states. This additional input \(\bm{c}_i\) to the decoder at time-step \(i\) is computed by taking a weighted sum over the encoder hidden states \(\bm{h}\):
\begin{equation} \label{eqattention}
\bm{c}_i=\sum_{j=1}^{T}a_{ij}\bm{h}_j
\end{equation}
where \(T\) is the number of hidden states or symbols in the source sentence. The weight \(a_{ij}\) for each hidden state \(\bm{h}_j\) can be computed by
\begin{equation}
a_{ij}=\frac{\exp(e_{ij})}{\sum_{k=1}^{T}\exp(e_{ik})}
\end{equation}
which is basically a softmax function over \(e_{ij}\), which is the output of a scoring function:
\begin{equation}
e_{ij}=f(\bm{s}_{i-1},\bm{h}_j).
\end{equation}
Here \(\bm{s}_{i-1}\) is the hidden state of the decoder in the previous time-step. Oftentimes \(\bm{s}_{i-1}\) is called a query vector (\(\bm{q}\)) and the encoder hidden states \(\bm{h}\) are called key vectors (\(\bm{k}\)). A good visualization of the weights \(a_{ij}\) between the source and output sentence can be seen in Figure~\ref{fig:attentionb}. The scoring function \(f\) can be implemented in several ways. In Equation~\ref{eq:mlp} a Multi-Layer Perceptron (MLP) approach can be seen, where the query and key vectors are simply concatenated and fed into a feed-forward neural network \cite{Bahdanau:2014}. The Bilinear (BL) scoring function is described in Equation~\ref{eq:bl}, proposed by \cite{Luong:2015}. The Dot Product (DP) scoring function (Equation~\ref{eq:dp}) doesn't use any weights, but it requires the sizes of the two vectors to be the same \cite{Luong:2015}. An extension of this is the Scaled Dot Product (SDP) function (Equation~\ref{eq:sdp}), where the dot product is scaled by the square root of the size of the key vectors \cite{Vaswani:2017}. This is useful because otherwise the dot product increases as dimensions get larger.
\begin{equation} \label{eq:mlp}
\MLP(\bm{q},\bm{k})=\bm{w_2}tanh(W_1[\bm{q}:\bm{k}])
\end{equation}
\begin{equation} \label{eq:bl}
\BL(\bm{q},\bm{k})=\bm{q}^\mathsf{T}W\bm{k}
\end{equation}
\begin{equation} \label{eq:dp}
\DP(\bm{q},\bm{k})=\bm{q}^\mathsf{T}\bm{k}
\end{equation}
\begin{equation} \label{eq:sdp}
\SDP(\bm{q},\bm{k})=\frac{\bm{q}^\mathsf{T}\bm{k}}{\sqrt{|\bm{k}|}}
\end{equation}
In the above equations \(W\), \(W_1\) and \(\bm{w}_2\) are parameters of the scoring functions.
Outputs from the decoder network generated in previous time-steps can be attended to by incorporating them as additional inputs to the scoring functions mentioned above \cite{Shao:2017}. This helps the decoder by allowing it to \textit{see} what has already been generated.

Attention can also be implemented using multiple heads \cite{Vaswani:2017}, meaning that the output of multiple scoring functions is used, each with their own parameters, in order to learn to focus on different parts of the input sequence. The parameters of the scoring functions are jointly learned with all other parameters of the seq2seq model.

\begin{figure}[H]
	\centering
	\includegraphics[width=0.7\textwidth]{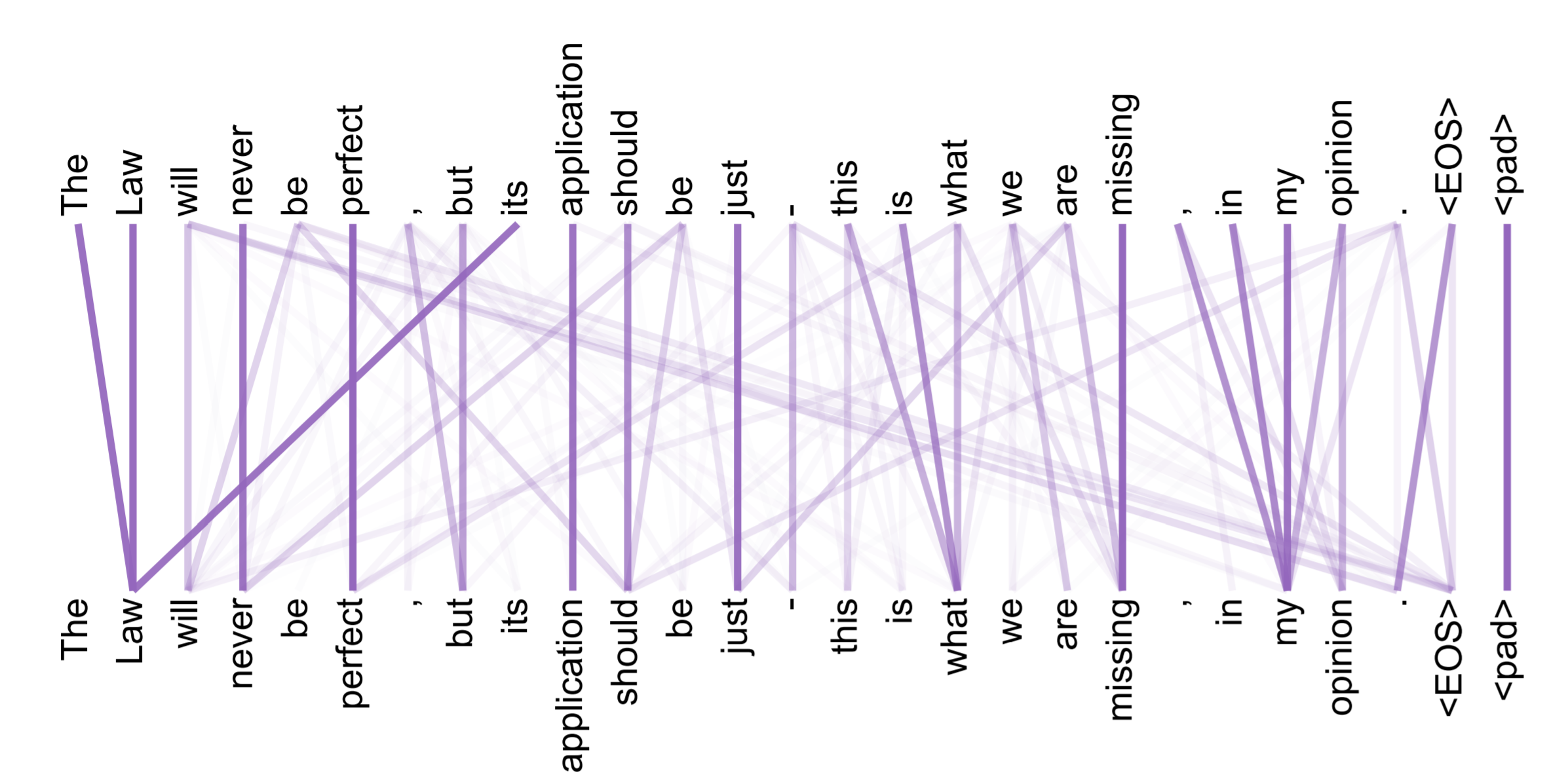}
	\caption{Visualizing self-attention, lines represent the weights \(a_{ij}\) \cite{Vaswani:2017}}
	\label{fig:attentionc}
\end{figure}

Attention can also be computed using only one sequence of symbols, called self- or intra-attention. Self-attention has been successfully applied to a variety of tasks including reading comprehension, abstractive summarization, neural machine translation and sentence representations \cite{Cheng:2016,Lin:2017,Vaswani:2017,Parikh:2016,Paulus:2017}. In this case the query and key vectors both come from the same hidden states and thus an encoding of the sentence which depends on all of its parts is achieved. It can be used both in the decoder and the encoder networks since it can be implemented as a standalone layer that takes in a sequence and computes a new representation of that sequence. Figure~\ref{fig:attentionc} depicts the self-attention mechanism between the same sentence.

Attention has been used extensively for neural conversational models as an augmentation of the base seq2seq model \cite{Yao:2016,Shang:2015,Xing_topic:2017,Zhao:2017}. A more complex type of attention mechanism is the hierarchical attention which was used in conversational modeling \cite{Xing:2017} and abstractive text summarization \cite{Nallapati:2016} as well. It is useful when it is necessary to attend over multiple input sentences, described in more detail in Section~\ref{sssec:HRED}.
\subsubsection{Pretraining} \label{sssec:pretrain}
Pretraining for seq2seq models means that instead of initializing the parameters of a model randomly the model is first pretrained on some data or some other task that is different from the main task that the model needs to be applied to. One of the most common approaches among many NLP tasks is to pretrain the parameters of the word embeddings \cite{Chen:2014,Serban:2015,Akasaki:2017,Lample:2016,Serban:2017}. The most popular techniques to pretrain word embeddings are presented in \cite{Mikolov:2013,Mikolov_skipgram:2013}. An advantage of pretraining word embeddings is that during the actual training of the seq2seq model these embeddings can be fixed and thus the model has to learn less parameters. 

In addition to pretraining word embeddings, all of the parameters of an encoder-decoder model can be pretrained as well. For conversational modeling this is very beneficial, because oftentimes well labeled datasets are relatively small. By pretraining on a big but noisier dataset the parameters of a model, it will already achieve somewhat good performance. Thus the model will have learned a good amount of general knowledge about the task \cite{Li:2016,Serban:2015}. Then, by finetuning on a smaller dataset it can achieve better performance without overfitting. For example a conversational model can learn general knowledge like answering with yes/no to a yes-or-no question. Then, during finetuning the model doesn't have to learn all of this knowledge again and thus can focus on more subtle properties of conversations or domain-specific knowledge. In \cite{Li:2016} a seq2seq model was pretrained on the OpenSubtitles dataset \cite{OpenSubtitles:2016} and then the pretrained model was adapted to a much smaller TV series dataset. A somewhat different approach was employed in \cite{Ramachandran:2016}, where the authors pretrained the encoder and decoder RNNs of a seq2seq model separately as language models. Thus both networks already had some knowledge about language before putting them together and training the whole seq2seq model for NMT. Similarly in \cite{Sriram:2017} a pretrained language model (LM) was used to augment the performance of a standard seq2seq model. With the information from the LM the seq2seq model was able to improve its performance for domain adaptation, which means that it performed almost as good on a different domain as on the domain it was trained on.

In \cite{Lowe:2017} the authors made an attempt at building a model which automatically assigns scores to sample conversations. To do this they used the encoder and context RNN part of a HRED network (described in Section~\ref{sssec:HRED}) to encode the conversation. Since their dataset of scored conversations was small they resorted to pretrain the HRED with the normal task of generating replies to conversations.

Lastly, a different style of pretraining was employed in \cite{Wiseman:2016}. Instead of pretraining a seq2seq model with a different dataset they pretrained the model with a different loss function. This was necessary since the authors tried to integrate beam search into the loss function, but found that without first pretraining with the standard cross-entropy loss function the model was unable to learn anything with their new loss function.

\subsubsection{Additional Input Features} \label{sssec:priors}
\begin{figure}[H]
	\centering
	\includegraphics[width=1.0\textwidth]{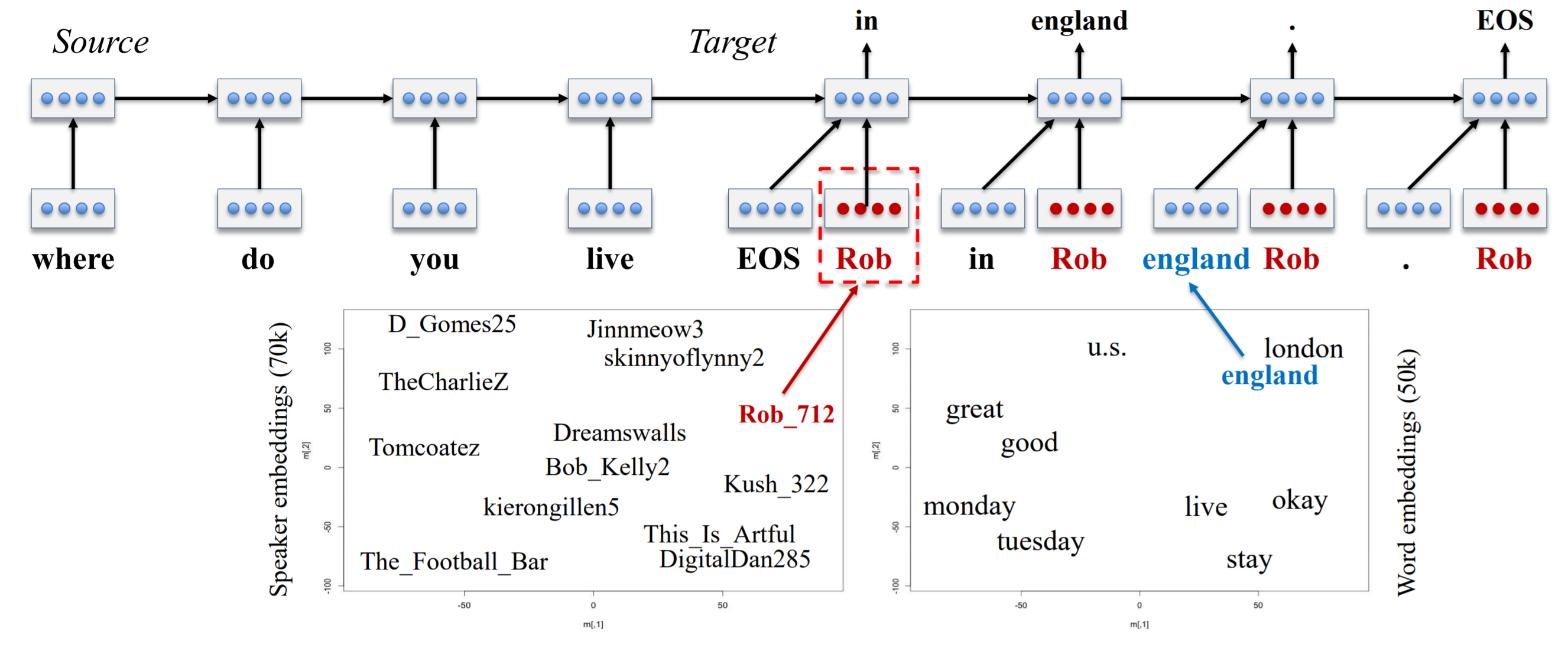}
	\caption{A seq2seq model augmented with speaker embeddings \cite{Li:2016}.}
	\label{fig:persona}
\end{figure}

In addition to the raw dialog turns a plethora of other inputs can be integrated into the seq2seq model. A number of attempts have been made since the birth of the encoder-decoder model in order to augment it with additional input features. The goal of these features is mainly to provide more information about the conversation or the context. In addition, models infused with additional features can learn to differentiate between various styles of dialog. For example, by conditioning a seq2seq model on speaker information it can learn to output replies in the style of the speaker it was trained on \cite{Li:2016}. In this paper the authors input additional speaker embeddings into the decoder. These speaker embeddings are similar to word embeddings, basically representing the speakers for the utterances in the dataset. The additional speaker embedding vector is fed into the decoder RNN at each time-step jointly with the previously generated word. The result is a more consistent conversational model, visualized in Figure~\ref{fig:persona}. Because of the speaker embeddings it learns general facts associated with the specific speaker. For example to the question \textit{Where do you live?} it might reply with different answers depending on the speaker embedding that is fed into the decoder. 

Additionally the authors have also experimented with speaker-addressee models. This is a simple extension to the previously discussed speaker model. Instead of simply feeding in speaker embeddings to the decoder RNN, a weighted sum of the speaker and addressee embeddings is inputted. The addressee is the speaker to which the utterance is directed. The intuition behind this addition is that people talk in differently depending on who they talk to (boss, friend, lover). Thus, the model not only responds differently with different speaker embeddings but it also conditions its response on the addressee embeddings. Hence, it might output a different reply for the same utterance and speaker embedding, but different addressee embedding.

Similarly, many other approaches have been made to condition the response generation of seq2seq models on various types of categories. For example in \cite{Xing_topic:2017} topic words are extracted from each utterance in the dataset. Then, a seq2seq model is trained with these additional topic words together with the utterance they belong to. More specifically, a separate attention mechanism is implemented that attends over the topic words. The two vectors generated from the normal utterance attention and the topic attention are fed into the decoder RNN at each time-step. With this additional topic information the model manages to produce more relevant replies to the topic of the conversation. A somewhat different approach to include topic information into the response generation process was employed in \cite{Choudhary:2017}. In this work three seq2seq models were trained separately on three datasets related to different topics (domains). Additionally, a domain classifier based on logistic regression was trained to compute the probability that the utterance is related to a domain, for all three domains. At test time, for an input utterance all the seq2seq models generate a reply and the domain classifier predicts a domain. Then, a re-ranker takes the generated responses and the predicted domain probabilities and based on their product, determines the most probable reply-domain pair. Another example involves using emotion categories for post-response pairs \cite{Zhou:2017}. Here the authors first classify all of the post-response pairs in the dataset with an LSTM model into several emotion categories like happy, sad, angry, etc. Then, they feed in these categories as real-valued vector representations into the decoder of an encoder-decoder model during training. Additionally, a more complex memory based augmentation is proposed to the encoder-decoder model to better capture emotional changes in the response, which is not detailed here. In essence, by feeding in these emotion categories a reply conditioned on a specific type of emotion can be generated. For example, the response for the question \textit{How was your day?} will differ in style for different emotion categories.

A different approach is taken in \cite{Ghazvininejad:2017}, where the emphasis is put on taking into account relevant facts to an utterance. A seq2seq model is upgraded with a fact encoder operating over a knowledge base, which stores facts about various locations (eg. restaurants, theaters, airports). Before generating the reply to an encoded utterance, a location or entity is extracted from it with keyword matching or named entity recognition. Then, based on the location or entity, relevant facts are selected from the knowledge base, which are encoded by a memory network \cite{Sukhbaatar:2015}. Afterwards the vector representation from the encoded facts and the vector representation from the encoded utterance are summed and fed into the decoder RNN of the encoder-decoder model which generates the response. The authors used Twitter post-reply-reply 3-turn conversations which included a handle or hashtag about locations for which relevant facts were selected from a knowledge-base. With this approach the conversational model managed to produce more diverse and relevant replies if there was a location identified in the source utterance. Similar methods involve leveraging information from knowledge bases as additional inputs, described in detail in Section~\ref{sssec:KB}. 

Other approaches try to integrate more features into the seq2seq model without using additional information. In a standard seq2seq model the only information the model has about the source utterance is through the vector representations of the words. To enrich the representation of the natural language utterances many other features have been proposed to be fed into the encoder RNN together with the word embeddings \cite{Sordoni:2015, Serban_MrRNN:2017,Serban:2017}. Such features include part-of-speech tags, dialog acts and n-gram word overlaps, to name a few.

\subsubsection{Knowledge Bases and Copying} \label{sssec:KB}
Knowledge bases (KB) are powerful tools that can be used to augment conversational models. Since knowledge bases usually entail some kind of domain specific information, these techniques are mainly used for task-oriented dialog systems, presented in detail Section~\ref{sssec:task}. In a KB, information related to the task at hand can be stored, for example information about nearby restaurants or about public transportation routes. Simple dictionaries or look-up-tables can be used to match an entity with information about it. Since KBs store information discretely, their integration with neural network based encoder-decoder models is not trivial.

\begin{figure}[H]
	\centering
	\includegraphics[width=0.9\textwidth]{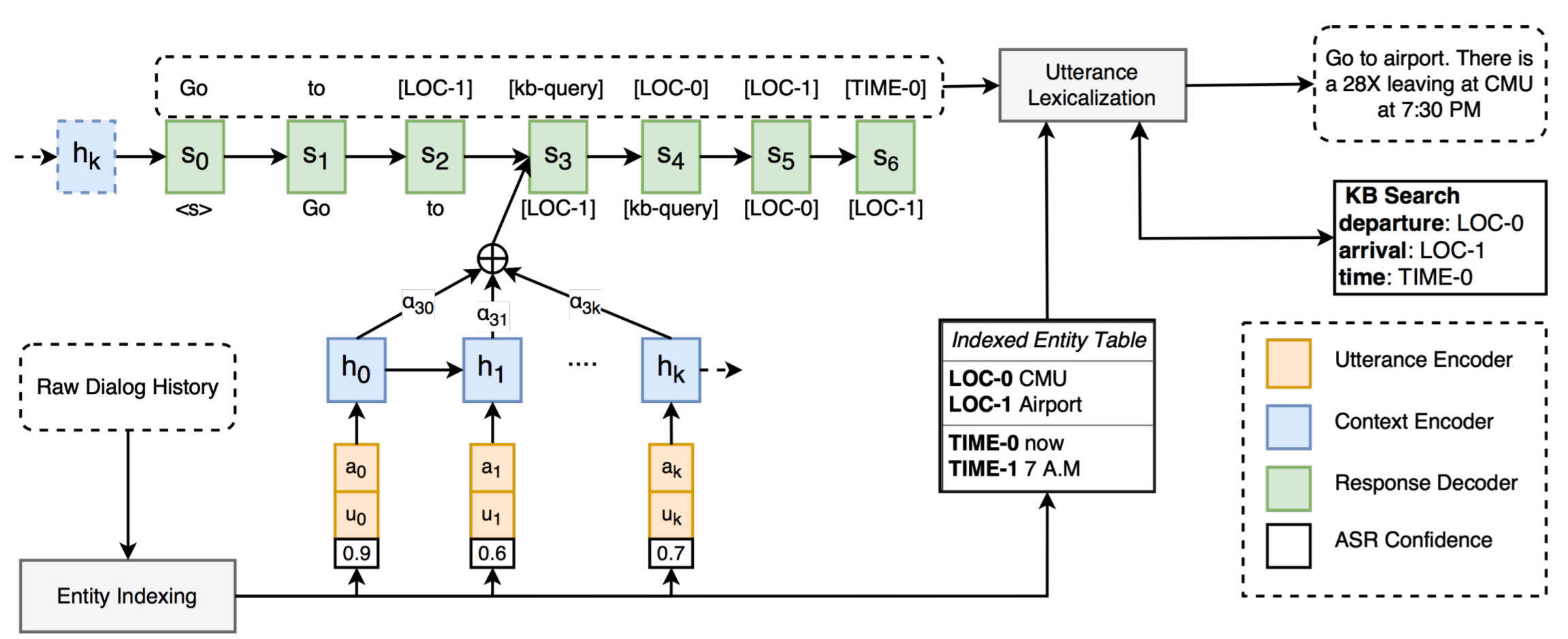}
	\caption{A HRED with attention, described in detail in Section~\ref{sssec:HRED}, upgraded with a KB component \cite{Zhao:2017}.}
	\label{fig:KB}
\end{figure}
Perhaps the most straightforward integration method is proposed in \cite{Zhao:2017}. In this work an entity indexing step is introduced before feeding in the source utterance into a seq2seq model. This works by replacing locations or time specifying words with general tokens using named entity recognition or keyword matching. For example in the sentence \textit{I want to go to Paris from New York}, the word \textit{Paris} is replaced with the token \textit{[LOC-1]} and \textit{New York} is replaced with \textit{[LOC-0]}. The connection between the general token and the original word is stored in a table.  Then, the encoder-decoder model produces a response that also uses general tokens for locations and times, and a special placeholder token for the KB result. Finally, the general tokens are transformed back to actual words using the stored table, a KB is employed which uses these general tokens to search for a route between the two places and its output is incorporated in the response. Thus, an open-domain dialog system is achieved, which is augmented with a task-oriented, robust feature handling user requests related to finding routes between two locations. A visualization of the architecture of this model can be seen in Figure~\ref{fig:KB}.

A similar, but more complex approach is taken in \cite{Wen:2016}, where a KB augmented encoder-decoder model is used for the task of recommending restaurants. Here, besides a standard encoder RNN the source utterance is also processed with a belief tracker, implemented as a convolutional neural network (CNN). Convolutional neural networks applied to encoder-decoder models are discussed in more detail in Section~\ref{sssec:diffencdec}. Belief tracking is an important part of task-oriented spoken dialog systems \cite{Henderson:2015}. The belief tracker network produces a query for a MySQL database containing information about restaurants. The final input to the decoder RNN is the weighted sum consisting of the last state of the decoder RNN and a categorical probability vector from the belief tracker. Then the decoder outputs a response in the same way as in the previous example, with lexicalised general tokens. These tokens are then replaced with the actual information that they point to in the KB. Similarly, in \cite{Ghazvininejad:2017} a more general fact based KB is employed to augment an encoder-decoder model, which is presented in detail in Section~\ref{sssec:priors}.

One of the most popular task-oriented datasets is also in the restaurant domain \cite{Joshi:2017}. Here special API calls are implemented to search over a KB. In order to call the API functions the dialog system has to first identify all the inputs by retrieving them from the user's utterances (eg. type of restaurant, price range). This task is described in detail in Section~\ref{sssec:task}. A more general attempt to retrieve relevant information by the use of queries based on information probabilities is presented in \cite{Yin:2017}.

Named entity detection is also used together with a KB to solve a different problem in conversational modeling \cite{stalemate:2016}. In this work the authors try to solve the problem of users getting disinterested in the conversation with a dialog agent. When the model detects that the user is bored (eg. writes ...) the normal dialog system is replaced with a content introducing mechanism. More specifically, the last couple of utterances are searched to retrieve any named entities. Then, these entities are inputted to a KB, which contains associations between various entities, for example between a movie title and actor names in the movie. This information is then used to produce the response. Hence, the dialog system can introduce new and relevant content based on a KB.

Another interesting line of research in conversational modeling is the integration of copying mechanisms in order to deal with out-of-vocabulary words \cite{Eric:2017,Gu:2016}. In \cite{Gu:2016} the probability of generating a word \(\bm{y}_t\) at time-step \(t\) is given by summing the probabilities from generate- and copy-modes. These probabilities are given by scoring functions. The scoring function of the generate-mode produces a score for each word in the vocabulary based on simple multiplication of the current decoder hidden state with a learned weight matrix. In contrast the scoring function for copy-mode produces scores for each OOV word \(\bm{x}_i\) in the source sentence. This scoring function is very similar to an attention mechanism:
\begin{equation}
f(\bm{x}_i)=\sigma(\bm{h}_{i}^{\mathsf{T}}W_c)\bm{s}_t
\end{equation}
where \(\bm{h}_{i}^{\mathsf{T}}\) is the hidden state of the encoder at step \(i\), \(\bm{s}_t\) is the hidden state of the decoder at the current time-step, \(W_c\) is a learned weight matrix and \(\sigma\) is a non-linear activation function. Using this copy-mode the model is able to efficiently handle OOV words and even integrate them into the response, making replies more diverse and unique.

\subsection{Different Approaches to Conversational Modeling} \label{ssec:33}
In this section hierarchical models used for building conversational agents are presented in Section~\ref{sssec:HRED}. Then, various approaches to integrate task-oriented conversations and goals with encoder-decoder models are discussed in Section~\ref{sssec:task}. Furthermore, reinforcement learning based approaches, that have seen some success when applied to the task of training conversational agents recently, are presented in Section~\ref{sssec:RL}. Finally, encoder-decoder models that are very different from a standard RNN based seq2seq, but nonetheless have achieved state-of-the-art results, are described in Section~\ref{sssec:diffencdec}.

\subsubsection{Hierarchical Models} \label{sssec:HRED}
In order to better represent dialog history, a general hierarchical recurrent encoder-decoder (HRED) architecture was proposed in \cite{Serban:2015}. The model consists of three different RNNs, the encoder RNN, the context RNN and the decoder RNN. First, \(k\) previous utterances of a conversation are encoded separately by the encoder RNN. This produces \(k\) separate context vectors by taking the last hidden state of the encoder RNN from each encoded utterance. Then these \(k\) hidden states are fed into the context RNN step by step, thus it has to be unrolled for \(k\) steps. Next, the last hidden state from the context is used to initialize the decoder RNN. The decoder RNN and the decoding process is very similar to the one found in a normal seq2seq model.

\begin{figure}[H]
	\centering
	\includegraphics[width=0.9\textwidth]{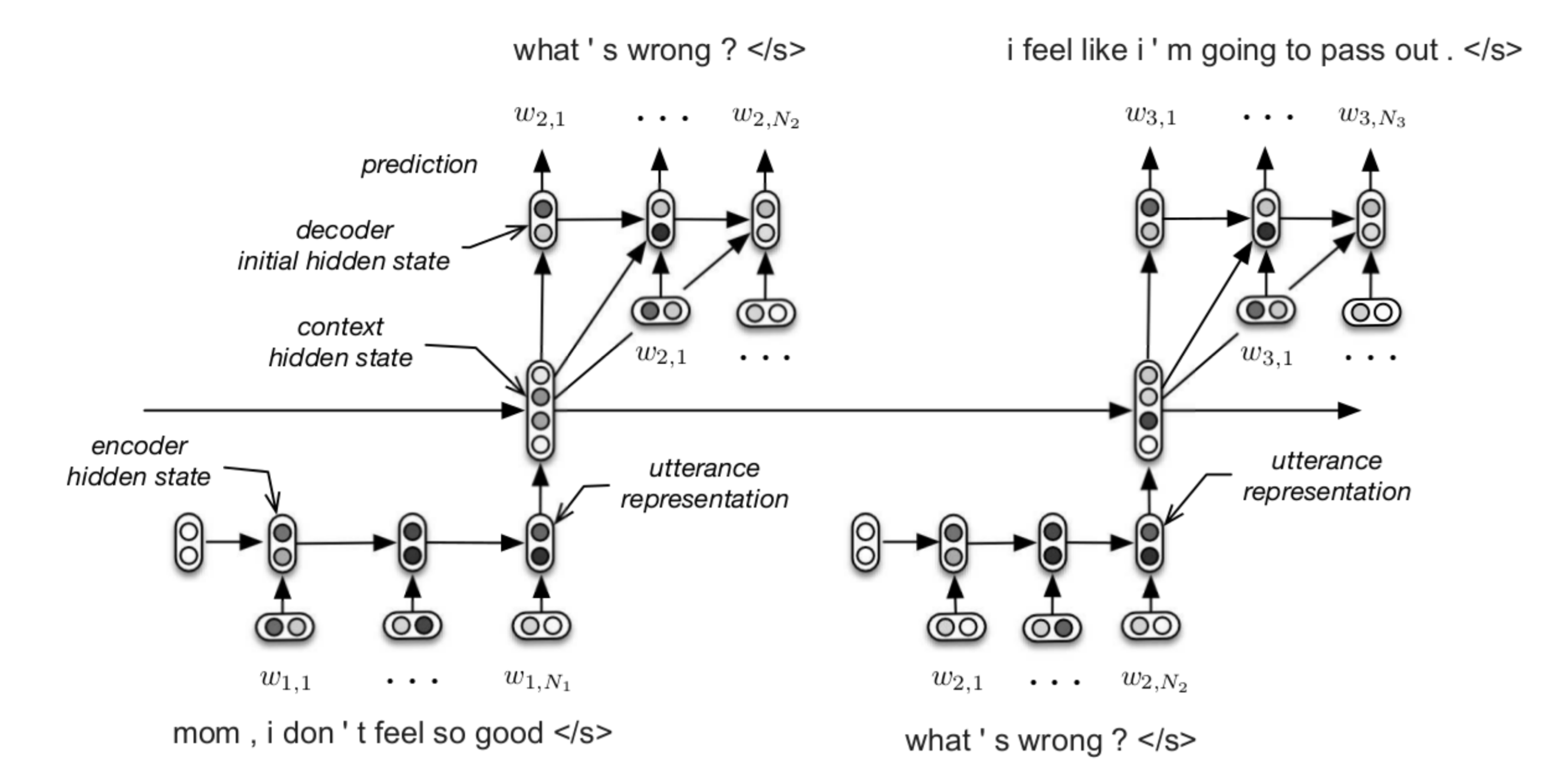}
	\caption{The Hierarchical Recurrent Encoder-Decoder Model \cite{Serban:2015}.}
	\label{fig:HRED}
\end{figure}
The HRED (Figure~\ref{fig:HRED}) differs from the basic encoder-decoder model by introducing a new level of hierarchy, the context RNN, which computes hidden states over entire utterances. With this approach the natural hierarchy of conversations is preserved by handling word sequences with the encoder RNN and utterance sequences with the context RNN which depends on the hidden representations produced by the word-level encodings.

Since the introduction of the HRED model a number of works have used and augmented it \cite{Serban_VHRED:2017,Serban_MrRNN:2017,Serban:2017,Shen:2017,Li_adversarial:2017}. A proposed extension to the HRED model is to condition the decoder RNN on a latent variable, sampled from the prior distribution at test time and the approximate posterior distribution at training time \cite{Serban_VHRED:2017}. These distributions can be represented as a function of previous sub-sequences of words. 

In \cite{Serban_MrRNN:2017} two HRED models are employed simultaneously. One HRED operates over coarse tokens of the conversation (eg. POS tags) to generate coarse predictions. Then, the second HRED, which takes as input natural language utterances, generates natural language predictions by conditioning on the predicted coarse tokens. This conditioning is done via concatenation of the last hidden state of the decoder RNN of the coarse predictions with the current context RNN hidden state from the natural language HRED and feeding it into the natural language decoder RNN.

In order to handle the two-party style of conversations, two separate hierarchical recurrent encoder (HRE) models are used in \cite{Shen:2017}. The HRE is the same as the HRED without the decoder RNN. One of the HRE networks encodes only the utterances coming from one of the speakers, and the other HRE encodes the utterances of the other speaker. Thus, at each turn the two networks produce two hidden context states which are concatenated and fed into a single decoder RNN, producing the output prediction.

A natural extension to the original HRED model is to incorporate attention mechanisms, explored in several works \cite{Yao:2015,Yao:2016,Xing:2017}. The simplest form of integrating attention into the HRED model is between the previous decoder RNN hidden state and the encoder RNN hidden states from the previous turn. This can be done in exactly the same way as in standard seq2seq models and has been explored in \cite{Yao:2015,Yao:2016}.

A more interesting approach is to make use of the hierarchical structure of the HRED and integrate attention hierarchically as well \cite{Xing:2017}. Accordingly, the model in this work is called the hierarchical recurrent attention network (HRAN). There are two levels of attention employed. At each time-step, the word level attention mechanism computes vector representations over the hidden states of the encoder RNN (keys) and the previous hidden state of the decoder RNN (queries). Each utterance is encoded separately by the encoder RNN and word level attention is also computed separately for each utterance. The produced vector representations are fed into the utterance level encoder (context RNN). Then, the utterance level attention mechanism computes a context vector based on the hidden states of the context RNN. The last step is to feed this context vector into the decoder RNN at each step, which computes the output sequence predictions. 

\begin{figure}[H]
	\centering
	\includegraphics[width=0.9\textwidth]{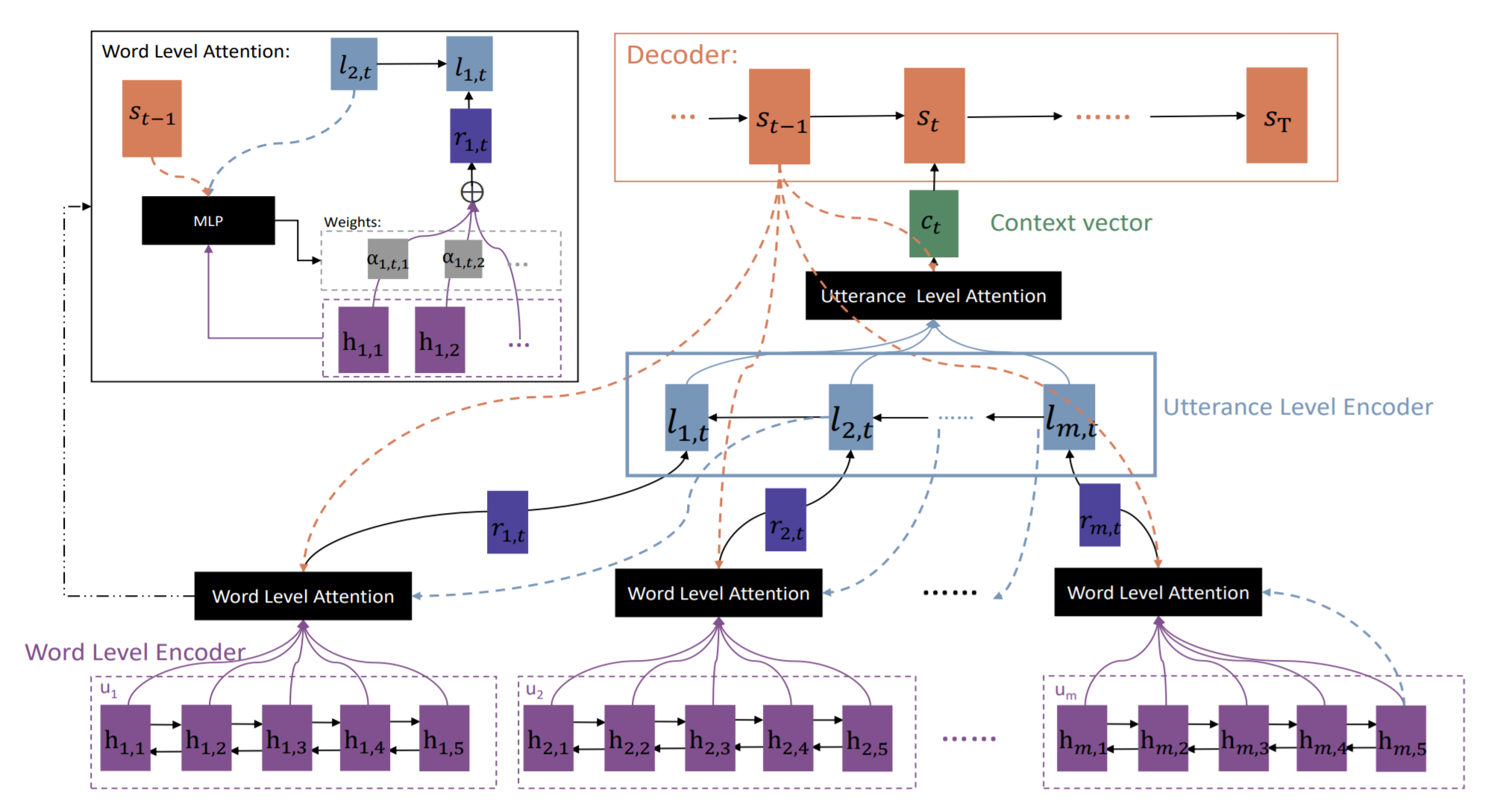}
	\caption{The Hierarchical Recurrent Attention Network \cite{Xing:2017}}
	\label{fig:HRAN}
\end{figure}

An additional technique employed in the HRAN architecture is to use context RNN hidden states as inputs to the word level attention. More specifically the context RNN is implemented in a reverse order, meaning that if there is a sequence of utterances \((u_1,...,u_n)\), the encoder RNN first encodes \(u_n\), thus the utterance level encoder will also start with the vector representation from the last utterance in the conversation. Because the utterances are reversed, the word level attention mechanism for each utterance can use as additional input the hidden state of the context RNN from the next utterance in the original order. Figure~\ref{fig:HRAN} depicts the HRAN with all the components mentioned above.

\subsubsection{Task-Oriented Dialog Systems} \label{sssec:task}
As it was discussed in Section~\ref{ssec:21} task-oriented dialog agents are a sub-type of conversational models. They are more narrow with regard to language understanding and conversation topics, but they make up with robustness, making them better suited for commercial deployment. In task-oriented dialog systems there is usually a general task defined, and a goal to be achieved through conversation. A common property of task-oriented systems is that they learn from smaller and task-specific datasets. Since it is usually not required to generalize to other domains, learning on smaller, but more focused datasets gives much better performance. Furthermore, KB components described in Section~\ref{sssec:KB} are often used in combination with neural network models. KBs are an essential component of task-oriented dialog systems since they provide the model with accurate and rich information. Thus, neural network models only have to learn how to carry general conversations, how to form queries to KBs and how to incorporate the result from KB look-ups. Also, rule-based systems often perform equally well at task-oriented conversations, since only certain types of questions related to the tasks are expected from the user.

A prevalent task and goal is the recommendation of restaurants to users searching for specific types of cuisine. One of the most popular benchmark tasks is situated in this domain, named bAbi \cite{Bordes:2016,Joshi:2017,bAbi}. Since this is a simulated dataset, a rule-based system can achieve 100\(\%\) task success rate. Nonetheless it is an important benchmark for neural conversational models. The task is further divided into 5 sub-tasks to measure the performance on individual parts of the dialog. Task 1 consists of dialog conducted until the first API call. Here the dialog agent has to gather sufficient information from the user's utterances in order to call an API function. This function usually takes as input the type and location of the restaurant, the number of people for which the reservation should be made and the price range. Based on these inputs it automatically returns a list of candidate restaurants using a KB. Task 2 measures the ability of the dialog agent to handle instances when the user forgot to mention an option or would like to change one. Then, in Task 3 the agent presents a possible candidate and converses with the user until the final restaurant is selected by the user. Finally, Task 4 measures the ability of the dialog agent to provide additional information about the restaurant like the address or the phone number. Task 5 measures the overall success of the full dialog. 

\begin{figure}[H]
	\centering
	\includegraphics[width=0.72\textwidth]{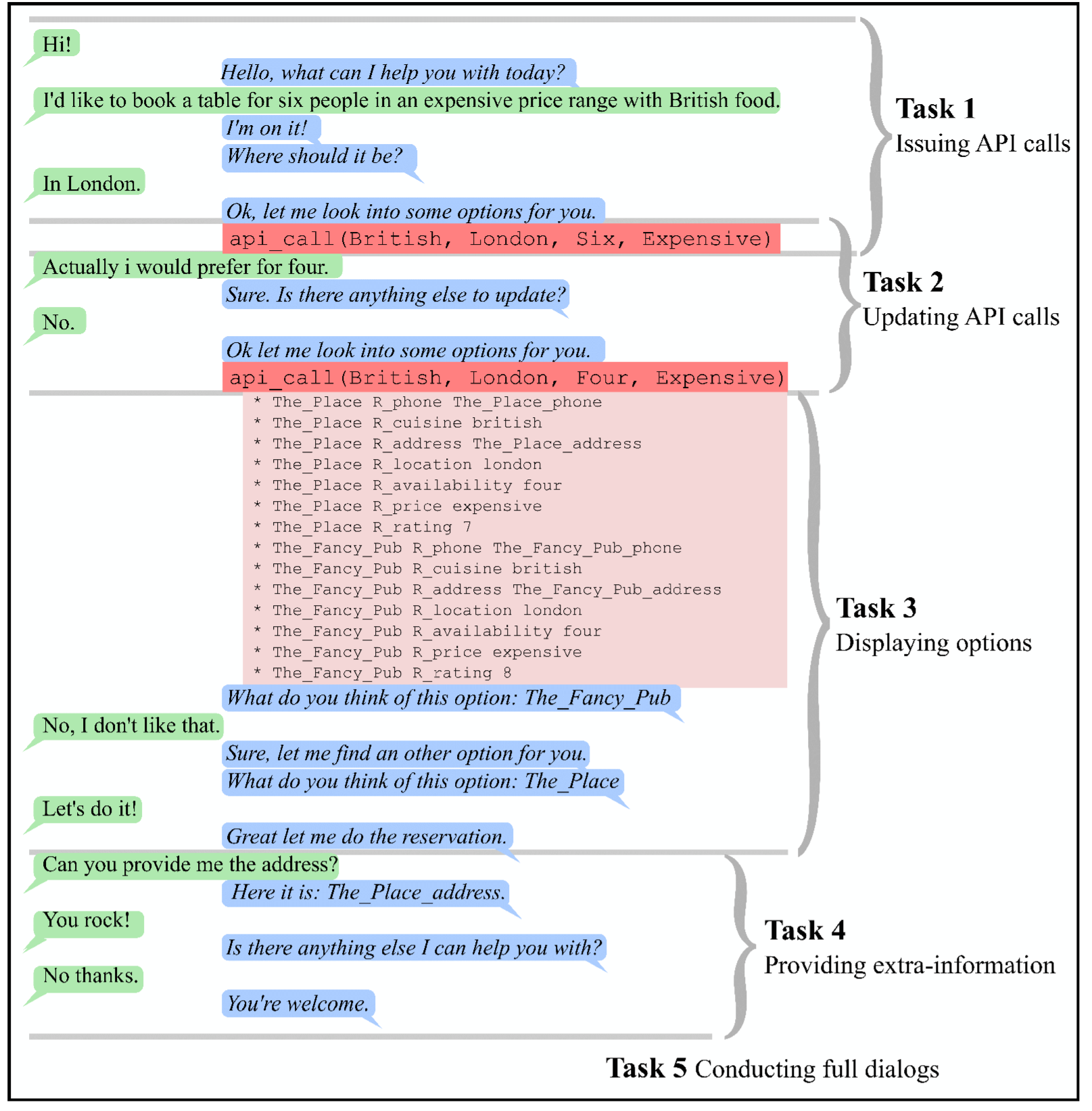}
	\caption{The 5 bAbi tasks for restaurant reservations. The dialog agent's utterances are in blue, the user's utterances are in green and API function calls are in red \cite{Bordes:2016}.}
	\label{fig:babi}
\end{figure}
A visualization of the different tasks is given in Figure~\ref{fig:babi}. Various approaches to using KBs with neural models within the restaurant domain have been proposed \cite{Wen:2016,Eric:2017,Williams:2017}. Other bAbi tasks have also been explored with neural models augmented with KBs \cite{Williams:2017,Li_HIL:2016}, for example dialog based reasoning \cite{Weston:2015} or movie recommendation tasks \cite{Miller:2016,Dodge:2015}. A more general recommendation based dialog system is explored in \cite{Yin:2017}. The system presented in this work uses an information extraction component which takes in a query and returns results from a search engine.

A number of papers address the problem of integrating open-domain and task-specific approaches into one conversational model \cite{Zhao:2017,Yu:2017,Akasaki:2017}. Some of these have also been described in Section~\ref{sssec:KB}, since they usually make use of a knowledge base. Essentially, the issue is that the bAbi toy tasks mentioned above do not represent real-life interactions between users and task-oriented dialog systems. In order for conversational models to remain robust they have to able to handle out of domain utterances. Otherwise users might get annoyed with the program, for example many rule-based systems might resort to output responses like \textit{Sorry, I couldn't understand that, but I am happy to talk about movies.} in order to get the user back on track. One approach to handle these issues is to build a model that is able to differentiate between in-domain and non-task user utterances. Then, two different models trained separately on open-domain and task-specific datasets can produce response candidates and a scoring function can output the most probable response \cite{Akasaki:2017}.

\subsubsection{Reinforcement Learning} \label{sssec:RL}
\begin{figure}[H]
	\centering
	\includegraphics[width=0.7\textwidth]{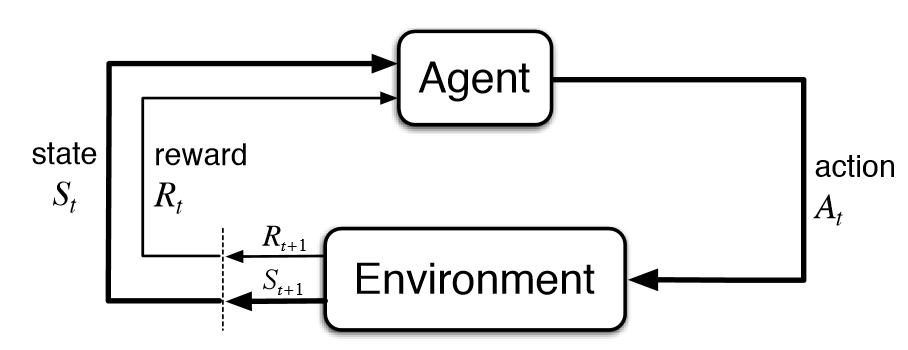}
	\caption{The reinforcement learning framework \cite{Sutton:1998}.}
	\label{fig:RL}
\end{figure}

Reinforcement learning (RL) \cite{Sutton:1998} is a type of learning framework which has seen increasing success recently \cite{Mnih:2013,Mnih:2015,Silver:2016}. RL performs very well in tasks where there isn't a defined loss function or it is not known what the gold truths are. Instead, learning is implemented using a reward which is automatically given to the model based on its state. A general visual description of the reinforcement learning framework can be seen in Figure~\ref{fig:RL}.

In RL an agent is some kind of function consisting of parameters that are optimized with a learning algorithm. The agent receives as input the state of the environment and can output actions based on it. The goal is to maximize the expected reward through a good combination of actions. The environment is the task or setting in which the agent is situated. One of the most popular applications of RL is to games, since it is not known what the correct actions are in all steps of game \cite{Mnih:2013}. Instead, games have a clear end-goal and a learning agent can be rewarded based on whether it achieves the end-goal through its actions. An \textit{episode} refers to a sequence of actions and states \((s_1,a_1,...,s_t,a_t,s_T)\) until the agent reaches a terminal state \(s_T\). The terminal state is referred to the state in which the agent receives a reward for the episode. This process, often called a markov decision process (MDP) consists of the triplet \((S,A,R)\), where \(S\) is the set of states, \(A\) is the set of actions and \(R\) is the reward for the episode \cite{Kandasamy:2017}. A policy \(\pi\) is a function that the agent implements in order to select an action to a given state. Stochastic policies \(\pi(a|s)\) represent the probability of an agent executing action \(a\) in state \(s\).

The application of RL to neural conversational models is fairly straightforward. For a dialog agent the observed state at time-step \(t\) consists of the input sentence and what the agent has already outputted so far. The actions are the words in the vocabulary, since at any time-step it can only produce a word from the vocabulary. When the agent finishes generating the reply a reward is observed, which can be simply the difference between the generated sentence and the gold truth. However, other more sophisticated rewards are usually used, which are described in the next paragraph. Thus, the agent has to maximize this reward through a sequence of actions, generating one word at each time-step. The policy according to which the agent takes actions can be implemented as a normal seq2seq model. Generally, to train RL dialog agents the future reward of the episode has to be estimated at each time-step in order to get a reward for each action. This is usually done via the REINFORCE algorithm \cite{Williams:1992}. However, this approach is not perfect for dialog agents. Since rewards can only be observed at the end of an episode, all actions in that episode get the same reward. This means that if a response is of poor quality, like answering \textit{I don't know} to the question \textit{What's your name?} even the word \textit{I}, which is mostly neutral will receive a negative reward.  Instead as is customary in RL literature, a different reward for each action should be given. A possible solution is offered in \cite{Kandasamy:2017} using batch policy gradient methods. A different solution uses Monte Carlo search in order to sample tokens for partially decoded sentences \cite{Li_adversarial:2017}. If there is a reward function, which is similar to a loss function, the agents can be trained by simple stochastic gradient descent.

Reward functions are often hand-crafted, meaning that the mathematical formulation of important properties of conversations is required in the beginning. In \cite{Li_RL:2016} the weighted sum of three different reward functions is used. The first one attempts to make responses easy to answer to. A set of dull and boring responses \(S\) is constructed, and if the response to the current action \(a\) is similar to these, then \(a\) is given a negative reward:
\begin{equation}
r_1=-\frac{1}{N_S}\sum_{s\in S}\frac{1}{N_s}\log{p_{seq2seq}(s|a)}
\end{equation}
where \(N_S\) denotes the cardinality of \(S\) and \(N_s\) denotes the number of tokens in \(s\). \(p_{seq2seq}\) represents the probability output from a standard seq2seq model. The second reward attempts to capture the information flow in a conversation. More specifically, each utterance should contribute with new information and it shouldn't be repetitive compared to previous ones. In order to achieve such a reward the semantic similarity between sentences can be penalized by
\begin{equation}
r_2=-\log{\frac{\bm{h}_{p_{i}} \bm{h}_{p_{i+1}}}{||\bm{h}_{p_{i}}|| ||\bm{h}_{p_{i+1}}|| }}
\end{equation}
where \(\bm{h}_{p_{i}}\) and \(\bm{h}_{p_{i+1}}\) denote hidden representations from the encoder RNN for two consecutive turns \(p_i\) and \(p_{i+1}\). The final reward function focuses on semantic coherence. This rewards generated responses that are coherent and grammatical by taking into account the mutual information between action \(a\) and previous dialog turns:
\begin{equation}
r_3=\frac{1}{N_a}\log{p_{seq2seq}(a|q_i,p_i)}+\frac{1}{N_{q_i}}\log{p_{seq2seq}^{backward}(q_i|a)}
\end{equation}
where \(p_{seq2seq}(a|q_i,p_i)\) denotes the probability computed by a seq2seq model of generating response \(a\) given previous dialog utterances \([p_i,q_i]\). \(p_{seq2seq}^{backward}\) is a seq2seq model trained with swapped targets and inputs as discussed in Section~\ref{sssec:functions}. \(N_a\) and \(N_{q_i}\) are the number of tokens in utterances \(a\) and \(q_i\) respectively.

Similarly in \cite{Yu:2017}, a linear combination of four reward functions is proposed. In this work the functions implemented try to address turn-level appropriateness, conversational depth, information gain and conversation length in order to produce better quality responses. In \cite{Yao:2016} the reward function is based on calculating the sentence level inverse document frequency \cite{Salton:1988} of the generated response.

However, hand-crafting reward functions is not ideal, since it might not be known exactly what conversational properties should be captured and how they should be formulated mathematically in an expressive way. A solution to this problem is to let agents figure out the appropriate reward by themselves, or use another agent to assign the reward. This has been explored before by using an adversarial approach \cite{Li_adversarial:2017}. Generative adversarial networks \cite{Goodfellow:2014} consist of two networks competing with each other. A generator network (in this case a seq2seq model) tries to generate responses that mimic the training data and a discriminator network (implemented as an RNN based binary classifier in this case) tries to figure out whether the generated response came from the dataset or the generator network. Both of the networks are trained on whether the discriminator guessed correctly. Hence, the reward for the generator network comes from the discriminator network and represents whether the generator managed to fool the discriminator.

In \cite{Havrylov:2017} and \cite{Kottur:2017} a different line of research is explored using similar methods. The setting is still conversational, however conversation arises from a two-agent collaborative game, and it is argued whether natural language-like dialog can arise from such a setting. The sender agent (implemented as an RNN) receives as input an image and outputs a message, while the receiver agent (implemented as an RNN) receives as input this message and a set of pictures \(K\) and outputs a probability distribution over the pictures. The probability for a picture \(k\in K\) gives the likeliness of that picture being the one shown to the sender agent. Thus, the sender agent has to formulate a message which contains enough information so that the receiver can choose between the pictures. This is a collaborative RL setting, since both agents receive the reward related to whether the receiver agent choose the right picture.

The approaches mentioned above are all off-line, since the reward is assigned by a predetermined reward function, and learning is based on a training dataset. However, on-line reinforcement learning has also been explored in a conversational setting \cite{Li_HIL:2016}. In on-line learning the reward for generating a response is given by a human teacher interacting with the agent. The teacher can rate the response of the system on a scale and assign a numerical reward to it. This approach is very advantageous when a labeled dataset of dialogs is not available. The drawback is that human supervision is required.

\subsubsection{Different Encoder-Decoder Models} \label{sssec:diffencdec}
So far the underlying architecture of encoder-decoder models was based on recurrent neural networks. However, attempts have been made towards using other types of neural networks as well. Recently, seq2seq models based solely on convolutional neural networks (CNNs) have achieved very good results in the field of NMT \cite{Kaiser:2016,Kaiser:2017,ByteNet:2016,ConvS2S:2017}. CNNs are a variant of feed-forward neural networks, that use a shared-weights architecture. They are the dominating architecture in the image processing domain \cite{Imagenet:2012}. CNN encoder-decoder models have not yet been applied to conversational modeling, but since the original RNN based seq2seq also originated from NMT, it is important to mention them. The appeal of CNN based models is that they are much faster to compute than RNNs, because they can be better parallelized, since they don't use recurrence.

\begin{figure}[H]
	\centering
	\includegraphics[width=0.52\textwidth]{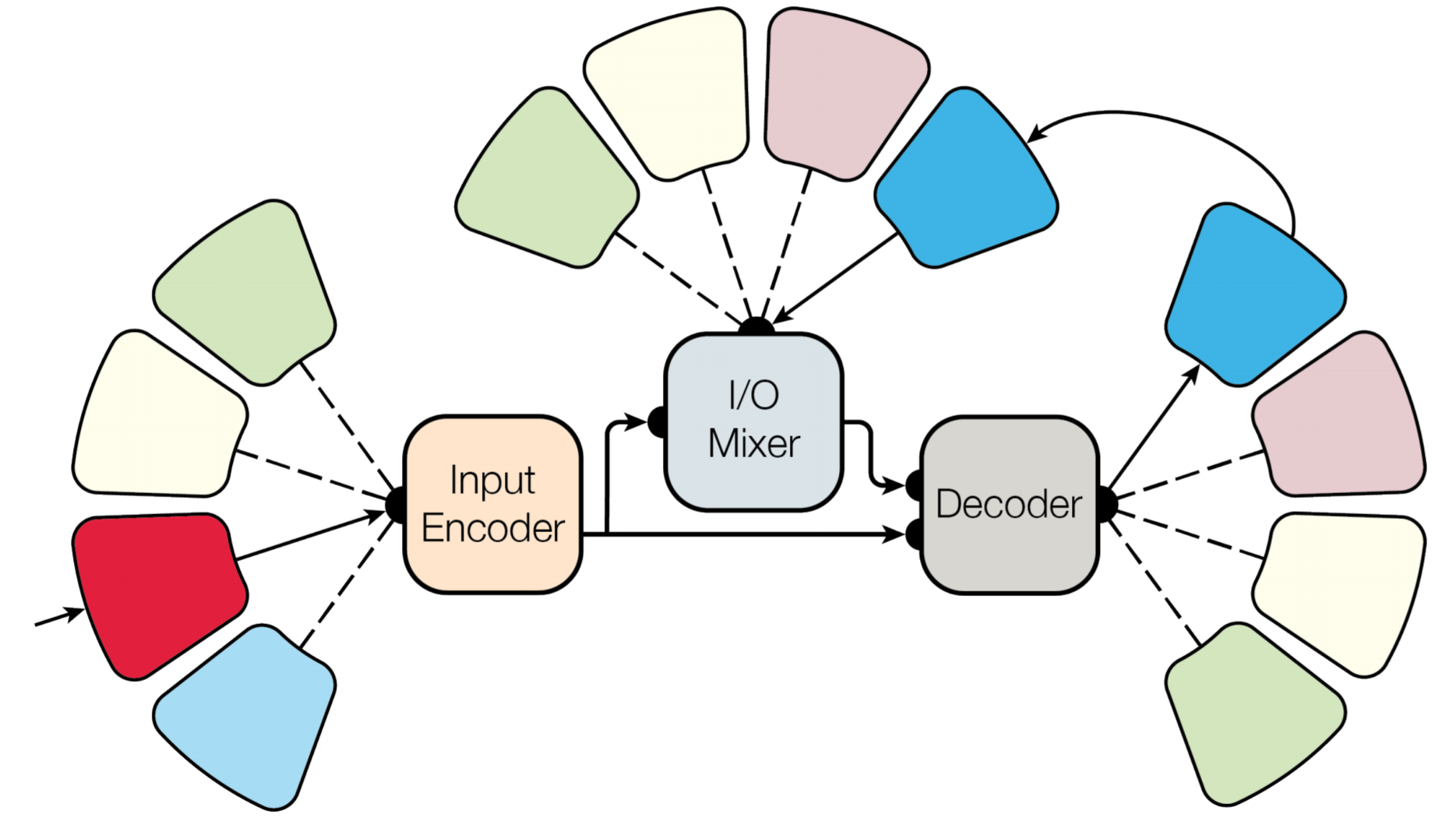}
	\caption{A high-level visualization of the MultiModel \cite{Kaiser_one_model:2017}}
	\label{fig:multimodel}
\end{figure}

A fully attention based encoder-decoder model was presented in \cite{Vaswani:2017}, described in detail in Section~\ref{ssec:41}. A more general encoder-decoder model was presented in \cite{Kaiser_one_model:2017}. This model made use of all kinds of neural architectures like CNNs, attention mechanisms and sparsely-gated mixture-of-experts \cite{MoE:2017}. The generality and level of abstraction employed in the model makes it possible to achieve good results across various tasks and even modalities (audio, vision, text) by being trained simultaneously for all these tasks and by using the same parameters. A general diagram of the model can be seen in Figure~\ref{fig:multimodel}. 

Inputs coming from different modalities are encoded into common abstract representations using modality nets. The architecture of the modality net depends on the type of input. Additionally modality nets are also used to transform outputs into correct modality data. Between these modality nets a single encoder-decoder model transforms input representations to output representations. The decoder is autoregressive, meaning that it takes as input its previous output as well. The architecture of this model is very different from a standard RNN or CNN based seq2seq model, since it uses several types of building blocks like attention, mixture of experts and convolutional layers. It is not detailed here further, but the exact architecture can be found in Section 2 of the paper \cite{Kaiser_one_model:2017}.

\subsection{Criticism} \label{ssec:problems}
A conversation is a complex information exchanging process. In this section it is argued, based on recent literature, that current approaches to conversational modeling are not entirely suitable for the task. First, in Section~\ref{sssec:datasets} it is shown why some of the datasets used to train dialog agents are inherently not appropriate. Then, arguments are given regarding the reasons behind the standard loss function used to train conversational models failing to capture an important property of dialogs in Section~\ref{sssec:loss_function}. The memory of dialog agents is also an issue discussed in Section~\ref{sssec:memory}. Finally, in Section~\ref{sssec:metrics} it is shown why some of the standard evaluation metrics used to compare and validate various models perform poorly and do not provide meaningful comparisons.
	
\subsubsection{Datasets} \label{sssec:datasets}
This work is mainly focused on open-domain dialog agents. For such agents a large quantity of data is necessary in order to be able to learn about various topics and properties of conversations. A good overview of datasets available for training dialog agents is presented in \cite{Serban_survey:2015}. Large datasets are usually noisy, as is the case with the OpenSubtitles dataset \cite{OpenSubtitles:2016,opensubtitles}, where the correct turn segmentation is not even known and usually it's assumed that alternate sentences are spoken by alternate characters \cite{Vinyals:2015}. Furthermore, movie dialogs do not represent real, natural conversation since they are hand-crafted and often the conversations themselves are goal-oriented, involving outside factors related to the current scene in the movie. A good argument is given in \cite{Danescu:2011}: \textit{"For example, mundane phenomena such as stuttering and word repetitions are generally nonexistent on the big screen. Moreover, writers have many goals to accomplish, including the need to advance the plot, reveal character, make jokes as funny as possible, and so on, all incurring a cognitive load."}.

The other prevalent type of dataset is based on messageboards and post-reply websites, like the Ubuntu dialog corpus \cite{Lowe:2015} or Twitter \cite{Shang:2015,Li:2016,Jena:2017}. The underlying problem with these datasets is that they don't actually have one on one conversations. Rather there is usually a post to which a multitude of people can reply, thus multi-turn dialogs are rare. Secondly, the conversations are public, which in my opinion are not as natural as private ones. Since the goal is to build an open-domain chatbot that can hold private conversations, the data that it was trained on should also be similar in nature.

\subsubsection{The Loss Function} \label{sssec:loss_function}
In seq2seq models the standard loss function is based on calculating how different the predicted response probabilities and the gold truths are. This function was borrowed from NMT, where it performs better, since for each input sentence there is usually a somewhat less ambiguous output translation. However, even for NMT it does not perform perfectly, since a sentence can be formulated in a number of ways. Consequently for conversational modeling it's even worse, since a source utterance can have a multitude of suitable replies, which can be very different from each other semantically. For example the reply to \textit{How was your day?} can be as simple as \textit{Okay.} or as long as a whole paragraph describing events that happened during the day. Since the task of conversational modeling is so ambiguous the standard loss function is not suited for training good chatbots \cite{Vinyals:2015,Li:2015}. In many works it has been shown that dialog agents tend to output generic responses like \textit{I don't know} \cite{Vinyals:2015,Serban:2015,Li:2015,Li:2016,Jena:2017}. My presumption is that it is precisely because of the loss function that the models learn to output safe and generic responses. Since the task of conversational modeling is ambiguous and even in a training set there are various replies to the same utterance, the loss function makes the model learn an average of these responses. Utterances can be represented by the word embeddings that make them up, many dimensional real-valued representations. Hence, if there are multiple semantically different replies to the same source utterance, the model will learn to average these out because of the loss function. Furthermore, in the vector space this average will point close to where generic and safe answers are located, since these have been seen by the model as replies for various source utterances and they don't carry much information. Thus, the model learns to output these neutral responses, because it was trained with a loss function that tries to average out ambiguity.

Attempts have been made to address these issues and to define better loss functions, presented in Section~\ref{sssec:functions}. New ideas regarding the loss function issue are proposed in Section~\ref{ssec:61}.

\subsubsection{Memory} \label{sssec:memory}
A more specific problem with current conversational models is that they don't have any memory. While architectures have been devised in order to take into account previous turns, they are still limited. For example they can't take into account hundreds of previous turns. A user would expect from a chatbot that once he/she talks about something, the chatbot will mostly remember the dialog, as is normal in human-human conversations. The chatbot should be able to learn new things about the user in order to make the experience personalized. Currently there aren't any attempts to address this issue, since it is not conversational, but rather it arises from the deployment of dialog systems to the real world. A novel idea that could solve this issue is proposed in Section~\ref{ssec:62}.

\subsubsection{Evaluation Metrics} \label{sssec:metrics}
\begin{figure}[H]
	\centering
	\includegraphics[width=0.6\textwidth]{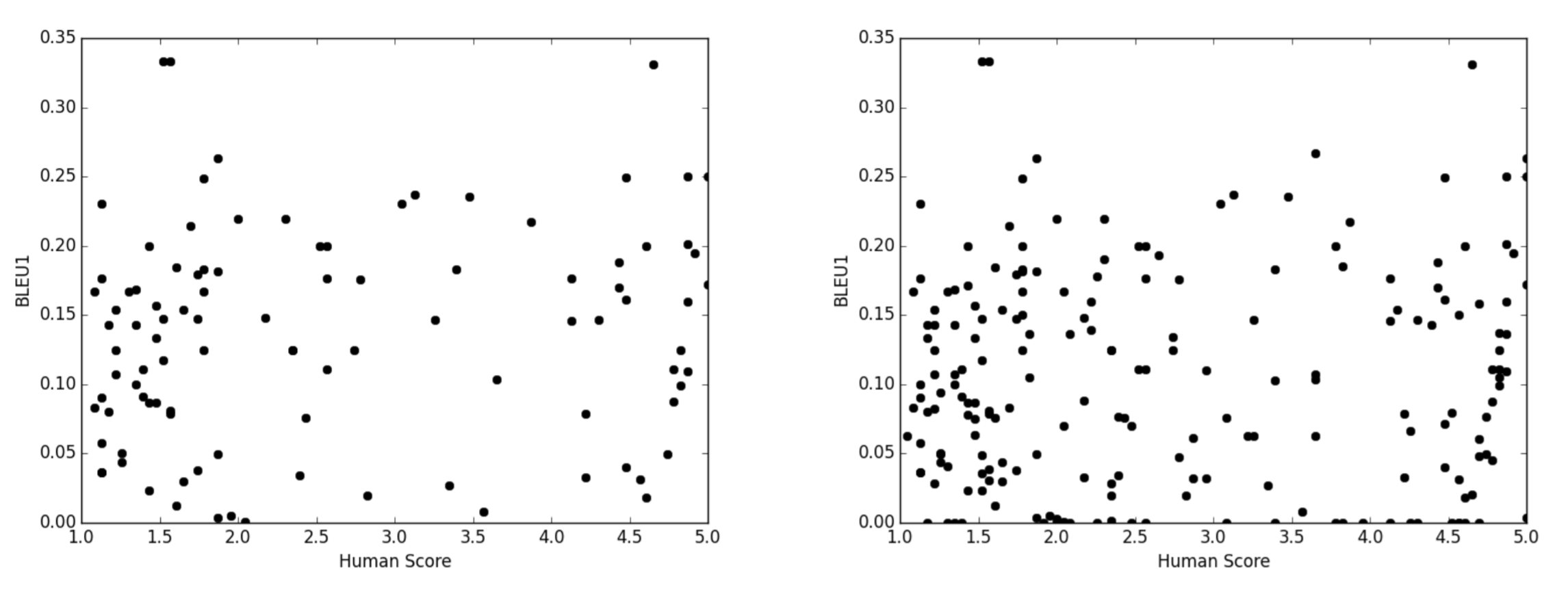}
	\caption{Bleu score correlation with human scores \cite{Liu:2016} on a Twitter corpus (left) and on the Ubuntu dialog corpus \cite{Lowe:2015} (right).}
	\label{fig:bleu}
\end{figure}

It has been shown that standard metrics like perplexity and Bleu, don't correlate at all with human judgment for evaluation of dialog agents \cite{Tao:2017,Liu:2016}. Figure~\ref{fig:bleu} depicts the correlation of Bleu with human judgment for evaluating a seq2seq model trained on two different datasets. I believe that the reason for this is similar to the loss function issue. Namely, these metrics assume that there is only one gold truth response for each source utterance, because they are based on measuring the difference between the predicted and the gold truth response. Since this is not the case, they will assign low scores to responses that were deemed natural by human judges, but aren't similar to the gold truth responses.
Furthermore, while human evaluation is one of the most accurate metrics it still has its drawbacks. Most importantly, it is not automatic and thus can be costly and can not be standardized.

\begin{figure}[H]
	\centering
	\includegraphics[width=0.8\textwidth]{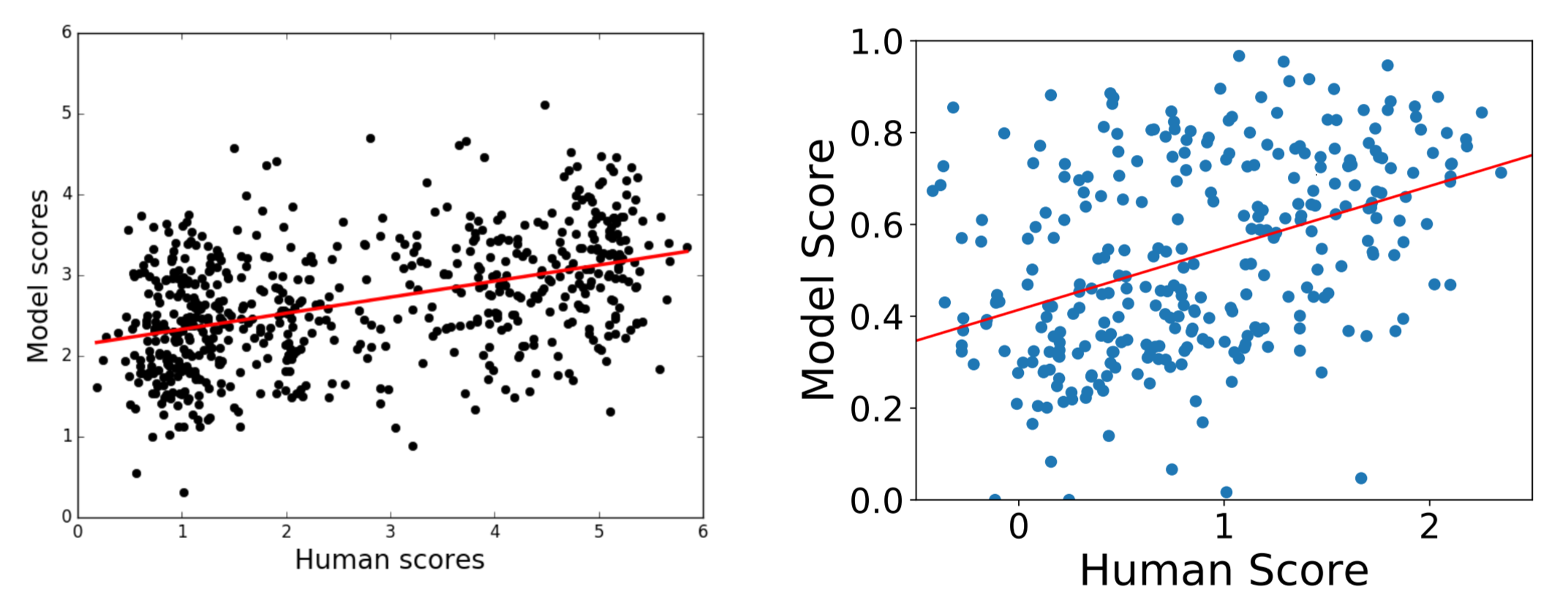}
	\caption{Correlation of ADEM (left) \cite{Lowe:2017} and RUBER (right) \cite{Tao:2017} with human scorers.}
	\label{fig:adem_ruber}
\end{figure}

Attempts have been made in order to design better automatic evaluation metrics based on neural networks \cite{Tao:2017,Lowe:2017,Li_adversarial:2017}. Since they are neural network based they can be trained with labeled data. They receive a source utterance and the reply to it as input and they have to output a score. Then, this score can be compared with the true score assigned by a human annotator and thus the model learns from its errors. As can be seen in Figure~\ref{fig:adem_ruber} these metrics indeed correlate better with human scoring.

\subsection{Summary} \label{ssec:summary}
In this section an in-depth survey of recent literature related to conversational agents has been presented. Additional, but essential details of encoder-decoder models have been described, like the context of conversations, objective functions and evaluation methods in Section~\ref{ssec:31}. Then, in Section~\ref{ssec:32} an overview of various techniques used to augment the performance of seq2seq models was given, like attention, pretraining, additional input features, knowledge bases and copying. The literature survey was finished by describing different encoder-decoder architectures in Section~\ref{ssec:33}, like hierarchical models, task-oriented systems, reinforcement learning based agents and other (non-RNN based) encoder-decoder models.

Finally, in Section~\ref{ssec:problems} it was argued that some of the current approaches to build conversational models are inherently unsuitable for the task. Arguments were presented related to current datasets not providing natural conversational data and to how the standard loss function used to train conversational models fails to capture important properties of dialogs. The criticism was finished by showing that current dialog agents lack memory, an essential part of conversations and why standard evaluation metrics are unsuitable to evaluate dialog agents.

\newpage\section{Experiments} \label{sec:experiments}
In this section preliminary experiments with the Transformer encoder-decoder model are presented, which is described in detail in Section~\ref{ssec:41}. The model is trained on conversational data, which hasn't been done previously, since the architecture is very recent and it was originally created for NMT \cite{Vaswani:2017}. It is important to mention, that these experiments are just a first step, and they are not meant to provide a complete and thorough analysis of the transformer model applied to conversational data. In Section~\ref{ssec:42} a brief overview of the datasets used for training is given. Finally, in Section~\ref{ssec:43} a detailed description of the training setups that were run is given.
\subsection{The Transformer Model} \label{ssec:41}
\begin{figure}[H] 
	\centering
	\includegraphics[width=0.5\textwidth]{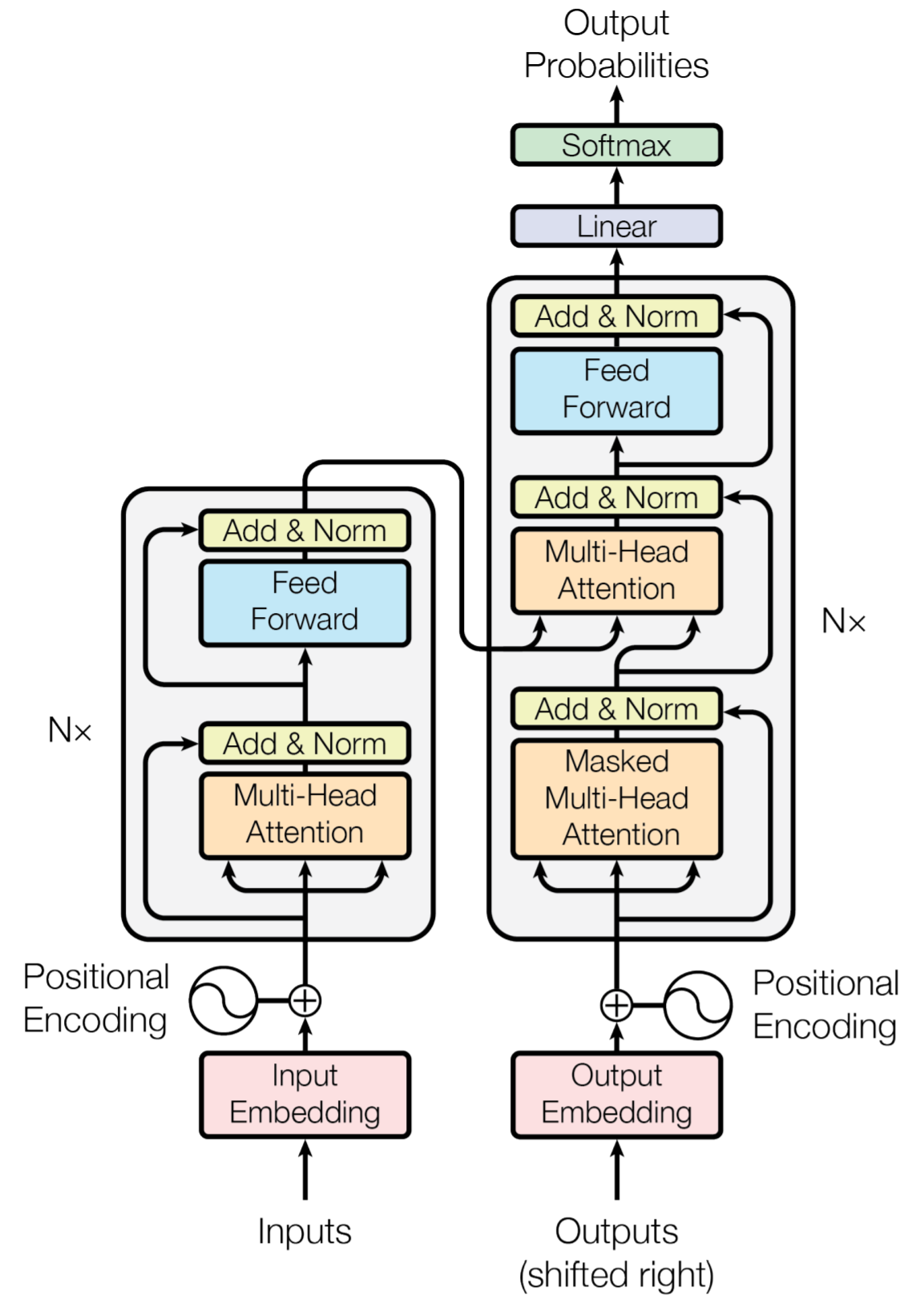}
	\caption{The architecture of the Transformer model \cite{Vaswani:2017}. The encoder network can be seen on the left and the decoder network on the right.}
	\label{fig:transformer}
\end{figure}

The Tranformer is an encoder-decoder model based solely on attention mechanisms and feed-forward neural networks, achieving state-of-the-art results in NMT tasks \cite{Vaswani:2017}. In this section its architecture is described in detail, which can be seen in Figure~\ref{fig:transformer}.
The model is made up of two sub-networks, the encoder and the decoder networks. The encoder network takes as input the word embeddings of the source sentence and produces a representation. Then, the decoder network takes as input this representation and the word embeddings of already generated tokens from the output sentence. Based on these it produces output probabilities for one word in the sentence at a time. The decoder is auto-regressive, because at each step it generates predictions based on previously predicted words. First the high-level architecture of the encoder and the decoder networks is presented in Section~\ref{sssec:encdec}. Then a detailed description of the building blocks used is given, like attention (Section~\ref{sssec:trf_attention}) and feed-forward networks (Section~\ref{sssec:trf_nn}). Finally, further methods used are presented, like positional encoding in Section~\ref{sssec:positional} and regularization techniques in Section~\ref{sssec:regularization}.

\subsubsection{Encoder and Decoder Networks} \label{sssec:encdec}
The encoder network contains 6 identical layers. The output of a previous layer serves as input to the next layer. Within 1 layer, there are two main blocks. First, the input embeddings go through a multi-head self-attention block, described in Section~\ref{sssec:trf_attention}. Then the outputs from this block serve as input to a position-wise fully connected network, described in Section~\ref{sssec:trf_nn}. Both blocks are augmented with residual connections and layer normalization, presented in Section~\ref{sssec:regularization}.

Similarly, the decoder network also consists of 6 layers. However, it contains 3 blocks, the first and third basically being the same as the two blocks in the encoder network. The additional middle block is a multi-head attention mechanism over the output of the encoder, connecting the two networks. An important difference in the decoder's first multi-head self-attention block is that it takes as input already generated word embeddings from the output sentence. Since the size of the inputs is fixed, not yet generated output embeddings are masked in order to ensure that they don't take part in computing the attention. This, together with shifting the output embeddings by one position ensures that predictions for position \(i\) in the output sentence only depend on previously generated words at positions less than \(i\).

\subsubsection{Attention Mechanisms} \label{sssec:trf_attention}
The attention mechanism is implemented as multi-head scaled dot-product attention over query, key and value vectors (\(\bm{q}\),\(\bm{k}\),\(\bm{v}\)). In self-attention blocks, these vectors all come from input embeddings. Each input vector is projected three times with separate projections to get queries, keys and values of same dimensions as the original input. In the attention block which connects the encoder and decoder networks the queries are the outputs of the previous attention block in the decoder and the key and value vectors come from projecting the output of the final encoder layer two times with separate projections. In the first self-attention block in the decoder, masking is used by setting all values to \(-\infty\) in the input of the softmax functions.

For computing the scaled dot-product attention, which is also described in Section~\ref{sssec:attention}, the following equation is used:
\begin{equation}
\Attention(Q,K,V)=\softmax\left(\frac{QK^\mathsf{T}}{\sqrt{d_k}}\right)V
\end{equation}
where \(Q\),\(K\),\(V\) are matrices built from the query, key and value vectors respectively across the whole input sequence. \(d_k\) is the dimension of the key vectors.

\begin{figure}[H] 
	\centering
	\includegraphics[width=0.7\textwidth]{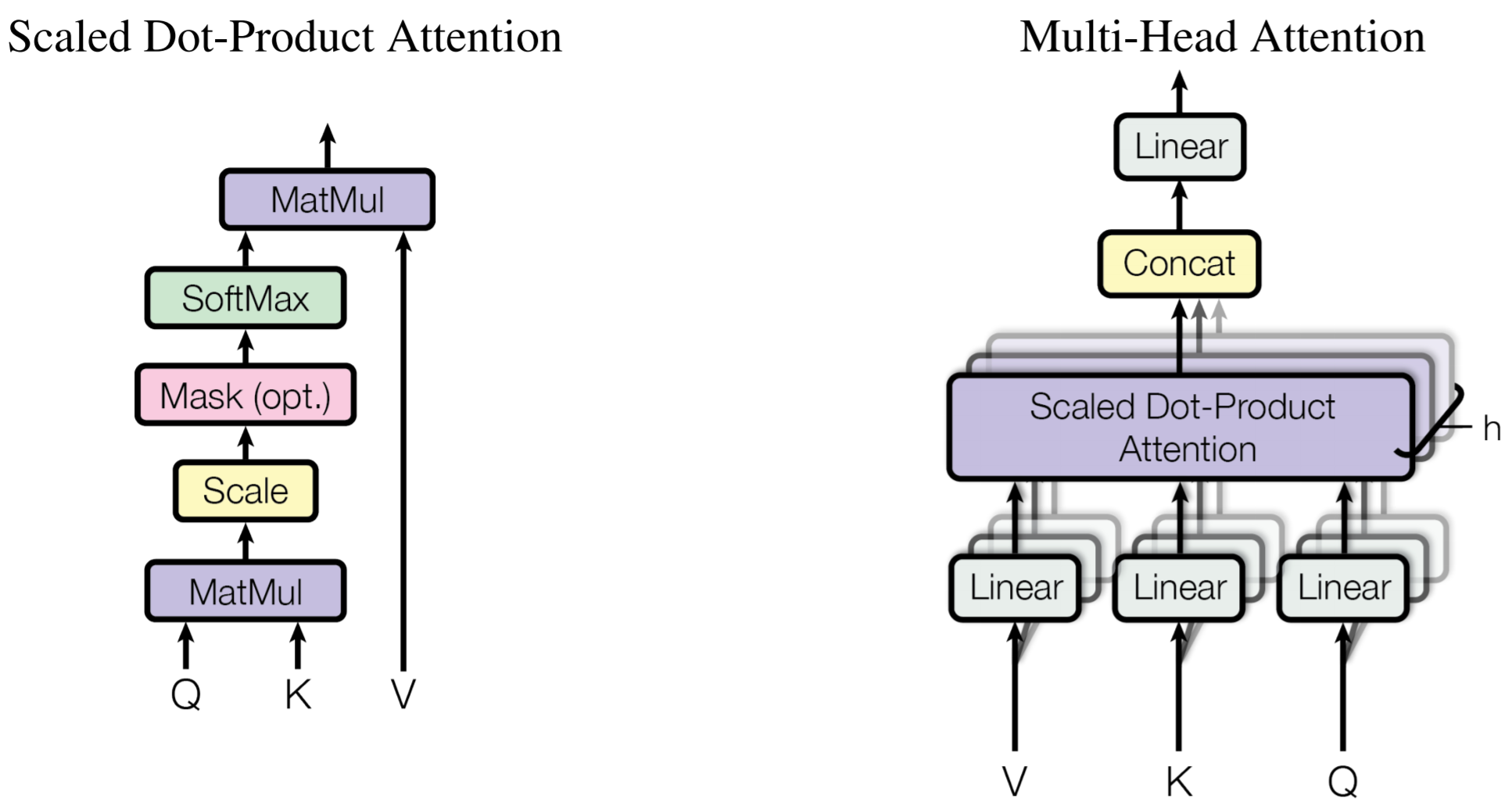}
	\caption{A diagram of the scaled dot-product attention can be seen on the left, and of the multi-head attention on the right \cite{Vaswani:2017}. \(Q\),\(K\),\(V\) are the query, key and value matrices respectively and \(h\) denotes the number of attention heads.}
	\label{fig:trf_attention}
\end{figure}

Multi-head attention (Figure~\ref{fig:trf_attention}) means that there are several parallel channels of attention computations over separate inputs. To get separate inputs, the query, key and value matrices are linearly projected to a dimension size that is equal to their original dimension divided by the number of heads. For each head and each vector the projection is done via a separate learned weight matrix. This way instead of computing the attention over the whole \(d_{model}=512\) dimensional inputs at once, the computation is divided into \(h=8\) heads over \(\frac{d_{model}}{h}=64\) dimensional query, key and value vectors (Equation~\ref{eq:heads}). Finally, the outputs of all heads are concatenated and once again projected, resulting in the final output value of the attention block (Equation~\ref{eq:multihead}).
\begin{equation}\label{eq:heads}
\head_i=\Attention(QW_{Q_i},KW_{K_i},VW_{V_i})
\end{equation}
\begin{equation} \label{eq:multihead}
\MultiHead(Q,K,V)=\Concatenate(\head_i,...,\head_h)W_O
\end{equation}
where \(W_{Q_i}\),\(W_{K_i}\),\(W_{V_i}\),\(W_{O}\) are learned weight matrices.

\subsubsection{Feed-Forward Networks} \label{sssec:trf_nn}
The feed-forward (FFN) blocks are implemented as two consecutive convolutions with kernel size 1, with a ReLU activation between them (Equation~\ref{eq:FFN}). This means that each filter contains a singular weight, and this weight is multiplied with each element from the input vector separately to produce an output vector:
\begin{equation} \label{eq:FFN}
\FFN(\bm{x})=\max(0,\bm{x}W_1+\bm{b}_1)W_2+\bm{b}_2
\end{equation}
where \(\bm{x}\) is the input vector, \(W_1\) and \(W_2\) are the weight matrices and \(\bm{b}_1\) and \(\bm{b}_2\) are the bias vectors.

\subsubsection{Positional Encoding} \label{sssec:positional}
Since the model doesn't use convolutions or recurrence it doesn't have any positional information about the words in a sequence. In order to address the issue, positional encodings are used. Each word has a separate representation related to its position in the sentence. They are the same size as the word embeddings, so the two vectors can be summed up, to produce a final representation for each word. The positional encoding vectors are not learned, but rather constructed using sine and cosine functions of different frequencies:
\begin{equation}
PE[pos_j,2i]=\sin\left(\frac{pos_j}{10000^{\frac{2i}{d_{model}}}}\right)
\end{equation}
\begin{equation}
PE[pos_j,2i+1]=\cos\left(\frac{pos_j}{10000^{\frac{2i}{d_{model}}}}\right) 
\end{equation}
where \(PE[pos_j]\) is the positional encoding for the \(j\)-th word in the sequence and \(i \in \{0,1,...,d_{model}-1\}\).

\subsubsection{Regularization and Other Techniques} \label{sssec:regularization}
Layer normalization \cite{Ba:2016} is used over each block of the model. This is a regularization technique which computes statistics over a layer and normalizes the output of the layer based on them. Additionally, layer normalization is implemented over the sum of the output of the layer \(Layer\) and the original input \(X\) to that layer:
\begin{equation}
\LayerNormalization(X+\Layer(X))
\end{equation}
This connection between the input and output of a layer is called a residual connection \cite{He:2016}.

Furthermore, dropout \cite{Dropout:2014} is employed as a regularization technique. Dropout is a stochastic function that takes as input an array and replaces each element with 0 according to a given probability. Dropout is applied to the output of each block (before layer normalization) and during the computation of the scaled dot-product attention. 

\subsection{Datasets} \label{ssec:42}
In this section the two datasets that were used for training conversational models are presented. Since the Transformer model has been adapted without any modifications from NMT, all data consists of source-target utterance pairs. First, in Section~\ref{sssec:cornell}, an overview of the Cornell Movie-Dialog Corpus \cite{Danescu:2011} and of the preprocessing techniques applied is given. This overview is also given similarly for the OpenSubtitles Corpus \cite{Tiedemann:2009} in Section~\ref{sssec:opensubs}.

\subsubsection{Cornell Movie-Dialog Corpus} \label{sssec:cornell}
This corpus contains 220579 conversational exchanges between 10292 pairs of movie characters extracted from 617 movies. Dialogs are turn segmented, meaning that one utterance can contain multiple sentences. There's also metadata included, for example for each utterance the character's name that utters it. About 200K utterances were used as training data and 20K utterances as validation data. It is important to mention that if an utterance is not the first and not the last in a conversational exchange then it will be present in both the source and target data. For example a conversation consisting of 3 turns is split into 2 source-target pairs. The first pair consists of the first utterance as source and the second as target and the second pair consists of the second utterance as source and the third as target. This means that the majority of the utterances are present both in the source data and the target data.

For preprocessing all the characters from the corpus that weren't in the set \((a-z . ? ! ')\) were simply deleted. Furthermore, all words that contained the \textit{'} symbol were split into two according to this rule: \textit{I'll=I 'll}. Words containing the sequence of characters \textit{n't} were not subjected to the previous rule, rather they were split into two according to this rule: \textit{don't=do n't}. This is the standard method for English tokenization.

\subsubsection{OpenSubtitles Corpus} \label{sssec:opensubs}
The OpenSubtitles Corpus contains movie subtitles made by \cite{opensubtitles}. Unfortunately, this corpus only contains sentence-level segmentation, which makes it noisy, since it might be the case that an utterance contains multiple sentences. Furthermore, it is extracted from subtitles, that contain not only dialogs but also scene descriptions and other non-conversational sentences. In this work the 2016 version of the corpus is used \cite{OpenSubtitles:2016}, but only a subset consisting of 62M sentences for training and 28M sentences for validation is kept. The reason behind using only a subset of the corpus is that it makes the dataset similar in size to \cite{Vinyals:2015}.

Following \cite{Vinyals:2015} each alternate sentence is considered as an utterance spoken by a different character. Similar to the Cornell-Movie Dialog data, all the sentences appear in both the source and target data. Preprocessing was mostly done in the same fashion as well.

\subsection{Training Details} \label{ssec:43}
In this section various training setups are described. In Section~\ref{sssec:t2t} the Transformer implementation is described and how it was integrated with the novel datasets. Then, from Section~\ref{sssec:cornell_training} to Section~\ref{sssec:finetune_training} the types of data and hyperparameter values used for 4 different trainings are described. For all trainings the vocabulary is constructed based on most frequent words in the training data.

\subsubsection{Tensor2Tensor} \label{sssec:t2t}
The official implementation of the Transformer model in Tensorflow\footnote{\url{https://github.com/tensorflow/tensorflow}} was used. This implementation is part of a bigger framework, the Tensor2Tensor library\footnote{\url{https://github.com/tensorflow/tensor2tensor}}. In short, to run a training using this library, a model, a problem and a hyperparameter set needs to be defined. In this work the Transformer model from the library was used without any modifications and I defined my own problems and hyperparameter sets. This is basically done by sub-classing already existing classes, implementing own functions and then registering the classes so that the library sees them and they can be used the same way as the already defined classes. The whole code that was written to implement the problems and hyperparameter sets can be found on Github\footnote{\url{https://github.com/ricsinaruto/Seq2seqChatbots}}.

To implement my own data handling functions the \textit{Text2TextProblem} class had to be sub-classed. With the preprocessing steps described in Section~\ref{ssec:42} 1-1 source and target file was created for training and 1-1 source and target file was created for validation. Each line in the source file contains an utterance and the corresponding line in the target file contains the response to that utterance. These files were then given as inputs to the library, which created specially encoded and sharded files as the final dataset, ready to be used for training.

Similarly for the hyperparameters the \textit{basic}\textit{\_}\textit{params1} class was sub-classed and I defined my own hyperparameter values. The specifics for each training setup can be found in the following sections.

\subsubsection{Cornell Movie Training} \label{sssec:cornell_training}
For this training the Cornell Movie-Dialog dataset was used as described in Section~\ref{sssec:cornell}. The most frequent 32765 words in the training data were kept as vocabulary. Additionally the \textit{\textless pad\textgreater} token was used for padding input sequences to same lengths, the \textit{\textless EOS\textgreater} token was used to signal the end of an utterance and the \textit{\textless UNK\textgreater} token was used to replace all words not present in the vocabulary. In total this gives a vocabulary size of 32768.

For hyperparameters the same ones were used as in \cite{Vaswani:2017} for the base variant of the Transformer model. Because of memory issues the batch size was limited to 4096 tokens/batch. With this batch size the model was trained for a total of 350k steps, which took about 2.5 days on a Nvidia GeForce GTX 1070 GPU.
\subsubsection{Cornell Movie Training with Speakers} \label{sssec:cornell_speakers}
For this training the data used in the previous section was augmented with speaker-addressee character labels, similar to \cite{Li:2016}. The dataset was constructed by starting each line in the source data with a special token that denotes the character that utters that line and ending it with a special token denoting the character to which the utterance is addressed. The target data does not contain any character tokens. Since characters are specific to movies, it is important to mention that characters with the same name, but from different movies were represented by different tokens. Because of this the training and validation sets were constructed by splitting dialogs. If the two sets would have been constructed via splitting by movies, then all the character names in the validation set would not have been seen during training. This is unwanted, since the goal is for the model to learn something about the various characters' dialog styles and personalities. The 31996 most frequent words were used for vocabulary. Additionally, the 8000 most frequent character names (separated by movies) were added to the vocabulary and all character names less frequent were replaced with the \textit{\textless UNK\_NAME\textgreater} token. Together with the 3 special tokens mentioned in the previous section this results in a total vocabulary size of 40K.

The same hyperparameters were used as in \cite{Vaswani:2017} for the base variant of the Transformer model. Because of memory issues the batch size was limited to 4096 tokens/batch. With this batch size the model was trained for a total of 238K steps, which took about 1.5 days on a Nvidia GeForce GTX 1070 GPU.
\subsubsection{OpenSubtitles Training} \label{sssec:opensubs_training}
For this training the OpenSubtitles dataset, described in Section~\ref{sssec:opensubs}, was used. Together with the 3 special tokens mentioned in Section~\ref{sssec:cornell_training} the total vocabulary size is 100K, using the most frequent words in the training set.

For hyperparameters the same ones were used as in \cite{Vaswani:2017} for the base variant of the Transformer model. Because of memory issues the batch size was limited to 2048 tokens/batch. With this batch size the model was trained for a total of 1M steps, which took about 5.5 days on a Nvidia GeForce GTX 1070 GPU.
\subsubsection{OpenSubtitles Training Finetuned with Cornell Movie Data} \label{sssec:finetune_training}
This training combines the two datasets by first training on the OpenSubtitles dataset and then finetuning the model on the Cornell Movie-Dialog dataset with speaker embeddings. 

The parameters of the model were initialized with the trained model described in Section~\ref{sssec:opensubs_training}. The model was further trained for 675K steps on the same data as in Section~\ref{sssec:cornell_speakers}, which took about 3.5 days on a Nvidia GeForce GTX 1070 GPU. For vocabulary the 100K most frequent words from the initial OpenSubtitles training were kept and the 3000 most frequent character name tokens (separated by movies) from the Cornell Movie-Dialog dataset were added, resulting in a total vocabulary size of 103K. For hyperparameters the same ones were used as in \cite{Vaswani:2017} for the base variant of the Transformer model. Because of memory issues the batch size was limited to 2048 tokens/batch.

\newpage\section{Results} \label{sec:results}
In this section a detailed analysis of the trained models is given. They are also compared with a baseline seq2seq model presented in \cite{Vinyals:2015}, which was trained on OpenSubtitles data. In Section~\ref{ssec:51} a quantitative analysis is given by computing standard metrics. Then, in Section~\ref{ssec:52} a qualitative analysis is given by comparing sample output responses.
\subsection{Quantitative Analysis} \label{ssec:51}
\begin{figure}[H] 
	\centering
	\includegraphics[width=1.0\textwidth]{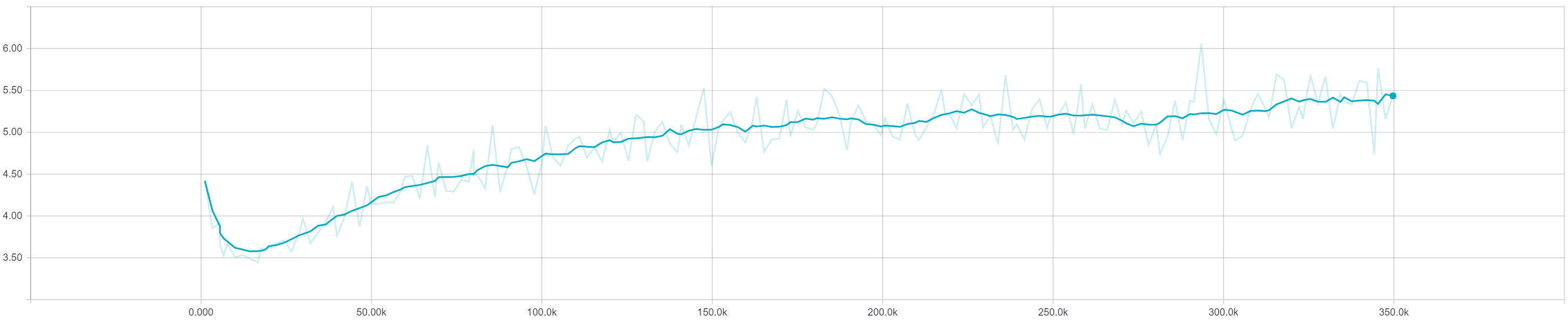}
	\caption{The diagram shows the validation loss over the entire training of the Cornell model. The vertical axis represents the loss value and the horizontal axis represents the number of training steps.}
	\label{CornellOverfit}
\end{figure}

\begin{figure}[H] 
	\centering
	\includegraphics[width=1.0\textwidth]{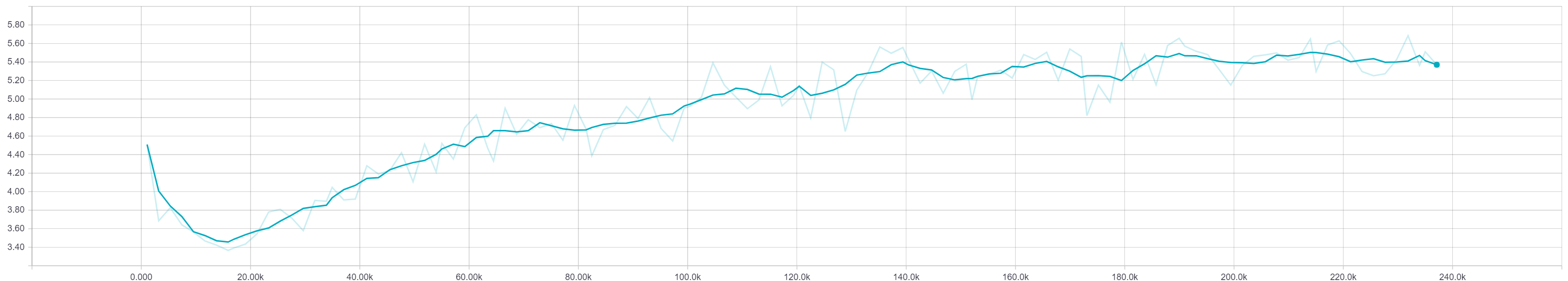}
	\caption{The diagram shows the validation loss over the entire training of the Cornell S model. The vertical axis represents the loss value and the horizontal axis represents the number of training steps.}
	\label{CornellSOverfit}
\end{figure}

\begin{figure}[H] 
	\centering
	\includegraphics[width=1.0\textwidth]{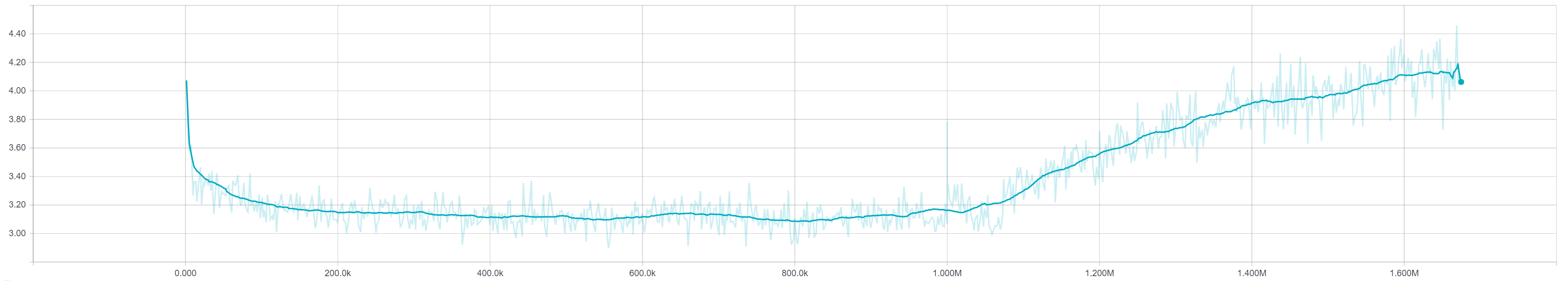}
	\caption{The diagram shows the validation loss over the entire training of the OpenSubtitles and the OpenSubtitles F model. In the first 1M steps the model was trained on OpenSubtitles data, then it was switched to speaker-annotated Cornell Movie-Dialog data. The vertical axis represents the loss value and the horizontal axis represents the number of training steps.}
	\label{Finetuned_loss}
\end{figure}

\begin{table}[H]
	\centering
	\begin{tabular}{|l|l|l|l|l|l|}
		\hline
		\textbf{Metrics} & \textbf{S2S} & \textbf{Cornell} & \textbf{Cornell S} & \textbf{OpenSubtitles} & \textbf{OpenSubtitles F}\\ \hline
		\textbf{Perplexity} & 17 & 17 & 15.4 & 11.7 & 24.6
		\\ \hline
		\textbf{Bleu} & - & 4.7 & 4.2 & 6.8 & 4.6
		\\ \hline
	\end{tabular}
	\caption{Perplexity and Bleu scores on validation data. S2S stands for the baseline seq2seq model, Cornell stands for the training setup presented in Section~\ref{sssec:cornell_training}, Cornell S stands for Section~\ref{sssec:cornell_speakers}, OpenSubtitles stands for Section~\ref{sssec:opensubs_training} and OpenSubtitles F stands for Section~\ref{sssec:finetune_training}.}	
	\label{table:scores}
\end{table}
In Table~\ref{table:scores} the perplexity and Bleu scores of the trained models can be seen on the respective validation datasets. Since the models trained on Cornell Movie-Dialog data overfitted quite quickly the dataset as can be seen in Figure~\ref{CornellOverfit} and Figure~\ref{CornellSOverfit}, the perplexity and Bleu scores presented are computed on the validation data before overfitting occurred. This means that for the Cornell model the scores were computed after 20K training steps and for the Cornell S model, scores were computed after 16K steps. For the OpenSubtitles training the model didn't seem to overfit, however after a quick drop the loss stagnated for the remainder of the training, which could be because the learning rate was not adequate or the model was too small to be able to learn more. Future work should address these issues, described in detail in Section~\ref{sec:future}. The effects can be seen in Figure~\ref{Finetuned_loss} up to step 1M, since afterwards the training was switched to the Cornell Movie-Dialog dataset. The scores corresponding to the OpenSubtitles F training are very poor, since the model immediately started to overfit after being switched from OpenSubtitles to Cornell Movie-Dialog data starting from step 1M in Figure~\ref{Finetuned_loss}.


\newpage\subsection{Qualitative Analysis} \label{ssec:52}
In this section output response samples from the various trainings are presented. For the sake of comparison with the baseline seq2seq model, the source utterances are a subset of the ones used in \cite{Vinyals:2015}. The source utterances were divided into several categories like in \cite{Vinyals:2015}. Generated responses by the trained models to the various source utterances can be seen from Table~\ref{table:basicNCM} to Table~\ref{table:philosophicalNCM}. Interestingly the best responses from the two Cornell trainings were not those outputted before the models started to overfit. Before overfitting the models usually outputted very generic responses similar to the ones seen in the OpenSubtitles column of the tables. By letting the models overfit a little on the training data more diverse and interesting responses were generated. This further suggests that the loss function is not in direct correlation with the quality of responses, since it gives a better score to safe and generic answers, like \textit{I don't know}. Specifically, the examples presented for the Cornell model are after training for 219K steps and for the Cornell S model after training for 238K steps. Also, for the OpenSubtitles F model, generated output responses after 370K steps of finetuning were used, since there was no further improvement observed in the quality of responses as training continued.

As can be seen in the tables, the OpenSubtitles F model, which was finetuned on Cornell Movie-Dialog data generates better responses than the OpenSubtitles variant, which proves that the finetuning had a beneficial effect on the quality of generated responses. This is again in contrast with the loss and metrics values, since these are much better for the OpenSubtitles model. Generally however the models trained solely on the Cornell Movie-Dialog dataset generate the most diverse and interesting responses. Despite this, none of the trainings achieved the performance of the baseline seq2seq model. Perhaps the most striking result is the OpenSubtitles model, since this provides the most direct comparison with the baseline seq2seq model, because similar and similarly sized datasets were used for training the two models. Several reasons are given for these results in Section~\ref{sec:future}, as well as possible solutions and future directions. It is important to note however that the source utterances were originally selected for evaluating the seq2seq model and they don't provide a complete qualitative picture of the performance of the conversational models, since for other source utterances the Transformer model based trainings might generate better results. In order to truly compare the trained models a set of random source utterances should be constructed and human evaluators employed to rank the generated responses.

\begin{table}[H]
	\centering
	\begin{tabular}{|L{2.3cm}|L{2.3cm}|L{2.3cm}|L{2.3cm}|L{2.3cm}|L{2.3cm}|}
		\hline
		\textbf{Source Utterance} & \textbf{S2S} & \textbf{Cornell} & \textbf{Cornell S} & \textbf{OpenSubtitles} & \textbf{OpenSubtitles F}\\ \hline
		hello ! & hello ! & hello . & hi . & hello . & hi .
		\\ \hline
		how are you ? & i 'm good . & fine . & fine . how are you ? & how are you ? & fine .
		\\ \hline
		what 's your name ? & i ’m julia . & my name is sir robin of camelot !  & david . & \textless UNK\textgreater . & bob . 
		\\ \hline
		when were you born ? & july 20th . & what ? & yesterday . & two years ago . & just a few minutes ago .
		\\ \hline
		what year were you born ? & 1977 . & eight years . . . & my father was the paper . & i don 't know . & \textless UNK\textgreater .
		\\ \hline
		where are you from ? & i ’m out in the boonies. & south america . & new york . & where are you ? & in here .
		\\ \hline
		are you a man or a woman ? & i ’m a woman . & i 'm a teacher .  & i never forget a pussy . . . cat . & yes . & yes .
		\\ \hline
		why are we here ? & i ’m not sure . & we ca n't stay here .  & the list . & i don 't know . & we 've got to get out of here . 
		\\ \hline
		okay bye ! & bye . & bye . & bye ! & bye . & bye .
		\\ \hline
		see you later . & bye . & where you off to ?  & bye ! & bye . & bye .
		\\ \hline
		
	\end{tabular}
	\caption{A set of general source utterances and the generated responses by the trained models.}	
	\label{table:basicNCM}
\end{table}
\begin{table}[H]
	\centering
	\begin{tabular}{|L{2.3cm}|L{2.3cm}|L{2.3cm}|L{2.3cm}|L{2.3cm}|L{2.3cm}|}
		\hline
		\textbf{Source Utterance} & \textbf{S2S} & \textbf{Cornell} & \textbf{Cornell S} & \textbf{OpenSubtitles} & \textbf{OpenSubtitles F}\\ \hline
		what do you like to talk about ? & nothing . & about things . about your wife .  & about things that are important to me . . . & i don 't know . & i do \textless UNK\textgreater know .
		\\ \hline
		what is your job ? & i 'm a lawyer . & my job my rules .  & i need a vacation . & i don 't know . & i . . . i . . .
		\\ \hline
		what do you do ? & i 'm a doctor . & i 'm a comedian .   & i 'm taking a piss . & i don 't know . & i 'm sorry .
		\\ \hline
		
	\end{tabular}
	\caption{Job and Personality.}	
	\label{table:personalityNCM}
\end{table}

\begin{table}[H]
	\centering
	\begin{tabular}{|L{2.3cm}|L{2.3cm}|L{2.3cm}|L{2.3cm}|L{2.3cm}|L{2.3cm}|}
		\hline
		\textbf{Source Utterance} & \textbf{S2S} & \textbf{Cornell} & \textbf{Cornell S} & \textbf{OpenSubtitles} & \textbf{OpenSubtitles F}\\ \hline
		who is skywalker ? & he is a hero . & a snake ! & ben . & i don 't know . & my brother . 
		\\ \hline
		who is bill clinton ? & he 's a billionaire . & name your brother . & i do n't know . & i don 't know . & he 's my brother . 
		\\ \hline
		is sky blue or black ? & blue . & blue eyes . & a car accident . & yes . & yes .
		\\ \hline
		does a cat have a tail ? & yes . & does it matter ?  & maybe . & no . & yes .
		\\ \hline
		does a cat have a wing ? & no & it 's a \textless UNK\textgreater . & not for lack of trying . & yes . & yes .
		\\ \hline
		can a cat fly ? & no . & you 're a woman are n't you ? & or a little bit . & no . & sure . 
		\\ \hline
		how many legs does a cat have ? & four , i think . & one .  & four . & i don 't know . & four .
		\\ \hline
		how many legs does a spider have ? & three , i think . & two . & thirty . & i don 't know . & four . 
		\\ \hline
		what is the color of the sky ? & blue . & blue !  & green . & i don 't know . & blue .
		\\ \hline
		what is the color of water ? & water . & what ? & white . & i don 't know . & blue .
		\\ \hline
		what is the color of blood ? & it is the same as a black eye . & it is the rabbit . & beautiful . & i don 't know . & blue .
		\\ \hline
		how much is two plus two ? & four . & i do n't know . & well do n't worry about it . & . & ten . 
		\\ \hline
		how much is ten minus two ? & seventy - two . & between ten and twelve .  & yeah . & . & ten . 
		\\ \hline
		
	\end{tabular}
	\caption{General Knowledge Q\&A.}	
	\label{table:genKnowledgeNCM}
\end{table}
\begin{table}[H]
	\centering
	\begin{tabular}{|L{2.3cm}|L{2.3cm}|L{2.3cm}|L{2.3cm}|L{2.3cm}|L{2.3cm}|}
		\hline
		\textbf{Source Utterance} & \textbf{S2S} & \textbf{Cornell} & \textbf{Cornell S} & \textbf{OpenSubtitles} & \textbf{OpenSubtitles F}\\ \hline
		my name is david . what is my name ? & david . & victor . & \textless UNK\textgreater . & what 's your name ? & bob .
		\\ \hline
		my name is john . what is my name ? & john . & i want to know . & what ? & \textless UNK\textgreater . & bob .
		\\ \hline
		
	\end{tabular}
	\caption{How different models handle context.}	
	\label{table:contextNCM}
\end{table}
\begin{table}[H]
	\centering
	\begin{tabular}{|L{2.3cm}|L{2.3cm}|L{2.3cm}|L{2.3cm}|L{2.3cm}|L{2.3cm}|}
		\hline
		\textbf{Source Utterance} & \textbf{S2S} & \textbf{Cornell} & \textbf{Cornell S} & \textbf{OpenSubtitles} & \textbf{OpenSubtitlesF}\\ \hline
		what is the purpose of life ? & to serve the greater good . & i do n't know . & my father . & i don 't know . & i do \textless UNK\textgreater know .
		\\ \hline
		what is the purpose of living ? & to live forever . & i do n't know . & a lot of people . & i don 't know . & i do \textless UNK\textgreater know .
		\\ \hline
		what is the purpose of existence ? & to find out what happens when we get to the planet earth . & i do n't know . & i do n't know . could be a lot of lives . & i don 't know . & i do \textless UNK\textgreater know .
		\\ \hline
		where are you now ? & i ’m in the middle of nowhere . & here . & here . & where are you ? & in here .
		\\ \hline
		what is the purpose of dying ? & to have a life . & i do n't know . & we 're not going somewhere . & i don 't know . & i do \textless UNK\textgreater know .
		\\ \hline
		what is the purpose of being intelligent ? & to find out what it is . & intelligent beyond our comprehension .  & we 're not sure . & i don 't know . & i do \textless UNK\textgreater know .
		\\ \hline
		what is the purpose of emotions ? & i don ’t know . & i do n't know sir .  & you know a little of this a little of that . & i don 't know . & i do \textless UNK\textgreater know .
		\\ \hline
		
	\end{tabular}
	\caption{Philosophical Q\&A.}	
	\label{table:philosophicalNCM}
\end{table}

In Table~\ref{table:CornellS1} generated output responses from the Cornell S training (after 145K steps) to various source utterances can be seen. The text of the source utterances is the same for all columns, however the speaker and addressee names are changed. More specifically for the \textit{Cornell S} column no names were given. For the columns \textit{Cornell S matching} and \textit{Cornell S matching R} the characters Ben and Mrs. Robinson were used from the movie \textit{The Graduate}. These characters were selected because they have many interactions in the training data, thus the model must have learned something about them. The difference between the two columns is that in the \textit{Cornell S matching} column Mrs. Robinson addresses Ben and in the \textit{Cornell S matching R} column Ben addresses Mrs. Robinson. For the final two columns the character of Ben is kept, however the other speaker embedding used is that of Joe from the movie \textit{Innerspace}. In the \textit{Cornell S different} column Ben addresses Joe and in the \textit{Cornell S different R} column Joe addresses Ben. These two characters were selected, because they have many utterances in the training data, however since they aren't in the same movie, they don't interact with each other.

There is a clear difference between the generated responses, solely in consequence of the speakers and addressees being different. The most striking influence can be seen where the source utterance asks the name of someone. For these questions the model learns to respond with actual corresponding names, however they are sometimes reversed. In the second row, when Mrs. Robinson asks \textit{what's your name?} from Ben, the model responds with Mrs. Robinson and when Ben asks the question from Mrs. Robinson the model responds with Benjamin. In row 8, if the first sentence is ignored, the responses to the question \textit{what is my name?} are the same as for the previous question, but since this question refers to the speaker, the replies are actually correct. It is important to note here that the tokens used for speakers and addressees are completely different from any names present in the dataset. Thus the model learned to come up with a sequence of 3 tokens (\textit{mrs . robinson}) as answer, while the actual token used to represent Mrs. Robinson is \textit{MRS.\_ROBINSON\_m77}.

Even when the characters did not interact in the training dataset, meaningful responses are generated. For example, in row 6 and column 5, when Ben says \textit{see you later.} to Joe, the model responds with \textit{benjamin ?}, which makes sense. Interestingly this was the only speaker-addressee combination where the model managed to answer with numbers to the utterances in the last two rows of the table. Moreover, in rows 8 and 9 for the question \textit{what is my name?}, when Joe asks Ben, the model responds with \textit{mrs . robinson}, which makes sense since the two characters haven't interacted before, thus if Ben is addressed it responds with the character name that it had the most interactions with.

The OpenSubtitles F training, which was also speaker annotated didn't produce such distinct and interesting answers, so the comparison of different source utterances is not presented here. This is probably due to the model being pretrained on the OpenSubtitles dataset. Since this dataset is much larger the model didn't manage to adapt to the smaller Cornell Movie-Dialog dataset, and thus its responses remained somewhat general and safe.

\begin{table}[H]
	\centering
	\begin{tabular}{|L{2.3cm}|L{2.3cm}|L{2.3cm}|L{2.3cm}|L{2.3cm}|L{2.3cm}|}
		\hline
		\textbf{Source Utterance} & \textbf{Cornell S} & \textbf{Cornell S matching} & \textbf{Cornell S matching R} & \textbf{Cornell S different} & \textbf{Cornell S different R}\\ \hline
		how are you ? & fine . how are you ? & very well . thank you . & i 'm in touch . & i ca n't how are you ? & i 'm fine . what 's up ?
		\\ \hline
		what 's your name ? & david . & mrs . robinson . & benjamin . & do n't you know ? & why do you do that ?
		\\ \hline
		when were you born ? & yesterday . & very interesting . & about three years . & about a year ago . & tuesday .
		\\ \hline
		where are you from ? & new york . & i 'm at the airport . & at the airport . & off the street . & with you .
		\\ \hline
		why are we here ? & the list . & well i just want you to leave . & i have to see you . & i just wanted to see you . & just to see you .
		\\ \hline
		see you later . & bye ! & now wait a minute . & benjamin ? & benjamin ? & okay .
		\\ \hline
		my name is david . what is my name ? & \textless UNK\textgreater . & sebastian . is that you ? & benjamin . & umyu name . & mrs . robinson .
		\\ \hline
		my name is john . what is my name ? & what ? & mrs . robinson . & benjamin . & my name 's john . & mrs . robinson .
		\\ \hline
		who is bill clinton ? & i do n't know . & no one knows . it 's his brother . & you know her . & well it 's not just me . & just some friends .
		\\ \hline
		is sky blue or black ? & a car accident . & oh it 's a lovely name . & do you know what it is ? & blue . & blue .
		\\ \hline
		does a cat have a tail ? & maybe . & yeah . i suppose so . & yes . & it does n't seem . & nope .
		\\ \hline
		how many legs does a cat have ? & four . & ten . & i do n't know . & not too big . & i lived here .
		\\ \hline
		what is the color of water ? & white . & oh it 's down . & i do n't really know where it is . & blue . & it 's like a song .
		\\ \hline
		how much is two plus two ? & well do n't worry about it . & what is it ? & i do n't know . & eight . & oh i do n't know .
		\\ \hline
		how much is ten minus two ? & yeah . & oh that . & not very . & twenty . & just about .
		\\ \hline
		
	\end{tabular}
	\caption{Cornell S output responses for various name combinations.}	
	\label{table:CornellS1}
\end{table}

\newpage\section{Future Work} \label{sec:future}
From the results presented in Section~\ref{ssec:52} the conclusion that the Transformer model simply doesn't perform as well as the seq2seq model could be drawn, however there are several reasons for the qualitatively worse results of the Transformer model trained with dialog data. First of all, the parameter space of the seq2seq model trained by \cite{Vinyals:2015} is much bigger than the base variant of the Transformer that was used in this work. Secondly, as it has been discussed in Section~\ref{ssec:51}, the Transformer trained on OpenSubtitles data seemed to stagnate which could be due to a too high learning rate or other hyperparameters not set correctly. Because of this it didn't achieve the point of convergence on the dataset and produced a lot of generic and safe responses. Since the same Transformer model and same hyperparameters were used as in the original work, which was tuned for NMT, it might very well be the case that the model would perform better with other hyperparameter configurations. Hyperparameter tuning is a general issue that subsequent research should focus on. Trying out larger Transformer variants for the OpenSubtitles dataset is also an important future direction.

Furthermore, it might be the case that the current standard loss function used has an even worse effect if used with the Transformer model. As discussed in Section~\ref{sssec:loss_function}, there are fundamental problems with this loss function and this is further proved in Section~\ref{sec:results}, where models that overfit the dataset and thus their validation loss increases, generate better and more diverse responses. Because of wrong hyperparameter settings or in consequence of the Transformer model not having a large enough parameter space, it couldn't even come close to overfitting the larger OpenSubtitles dataset, which is a good explanation of the safe and generic responses produced.

In addition to the ideas presented above to make the transformer model work better for conversational modeling, several ideas are proposed in the following sections towards solving some of the issues presented in Section~\ref{ssec:problems}. In Section~\ref{ssec:61} the loss function issue is tackled. Then, a conversational model that takes into account the passage of time is described in Section~\ref{ssec:62}. Finally, further ideas related to constructing better conversational models are presented in Section~\ref{ssec:63}.

\subsection{Ideas Towards Solving The Loss Function Issue} \label{ssec:61}
Since human-like conversation is basically artificial general intelligence (AGI), it is extremely difficult to tackle it up front. A complex model would have to be constructed and a lot of knowledge included about the world in meaningful ways. This section will instead focus on augmenting encoder-decoder models with techniques that could remedy the loss function issue presented in Section~\ref{sssec:loss_function}. I propose that a multitude of features and priors should be taken into consideration when modeling conversations. Since maximizing the log probability is essentially the same as maximizing the probability of a reply given an utterance, there should be other prior probabilities that the reply can be conditioned on. This would address the issue of having various different responses for one source utterance, since differentiation between them by using other priors would become possible. A number of similar augmentations have been proposed before \cite{Li:2016,Xing_topic:2017,Zhou:2017,Choudhary:2017}. They all try to feed additional information into seq2seq models, like personality, mood and topic categories. While they all show that generated responses from these models are somewhat more diverse than those outputted by a basic seq2seq model, I still believe that the conversational models are somewhat ambiguous and more priors should be used.

While at the birth of the seq2seq architecture conversational models were trained on utterance-reply pairs or by concatenating previous turns into one source sequence, there have been many efficient techniques proposed that can take into account conversation history, which are discussed in Section~\ref{sssec:context} and Section~\ref{sssec:HRED}. Conversation history is indeed one of the most important priors that should be taken into account, since a response to an utterance is usually grounded on information presented in previous turns. However, this problem is mostly solved by the aforementioned techniques.

I propose several priors on which a conversational model should be conditioned. More specifically, the speaker and addressee of each utterance and the mood of the speaker at the time of saying the utterance. Taking into account the persona of the speaker and addressee is important, since what one says in general depends on one's personality and past experiences and also on who one talks to. This is already a good starting point, because it differentiates between different answers to same questions. However, even for a single person the reply to the question \textit{How are you?} might be completely different depending on whether the person is happy, sad, annoyed, etc. By taking into account this mood prior, the data that the conversational model is trained on can be further disambiguated. Ideally, a dataset could be created where for each source sequence and set of priors only one correct output sequence exists. This would assure that the generated replies are diverse and not just an average of all possible responses. All of these priors and features can be represented by embeddings similar to word embeddings, which can be learned during training. However, as more features are added, more data is needed, so that model has sufficient examples from which it can learn what it \textit{means} to be in a mood or talk like a specific person. This is a problem with the Cornell Movie-Dialog dataset, which is annotated with speaker-addressees, but it is too small and the model can't generalize from so few examples. 

There are still problems with the above mentioned approach. Intuitively the speaker and addressee representations should have a lot of parameters, since they have to capture what makes one different from other people, which depends on all of one's past experiences. This is a lot of information to capture into a vector, for example to represent a single word 100-1000 dimensional vectors are used and since a lot of information (many words) are needed to describe a person, it would make the representation huge. There are two problems with these huge representations. The network would be very difficult to train because of limited computational resources and current datasets don't offer enough examples from which the network can learn such a huge space of parameters.

In order to combat these issues I propose a different approach. The priors previously mentioned are still used, however personality representations should be kept to learnable sizes. The intuition behind this is that people generally don't differ that much in basic world knowledge and conversational style. Thus, these representations could be kept smaller and instead the construction of a representation which captures general world-knowledge, language-knowledge and conversation-knowledge should be pursued. An example for world-knowledge would be the color of the sky, language-knowledge is about learning the meaning of words and conversation-knowledge is about learning to answer with yes/no to yes-or-no questions for example. The advantage of this representation is that it could also be trained on non-conversational data with unsupervised methods.

There is one final prior that needs to be taken into account to truly capture the conditioning space of conversations. Take for example the reply \textit{I got hit by a car.} to the question \textit{How are you?}. Obviously, this response has little to do with the persona of the speaker or any general knowledge about the world. Rather the response is simply conditioned on outside factors, which can be temporary. Temporary, because after a week has passed from the accident in the previous scenario, the person will probably not reply with this answer, since \textit{How are you?} is a question about the present. There are some ways in which these outside factors could be dealt with. For example the person's speaker representation could be slightly changed for this specific response, or a new representation that tries to encode outside factors could be used. The most simple approach would be to just cut out conversations grounded on outside factors from the dataset. Unfortunately, all of these methods require a dataset that is labeled with labels that show whether the response is based on some external factor, which can probably only be done manually and to my knowledge such a dataset does not exist currently.

The actual specifics of how such representations could be integrated with encoder-decoder models can have many variations of course, however this work does not go into details.

\subsection{Temporal Conditioning and Memory} \label{ssec:62}
An important part of conversations is timing and the involvement of both speakers. Current chatbot models are trained in such ways that they will only emit one response instantly after they receive the user's utterance. To make chatbots more human-like I propose an additional term to the loss function, which is based on the temporal delay between an utterance and a reply. Essentially the chatbot has to generate a reply and also guess how much time should it wait before emitting the reply. Twitter-style datasets usually have such annotations so the implementation of this feature should be straight-forward. 

The time-delay can be represented as a vector of probabilities over time frame categories. For example the model could have 10\% confidence that the reply should be delayed by 0 to 10 seconds and 90\% confidence that it should be delayed by 10 to 20 seconds. Such a representation is effective because a simple softmax can be used to get the probabilities of time frame categories and backpropagation can be extended by comparing the predicted time-delay vector with the one-hot ground-truth timing vector. Not only would this addition help achieve a more human-like delay (which could also be tunable) in the chatbot's responses, but it would allow a conversational model to periodically generate new utterances by itself, based on the conversation history. Regardless of whether the user has inputted an utterance or not, the conversation history can be fed into the model after the chatbot emitted an utterance, and a new utterance can be generated if the model thinks that it is necessary to further the conversation. With the time delay mechanism, this process can go on indefinitely, since the generation of the new utterance will only be considered after the time delay for the current utterance has passed. This feature would make the chatbot even more human-like and engaging, by furthering the conversation without user interaction.

Another concept tied to temporal conditioning is real-time model updates. Basically, at each turn the user's utterances can be backpropagated through the network, based on the conversation history. This would make the model update its weights and prior representations as the conversation goes on, thus remembering the dialog. The technique could either replace or complement the need for taking into account long conversation histories with hierarchical models. It would be especially useful for real-world deployment of chatbots, where users expect a chatbot to be able to remember what they said 1000 utterances ago, which can't be achieved solely through a hierarchical model. Through these real time updates of the parameters the model is capable to encode and learn new information about the user, which acts as memory. For example for a new user the persona representation is not known yet, however by interacting with the user the model can learn information specific to that user and it can encode it into the representation with real-time backpropagation.

\subsection{Additional Ideas} \label{ssec:63}
In this final section several further ideas are presented that could be used to build better conversational models. Reinforcement learning has achieved impressive results recently for game-like tasks where there is a clear end-goal, without any human supervision \cite{alphagozero}. Some approaches to adapt RL to the field of conversational modeling have been discussed in Section~\ref{sssec:RL}. My idea would be to train 2 encoder-decoder based conversational agents and then let them talk to each other. Furthermore, one of the chatbots would have the goal of getting a specific response from the other bot. If in some number of turns the response is uttered, then both bots could get a positive reward. In this environment several goals can be designed that would help further train the chatbots without any data or supervision. The downside of this idea is that hand-crafted goals have to be designed, which should capture important conversational properties.

New neural architectures should also be tried for the conversational domain, for example models that were described in Section~\ref{sssec:diffencdec}. Furthermore, an interesting technique called neural architecture search \cite{Zoph:2016} has been proposed to build new types of recurrent cells and even entire convolutional architectures. Since there already exist a plethora of different computational blocks used in conversational models, these could simply be input to the neural architecture search model, which could then generate an architecture that would presumably be better suited for the task of conversational modeling.

\newpage\section{Conclusion} \label{sec:conclusion}

A brief overview of the history of chatbots has been given and the encoder-decoder model has been described in detail. Afterwards an in-depth survey of scientific literature related to conversational models, published in the last 3 years, was presented. Various techniques and architectures were discussed, that were proposed to augment the encoder-decoder model and to make conversational agents more natural and human-like. Criticism was also presented regarding some of the properties of current chatbot models and it has been shown how and why several of the techniques currently employed are inappropriate for the task of modeling conversations. Furthermore, preliminary experiments were run by training the Transformer model on two different dialog datasets. The performance of the trainings was analyzed with the help of automatic evaluation metrics and by comparing output responses for a set of source utterances. Finally, it was concluded that further, more detailed experiments are needed in order to determine whether the Transformer model is truly worse than the standard RNN based seq2seq model for the task of conversational modeling. In addition to presenting directions and experiments that should be conducted with the Transformer model, several ideas were presented related to solving some of the issues brought up in the criticism that was given earlier in the paper. These ideas are an important direction for future research in the domain of conversational agents, since they are not model related, but rather try to solve fundamental issues with current dialog agents. Continuation of this work will focus on trying to make open-domain conversational models as human-like as possible by implementing the ideas presented.

\bibliographystyle{apalike}
\newpage\bibliography{chatbots}

\begin{thebibliography}{}

\bibitem[Akasaki and Kaji, 2017]{Akasaki:2017}
Akasaki, S. and Kaji, N. (2017).
\newblock Chat detection in an intelligent assistant: Combining task-oriented
  and non-task-oriented spoken dialogue systems.
\newblock {\em arXiv preprint arXiv:1705.00746}.

\bibitem[Apple, 2017]{Siri:2017}
Apple (2017).
\newblock Siri.
\newblock \url{https://www.apple.com/ios/siri/}.
\newblock Accessed: 2017-10-04.

\bibitem[Ba et~al., 2016]{Ba:2016}
Ba, J.~L., Kiros, J.~R., and Hinton, G.~E. (2016).
\newblock Layer normalization.
\newblock {\em arXiv preprint arXiv:1607.06450}.

\bibitem[Bahdanau et~al., 2014]{Bahdanau:2014}
Bahdanau, D., Cho, K., and Bengio, Y. (2014).
\newblock Neural machine translation by jointly learning to align and
  translate.
\newblock {\em arXiv preprint arXiv:1409.0473}.

\bibitem[Barone and Sennrich, 2017]{Rico:2017}
Barone, A. V.~M. and Sennrich, R. (2017).
\newblock A parallel corpus of python functions and documentation strings for
  automated code documentation and code generation.
\newblock {\em arXiv preprint arXiv:1707.02275}.

\bibitem[Bengio et~al., 2003]{Bengio:2003}
Bengio, Y., Ducharme, R., Vincent, P., and Jauvin, C. (2003).
\newblock A neural probabilistic language model.
\newblock {\em Journal of machine learning research}, 3(Feb):1137--1155.

\bibitem[Bordes et~al., 2016]{Bordes:2016}
Bordes, A., Boureau, Y.-L., and Weston, J. (2016).
\newblock Learning end-to-end goal-oriented dialog.
\newblock {\em arXiv preprint arXiv:1605.07683}.

\bibitem[Bottou, 2010]{SGD:2010}
Bottou, L. (2010).
\newblock Large-scale machine learning with stochastic gradient descent.
\newblock In {\em Proceedings of COMPSTAT'2010}, pages 177--186. Springer.

\bibitem[Britz, 2015]{RNN_pic:2017}
Britz, D. (2015).
\newblock Recurrent neural network tutorial.
\newblock
  \url{http://www.wildml.com/2015/09/recurrent-neural-networks-tutorial-part-1-introduction-to-rnns/}.
\newblock Accessed: 2017-10-09.

\bibitem[Carpenter, 2017]{Cleverbot:2017}
Carpenter, R. (2017).
\newblock Cleverbot.
\newblock \url{http://www.cleverbot.com/}.
\newblock Accessed: 2017-10-04.

\bibitem[Chen and Manning, 2014]{Chen:2014}
Chen, D. and Manning, C. (2014).
\newblock A fast and accurate dependency parser using neural networks.
\newblock In {\em Proceedings of the 2014 conference on empirical methods in
  natural language processing (EMNLP)}, pages 740--750.

\bibitem[Cheng et~al., 2016]{Cheng:2016}
Cheng, J., Dong, L., and Lapata, M. (2016).
\newblock Long short-term memory-networks for machine reading.
\newblock {\em arXiv preprint arXiv:1601.06733}.

\bibitem[Chiu et~al., 2017]{Chiu:2017}
Chiu, C.-C., Lawson, D., Luo, Y., Tucker, G., Swersky, K., Sutskever, I., and
  Jaitly, N. (2017).
\newblock An online sequence-to-sequence model for noisy speech recognition.
\newblock {\em arXiv preprint arXiv:1706.06428}.

\bibitem[Cho et~al., 2014]{Cho:2014}
Cho, K., Van~Merri{\"e}nboer, B., Gulcehre, C., Bahdanau, D., Bougares, F.,
  Schwenk, H., and Bengio, Y. (2014).
\newblock Learning phrase representations using rnn encoder-decoder for
  statistical machine translation.
\newblock {\em arXiv preprint arXiv:1406.1078}.

\bibitem[Choudhary et~al., 2017]{Choudhary:2017}
Choudhary, S., Srivastava, P., Ungar, L., and Sedoc, J. (2017).
\newblock Domain aware neural dialog system.
\newblock {\em arXiv preprint arXiv:1708.00897}.

\bibitem[Danescu-Niculescu-Mizil and Lee, 2011]{Danescu:2011}
Danescu-Niculescu-Mizil, C. and Lee, L. (2011).
\newblock Chameleons in imagined conversations: A new approach to understanding
  coordination of linguistic style in dialogs.
\newblock In {\em Proceedings of the 2nd Workshop on Cognitive Modeling and
  Computational Linguistics}, pages 76--87. Association for Computational
  Linguistics.

\bibitem[Das et~al., 2017]{Das:2017}
Das, A., Kottur, S., Moura, J.~M., Lee, S., and Batra, D. (2017).
\newblock Learning cooperative visual dialog agents with deep reinforcement
  learning.
\newblock {\em arXiv preprint arXiv:1703.06585}.

\bibitem[Dodge et~al., 2015]{Dodge:2015}
Dodge, J., Gane, A., Zhang, X., Bordes, A., Chopra, S., Miller, A., Szlam, A.,
  and Weston, J. (2015).
\newblock Evaluating prerequisite qualities for learning end-to-end dialog
  systems.
\newblock {\em arXiv preprint arXiv:1511.06931}.

\bibitem[Eric and Manning, 2017]{Eric:2017}
Eric, M. and Manning, C.~D. (2017).
\newblock A copy-augmented sequence-to-sequence architecture gives good
  performance on task-oriented dialogue.
\newblock {\em arXiv preprint arXiv:1701.04024}.

\bibitem[Facebook, 2017]{bAbi}
Facebook (2017).
\newblock The babi project.
\newblock \url{https://research.fb.com/downloads/babi/}.
\newblock Accessed: 2017-10-13.

\bibitem[Feng et~al., 2017]{Feng:2017}
Feng, Y., Zhang, S., Zhang, A., Wang, D., and Abel, A. (2017).
\newblock Memory-augmented neural machine translation.
\newblock {\em arXiv preprint arXiv:1708.02005}.

\bibitem[Gehring et~al., 2017]{ConvS2S:2017}
Gehring, J., Auli, M., Grangier, D., Yarats, D., and Dauphin, Y.~N. (2017).
\newblock Convolutional sequence to sequence learning.
\newblock {\em arXiv preprint arXiv:1705.03122}.

\bibitem[Ghazvininejad et~al., 2017]{Ghazvininejad:2017}
Ghazvininejad, M., Brockett, C., Chang, M.-W., Dolan, B., Gao, J., Yih, W.-t.,
  and Galley, M. (2017).
\newblock A knowledge-grounded neural conversation model.
\newblock {\em arXiv preprint arXiv:1702.01932}.

\bibitem[Goodfellow et~al., 2014]{Goodfellow:2014}
Goodfellow, I., Pouget-Abadie, J., Mirza, M., Xu, B., Warde-Farley, D., Ozair,
  S., Courville, A., and Bengio, Y. (2014).
\newblock Generative adversarial nets.
\newblock In {\em Advances in neural information processing systems}, pages
  2672--2680.

\bibitem[Google, 2017]{Google:2017}
Google (2017).
\newblock Google assistant.
\newblock \url{https://assistant.google.com/}.
\newblock Accessed: 2017-10-04.

\bibitem[Goyal et~al., 2017]{Goyal:2017}
Goyal, K., Neubig, G., Dyer, C., and Berg-Kirkpatrick, T. (2017).
\newblock A continuous relaxation of beam search for end-to-end training of
  neural sequence models.
\newblock {\em arXiv preprint arXiv:1708.00111}.

\bibitem[Gu et~al., 2016]{Gu:2016}
Gu, J., Lu, Z., Li, H., and Li, V.~O. (2016).
\newblock Incorporating copying mechanism in sequence-to-sequence learning.
\newblock {\em arXiv preprint arXiv:1603.06393}.

\bibitem[Harris, 1954]{Harris:1954}
Harris, Z.~S. (1954).
\newblock Distributional structure.
\newblock {\em Word}, 10(2-3):146--162.

\bibitem[Havrylov and Titov, 2017]{Havrylov:2017}
Havrylov, S. and Titov, I. (2017).
\newblock Emergence of language with multi-agent games: Learning to communicate
  with sequences of symbols.
\newblock {\em arXiv preprint arXiv:1705.11192}.

\bibitem[He et~al., 2016]{He:2016}
He, K., Zhang, X., Ren, S., and Sun, J. (2016).
\newblock Deep residual learning for image recognition.
\newblock In {\em Proceedings of the IEEE conference on computer vision and
  pattern recognition}, pages 770--778.

\bibitem[Henderson, 2015]{Henderson:2015}
Henderson, M. (2015).
\newblock Machine learning for dialog state tracking: A review.
\newblock In {\em Machine Learning in Spoken Language Processing Workshop}.

\bibitem[Hochreiter, 1998]{Hochreiter:1998}
Hochreiter, S. (1998).
\newblock The vanishing gradient problem during learning recurrent neural nets
  and problem solutions.
\newblock {\em International Journal of Uncertainty, Fuzziness and
  Knowledge-Based Systems}, 6(02):107--116.

\bibitem[Hochreiter and Schmidhuber, 1997]{Hochreiter:1997}
Hochreiter, S. and Schmidhuber, J. (1997).
\newblock Long short-term memory.
\newblock {\em Neural computation}, 9(8):1735--1780.

\bibitem[Jean et~al., 2014]{Jean:2014}
Jean, S., Cho, K., Memisevic, R., and Bengio, Y. (2014).
\newblock On using very large target vocabulary for neural machine translation.
\newblock {\em arXiv preprint arXiv:1412.2007}.

\bibitem[Jena et~al., 2017]{Jena:2017}
Jena, G., Vashisht, M., Basu, A., Ungar, L., and Sedoc, J. (2017).
\newblock Enterprise to computer: Star trek chatbot.
\newblock {\em arXiv preprint arXiv:1708.00818}.

\bibitem[Joshi et~al., 2017]{Joshi:2017}
Joshi, C.~K., Mi, F., and Faltings, B. (2017).
\newblock Personalization in goal-oriented dialog.
\newblock {\em arXiv preprint arXiv:1706.07503}.

\bibitem[Kaiser and Bengio, 2016]{Kaiser:2016}
Kaiser, {\L}. and Bengio, S. (2016).
\newblock Can active memory replace attention?
\newblock In {\em Advances in Neural Information Processing Systems}, pages
  3781--3789.

\bibitem[Kaiser et~al., 2017a]{Kaiser:2017}
Kaiser, L., Gomez, A.~N., and Chollet, F. (2017a).
\newblock Depthwise separable convolutions for neural machine translation.
\newblock {\em arXiv preprint arXiv:1706.03059}.

\bibitem[Kaiser et~al., 2017b]{Kaiser_one_model:2017}
Kaiser, L., Gomez, A.~N., Shazeer, N., Vaswani, A., Parmar, N., Jones, L., and
  Uszkoreit, J. (2017b).
\newblock One model to learn them all.
\newblock {\em arXiv preprint arXiv:1706.05137}.

\bibitem[Kalchbrenner et~al., 2016]{ByteNet:2016}
Kalchbrenner, N., Espeholt, L., Simonyan, K., Oord, A. v.~d., Graves, A., and
  Kavukcuoglu, K. (2016).
\newblock Neural machine translation in linear time.
\newblock {\em arXiv preprint arXiv:1610.10099}.

\bibitem[Kandasamy et~al., 2017]{Kandasamy:2017}
Kandasamy, K., Bachrach, Y., Tomioka, R., Tarlow, D., and Carter, D. (2017).
\newblock Batch policy gradient methods for improving neural conversation
  models.
\newblock {\em arXiv preprint arXiv:1702.03334}.

\bibitem[Konstas et~al., 2017]{Konstas:2017}
Konstas, I., Iyer, S., Yatskar, M., Choi, Y., and Zettlemoyer, L. (2017).
\newblock Neural amr: Sequence-to-sequence models for parsing and generation.
\newblock {\em arXiv preprint arXiv:1704.08381}.

\bibitem[Kottur et~al., 2017]{Kottur:2017}
Kottur, S., Moura, J.~M., Lee, S., and Batra, D. (2017).
\newblock Natural language does not emerge'naturally'in multi-agent dialog.
\newblock {\em arXiv preprint arXiv:1706.08502}.

\bibitem[Krizhevsky et~al., 2012]{Imagenet:2012}
Krizhevsky, A., Sutskever, I., and Hinton, G.~E. (2012).
\newblock Imagenet classification with deep convolutional neural networks.
\newblock In {\em Advances in neural information processing systems}, pages
  1097--1105.

\bibitem[Lample et~al., 2016]{Lample:2016}
Lample, G., Ballesteros, M., Subramanian, S., Kawakami, K., and Dyer, C.
  (2016).
\newblock Neural architectures for named entity recognition.
\newblock {\em arXiv preprint arXiv:1603.01360}.

\bibitem[Li et~al., 2015]{Li:2015}
Li, J., Galley, M., Brockett, C., Gao, J., and Dolan, B. (2015).
\newblock A diversity-promoting objective function for neural conversation
  models.
\newblock {\em arXiv preprint arXiv:1510.03055}.

\bibitem[Li et~al., 2016a]{Li:2016}
Li, J., Galley, M., Brockett, C., Spithourakis, G.~P., Gao, J., and Dolan, B.
  (2016a).
\newblock A persona-based neural conversation model.
\newblock {\em arXiv preprint arXiv:1603.06155}.

\bibitem[Li et~al., 2016b]{Li_HIL:2016}
Li, J., Miller, A.~H., Chopra, S., Ranzato, M., and Weston, J. (2016b).
\newblock Dialogue learning with human-in-the-loop.
\newblock {\em arXiv preprint arXiv:1611.09823}.

\bibitem[Li et~al., 2016c]{Li_RL:2016}
Li, J., Monroe, W., Ritter, A., Galley, M., Gao, J., and Jurafsky, D. (2016c).
\newblock Deep reinforcement learning for dialogue generation.
\newblock {\em arXiv preprint arXiv:1606.01541}.

\bibitem[Li et~al., 2017]{Li_adversarial:2017}
Li, J., Monroe, W., Shi, T., Ritter, A., and Jurafsky, D. (2017).
\newblock Adversarial learning for neural dialogue generation.
\newblock {\em arXiv preprint arXiv:1701.06547}.

\bibitem[Li et~al., 2016d]{stalemate:2016}
Li, X., Mou, L., Yan, R., and Zhang, M. (2016d).
\newblock Stalematebreaker: A proactive content-introducing approach to
  automatic human-computer conversation.
\newblock {\em arXiv preprint arXiv:1604.04358}.

\bibitem[Lin et~al., 2017]{Lin:2017}
Lin, Z., Feng, M., Santos, C. N.~d., Yu, M., Xiang, B., Zhou, B., and Bengio,
  Y. (2017).
\newblock A structured self-attentive sentence embedding.
\newblock {\em arXiv preprint arXiv:1703.03130}.

\bibitem[Lison and Bibauw, 2017]{Lison:2017}
Lison, P. and Bibauw, S. (2017).
\newblock Not all dialogues are created equal: Instance weighting for neural
  conversational models.
\newblock {\em arXiv preprint arXiv:1704.08966}.

\bibitem[Lison and Tiedemann, 2016]{OpenSubtitles:2016}
Lison, P. and Tiedemann, J. (2016).
\newblock Opensubtitles2016: Extracting large parallel corpora from movie and
  tv subtitles.
\newblock In {\em LREC}.

\bibitem[Liu et~al., 2016]{Liu:2016}
Liu, C.-W., Lowe, R., Serban, I.~V., Noseworthy, M., Charlin, L., and Pineau,
  J. (2016).
\newblock How not to evaluate your dialogue system: An empirical study of
  unsupervised evaluation metrics for dialogue response generation.
\newblock {\em arXiv preprint arXiv:1603.08023}.

\bibitem[Lowe et~al., 2017]{Lowe:2017}
Lowe, R., Noseworthy, M., Serban, I.~V., Angelard-Gontier, N., Bengio, Y., and
  Pineau, J. (2017).
\newblock Towards an automatic turing test: Learning to evaluate dialogue
  responses.
\newblock {\em arXiv preprint arXiv:1708.07149}.

\bibitem[Lowe et~al., 2015]{Lowe:2015}
Lowe, R., Pow, N., Serban, I., and Pineau, J. (2015).
\newblock The ubuntu dialogue corpus: A large dataset for research in
  unstructured multi-turn dialogue systems.
\newblock {\em arXiv preprint arXiv:1506.08909}.

\bibitem[Luong et~al., 2015]{Luong:2015}
Luong, M.-T., Pham, H., and Manning, C.~D. (2015).
\newblock Effective approaches to attention-based neural machine translation.
\newblock {\em arXiv preprint arXiv:1508.04025}.

\bibitem[Luong et~al., 2014]{Luong:2014}
Luong, M.-T., Sutskever, I., Le, Q.~V., Vinyals, O., and Zaremba, W. (2014).
\newblock Addressing the rare word problem in neural machine translation.
\newblock {\em arXiv preprint arXiv:1410.8206}.

\bibitem[Manning et~al., 1999]{Manning:1999}
Manning, C.~D., Sch{\"u}tze, H., et~al. (1999).
\newblock {\em Foundations of statistical natural language processing}, volume
  999.
\newblock MIT Press.

\bibitem[Marietto et~al., 2013]{Marietto:2013}
Marietto, M. d. G.~B., de~Aguiar, R.~V., Barbosa, G. d.~O., Botelho, W.~T.,
  Pimentel, E., Fran{\c{c}}a, R. d.~S., and da~Silva, V.~L. (2013).
\newblock Artificial intelligence markup language: A brief tutorial.
\newblock {\em arXiv preprint arXiv:1307.3091}.

\bibitem[Microsoft, 2017a]{Cortana:2017}
Microsoft (2017a).
\newblock Cortana.
\newblock \url{https://www.microsoft.com/en-us/windows/cortana}.
\newblock Accessed: 2017-10-04.

\bibitem[Microsoft, 2017b]{Microsoft:2017}
Microsoft (2017b).
\newblock Microsoft bot framework.
\newblock \url{https://dev.botframework.com/}.
\newblock Accessed: 2017-10-04.

\bibitem[Mikolov et~al., 2013a]{Mikolov:2013}
Mikolov, T., Chen, K., Corrado, G., and Dean, J. (2013a).
\newblock Efficient estimation of word representations in vector space.
\newblock {\em arXiv preprint arXiv:1301.3781}.

\bibitem[Mikolov et~al., 2013b]{Mikolov_skipgram:2013}
Mikolov, T., Sutskever, I., Chen, K., Corrado, G.~S., and Dean, J. (2013b).
\newblock Distributed representations of words and phrases and their
  compositionality.
\newblock In {\em Advances in neural information processing systems}, pages
  3111--3119.

\bibitem[Miller et~al., 2016]{Miller:2016}
Miller, A., Fisch, A., Dodge, J., Karimi, A.-H., Bordes, A., and Weston, J.
  (2016).
\newblock Key-value memory networks for directly reading documents.
\newblock {\em arXiv preprint arXiv:1606.03126}.

\bibitem[Miller et~al., 2017]{Miller:2017}
Miller, A.~H., Feng, W., Fisch, A., Lu, J., Batra, D., Bordes, A., Parikh, D.,
  and Weston, J. (2017).
\newblock Parlai: A dialog research software platform.
\newblock {\em arXiv preprint arXiv:1705.06476}.

\bibitem[Mitchell, 1998]{GA:1998}
Mitchell, M. (1998).
\newblock {\em An introduction to genetic algorithms}.
\newblock MIT press.

\bibitem[Mnih et~al., 2013]{Mnih:2013}
Mnih, V., Kavukcuoglu, K., Silver, D., Graves, A., Antonoglou, I., Wierstra,
  D., and Riedmiller, M. (2013).
\newblock Playing atari with deep reinforcement learning.
\newblock {\em arXiv preprint arXiv:1312.5602}.

\bibitem[Mnih et~al., 2015]{Mnih:2015}
Mnih, V., Kavukcuoglu, K., Silver, D., Rusu, A.~A., Veness, J., Bellemare,
  M.~G., Graves, A., Riedmiller, M., Fidjeland, A.~K., Ostrovski, G., et~al.
  (2015).
\newblock Human-level control through deep reinforcement learning.
\newblock {\em Nature}, 518(7540):529--533.

\bibitem[Nallapati et~al., 2016]{Nallapati:2016}
Nallapati, R., Zhou, B., Gulcehre, C., Xiang, B., et~al. (2016).
\newblock Abstractive text summarization using sequence-to-sequence rnns and
  beyond.
\newblock {\em arXiv preprint arXiv:1602.06023}.

\bibitem[Olah, 2015]{LSTM_article}
Olah, C. (2015).
\newblock Understanding lstm networks.
\newblock \url{http://colah.github.io/posts/2015-08-Understanding-LSTMs/}.
\newblock Accessed: 2017-10-08.

\bibitem[opensubtitles.org, 2017]{opensubtitles}
opensubtitles.org (2017).
\newblock Opensubtitles.
\newblock \url{https://www.opensubtitles.org/}.
\newblock Accessed: 2017-10-08.

\bibitem[Papineni et~al., 2002]{Papineni:2002}
Papineni, K., Roukos, S., Ward, T., and Zhu, W.-J. (2002).
\newblock Bleu: a method for automatic evaluation of machine translation.
\newblock In {\em Proceedings of the 40th annual meeting on association for
  computational linguistics}, pages 311--318. Association for Computational
  Linguistics.

\bibitem[Parikh et~al., 2016]{Parikh:2016}
Parikh, A.~P., T{\"a}ckstr{\"o}m, O., Das, D., and Uszkoreit, J. (2016).
\newblock A decomposable attention model for natural language inference.
\newblock {\em arXiv preprint arXiv:1606.01933}.

\bibitem[Park et~al., 2008]{Park:2008}
Park, Y., Patwardhan, S., Visweswariah, K., and Gates, S.~C. (2008).
\newblock An empirical analysis of word error rate and keyword error rate.
\newblock In {\em INTERSPEECH}, pages 2070--2073.

\bibitem[Paulus et~al., 2017]{Paulus:2017}
Paulus, R., Xiong, C., and Socher, R. (2017).
\newblock A deep reinforced model for abstractive summarization.
\newblock {\em arXiv preprint arXiv:1705.04304}.

\bibitem[Ramachandran et~al., 2016]{Ramachandran:2016}
Ramachandran, P., Liu, P.~J., and Le, Q.~V. (2016).
\newblock Unsupervised pretraining for sequence to sequence learning.
\newblock {\em arXiv preprint arXiv:1611.02683}.

\bibitem[Rumelhart et~al., 1988]{RNN:1988}
Rumelhart, D.~E., Hinton, G.~E., Williams, R.~J., et~al. (1988).
\newblock Learning representations by back-propagating errors.
\newblock {\em Cognitive modeling}, 5(3):1.

\bibitem[Salton and Buckley, 1988]{Salton:1988}
Salton, G. and Buckley, C. (1988).
\newblock Term-weighting approaches in automatic text retrieval.
\newblock {\em Information processing \& management}, 24(5):513--523.

\bibitem[Schuster and Paliwal, 1997]{Schuster:1997}
Schuster, M. and Paliwal, K.~K. (1997).
\newblock Bidirectional recurrent neural networks.
\newblock {\em IEEE Transactions on Signal Processing}, 45(11):2673--2681.

\bibitem[Sennrich et~al., 2015]{Sennrich:2015}
Sennrich, R., Haddow, B., and Birch, A. (2015).
\newblock Neural machine translation of rare words with subword units.
\newblock {\em arXiv preprint arXiv:1508.07909}.

\bibitem[Serban et~al., 2017a]{Serban_MrRNN:2017}
Serban, I.~V., Klinger, T., Tesauro, G., Talamadupula, K., Zhou, B., Bengio,
  Y., and Courville, A.~C. (2017a).
\newblock Multiresolution recurrent neural networks: An application to dialogue
  response generation.
\newblock In {\em AAAI}, pages 3288--3294.

\bibitem[Serban et~al., 2015]{Serban_survey:2015}
Serban, I.~V., Lowe, R., Charlin, L., and Pineau, J. (2015).
\newblock A survey of available corpora for building data-driven dialogue
  systems.
\newblock {\em arXiv preprint arXiv:1512.05742}.

\bibitem[Serban et~al., 2017b]{Serban:2017}
Serban, I.~V., Sankar, C., Germain, M., Zhang, S., Lin, Z., Subramanian, S.,
  Kim, T., Pieper, M., Chandar, S., Ke, N.~R., et~al. (2017b).
\newblock A deep reinforcement learning chatbot.
\newblock {\em arXiv preprint arXiv:1709.02349}.

\bibitem[Serban et~al., 2016]{Serban:2015}
Serban, I.~V., Sordoni, A., Bengio, Y., Courville, A.~C., and Pineau, J.
  (2016).
\newblock Building end-to-end dialogue systems using generative hierarchical
  neural network models.
\newblock In {\em AAAI}, pages 3776--3784.

\bibitem[Serban et~al., 2017c]{Serban_VHRED:2017}
Serban, I.~V., Sordoni, A., Lowe, R., Charlin, L., Pineau, J., Courville,
  A.~C., and Bengio, Y. (2017c).
\newblock A hierarchical latent variable encoder-decoder model for generating
  dialogues.
\newblock In {\em AAAI}, pages 3295--3301.

\bibitem[Shang et~al., 2015]{Shang:2015}
Shang, L., Lu, Z., and Li, H. (2015).
\newblock Neural responding machine for short-text conversation.
\newblock {\em arXiv preprint arXiv:1503.02364}.

\bibitem[Shannon, 2017]{Shannon:2017}
Shannon, M. (2017).
\newblock Optimizing expected word error rate via sampling for speech
  recognition.
\newblock {\em arXiv preprint arXiv:1706.02776}.

\bibitem[Shao et~al., 2017]{Shao:2017}
Shao, Y., Gouws, S., Britz, D., Goldie, A., Strope, B., and Kurzweil, R.
  (2017).
\newblock Generating high-quality and informative conversation responses with
  sequence-to-sequence models.
\newblock In {\em Proceedings of the 2017 Conference on Empirical Methods in
  Natural Language Processing}, pages 2200--2209.

\bibitem[Shazeer et~al., 2017]{MoE:2017}
Shazeer, N., Mirhoseini, A., Maziarz, K., Davis, A., Le, Q., Hinton, G., and
  Dean, J. (2017).
\newblock Outrageously large neural networks: The sparsely-gated
  mixture-of-experts layer.
\newblock {\em arXiv preprint arXiv:1701.06538}.

\bibitem[Shen et~al., 2017]{Shen:2017}
Shen, X., Su, H., Li, Y., Li, W., Niu, S., Zhao, Y., Aizawa, A., and Long, G.
  (2017).
\newblock A conditional variational framework for dialog generation.
\newblock {\em arXiv preprint arXiv:1705.00316}.

\bibitem[Silver et~al., 2016]{Silver:2016}
Silver, D., Huang, A., Maddison, C.~J., Guez, A., Sifre, L., Van Den~Driessche,
  G., Schrittwieser, J., Antonoglou, I., Panneershelvam, V., Lanctot, M.,
  et~al. (2016).
\newblock Mastering the game of go with deep neural networks and tree search.
\newblock {\em Nature}, 529(7587):484--489.

\bibitem[Silver et~al., 2017]{alphagozero}
Silver, D., Schrittwieser, J., Simonyan, K., Antonoglou, I., Huang, A., Guez,
  A., Hubert, T., Baker, L., Lai, M., Bolton, A., Chen, Y., Lillicrap, T., Hui,
  F., Sifre, L., van~den Driessche, G., Graepel, T., and Hassabis, D. (2017).
\newblock Mastering the game of go without human knowledge.
\newblock {\em Nature}, 550(7676):354--359.

\bibitem[Song et~al., 2016]{Song:2016}
Song, Y., Yan, R., Li, X., Zhao, D., and Zhang, M. (2016).
\newblock Two are better than one: An ensemble of retrieval-and
  generation-based dialog systems.
\newblock {\em arXiv preprint arXiv:1610.07149}.

\bibitem[Sordoni et~al., 2015]{Sordoni:2015}
Sordoni, A., Galley, M., Auli, M., Brockett, C., Ji, Y., Mitchell, M., Nie,
  J.-Y., Gao, J., and Dolan, B. (2015).
\newblock A neural network approach to context-sensitive generation of
  conversational responses.
\newblock {\em arXiv preprint arXiv:1506.06714}.

\bibitem[Sriram et~al., 2017]{Sriram:2017}
Sriram, A., Jun, H., Satheesh, S., and Coates, A. (2017).
\newblock Cold fusion: Training seq2seq models together with language models.
\newblock {\em arXiv preprint arXiv:1708.06426}.

\bibitem[Srivastava et~al., 2014]{Dropout:2014}
Srivastava, N., Hinton, G.~E., Krizhevsky, A., Sutskever, I., and
  Salakhutdinov, R. (2014).
\newblock Dropout: a simple way to prevent neural networks from overfitting.
\newblock {\em Journal of machine learning research}, 15(1):1929--1958.

\bibitem[Sukhbaatar et~al., 2015]{Sukhbaatar:2015}
Sukhbaatar, S., Weston, J., Fergus, R., et~al. (2015).
\newblock End-to-end memory networks.
\newblock In {\em Advances in neural information processing systems}, pages
  2440--2448.

\bibitem[Sutskever et~al., 2014]{Sutskever:2014}
Sutskever, I., Vinyals, O., and Le, Q.~V. (2014).
\newblock Sequence to sequence learning with neural networks.
\newblock In {\em Advances in neural information processing systems}, pages
  3104--3112.

\bibitem[Sutton and Barto, 1998]{Sutton:1998}
Sutton, R.~S. and Barto, A.~G. (1998).
\newblock {\em Reinforcement learning: An introduction}, volume~1.
\newblock MIT press Cambridge.

\bibitem[Tao et~al., 2017]{Tao:2017}
Tao, C., Mou, L., Zhao, D., and Yan, R. (2017).
\newblock Ruber: An unsupervised method for automatic evaluation of open-domain
  dialog systems.
\newblock {\em arXiv preprint arXiv:1701.03079}.

\bibitem[Tensorflow, 2017]{deep_seq2seq}
Tensorflow (2017).
\newblock Sequence-to-sequence models.
\newblock \url{https://www.tensorflow.org/tutorials/seq2seq}.
\newblock Accessed: 2017-10-08.

\bibitem[Tiedemann, 2009]{Tiedemann:2009}
Tiedemann, J. (2009).
\newblock News from opus-a collection of multilingual parallel corpora with
  tools and interfaces.
\newblock In {\em Recent advances in natural language processing}, volume~5,
  pages 237--248.

\bibitem[Vaswani et~al., 2017]{Vaswani:2017}
Vaswani, A., Shazeer, N., Parmar, N., Uszkoreit, J., Jones, L., Gomez, A.~N.,
  Kaiser, L., and Polosukhin, I. (2017).
\newblock Attention is all you need.
\newblock {\em arXiv preprint arXiv:1706.03762}.

\bibitem[Vinyals and Le, 2015]{Vinyals:2015}
Vinyals, O. and Le, Q. (2015).
\newblock A neural conversational model.
\newblock {\em arXiv preprint arXiv:1506.05869}.

\bibitem[Wallace, 2009]{Wallace:2009}
Wallace, R.~S. (2009).
\newblock The anatomy of alice.
\newblock {\em Parsing the Turing Test}, pages 181--210.

\bibitem[Weizenbaum, 1966]{Weizenbaum:1966}
Weizenbaum, J. (1966).
\newblock Eliza—a computer program for the study of natural language
  communication between man and machine.
\newblock {\em Communications of the ACM}, 9(1):36--45.

\bibitem[Wen et~al., 2016]{Wen:2016}
Wen, T.-H., Vandyke, D., Mrksic, N., Gasic, M., Rojas-Barahona, L.~M., Su,
  P.-H., Ultes, S., and Young, S. (2016).
\newblock A network-based end-to-end trainable task-oriented dialogue system.
\newblock {\em arXiv preprint arXiv:1604.04562}.

\bibitem[Werbos, 1990]{Werbos:1990}
Werbos, P.~J. (1990).
\newblock Backpropagation through time: what it does and how to do it.
\newblock {\em Proceedings of the IEEE}, 78(10):1550--1560.

\bibitem[Weston et~al., 2015]{Weston:2015}
Weston, J., Bordes, A., Chopra, S., Rush, A.~M., van Merri{\"e}nboer, B.,
  Joulin, A., and Mikolov, T. (2015).
\newblock Towards ai-complete question answering: A set of prerequisite toy
  tasks.
\newblock {\em arXiv preprint arXiv:1502.05698}.

\bibitem[Williams et~al., 2017]{Williams:2017}
Williams, J.~D., Asadi, K., and Zweig, G. (2017).
\newblock Hybrid code networks: practical and efficient end-to-end dialog
  control with supervised and reinforcement learning.
\newblock {\em arXiv preprint arXiv:1702.03274}.

\bibitem[Williams, 1992]{Williams:1992}
Williams, R.~J. (1992).
\newblock Simple statistical gradient-following algorithms for connectionist
  reinforcement learning.
\newblock {\em Machine learning}, 8(3-4):229--256.

\bibitem[Wiseman and Rush, 2016]{Wiseman:2016}
Wiseman, S. and Rush, A.~M. (2016).
\newblock Sequence-to-sequence learning as beam-search optimization.
\newblock {\em arXiv preprint arXiv:1606.02960}.

\bibitem[Worswick, 2017]{Mitsuku:2017}
Worswick, S. (2017).
\newblock Mitsuku.
\newblock \url{http://www.mitsuku.com/}.
\newblock Accessed: 2017-10-04.

\bibitem[Wu et~al., 2016]{googleNMT:2016}
Wu, Y., Schuster, M., Chen, Z., Le, Q.~V., Norouzi, M., Macherey, W., Krikun,
  M., Cao, Y., Gao, Q., Macherey, K., et~al. (2016).
\newblock Google's neural machine translation system: Bridging the gap between
  human and machine translation.
\newblock {\em arXiv preprint arXiv:1609.08144}.

\bibitem[Xing et~al., 2017a]{Xing_topic:2017}
Xing, C., Wu, W., Wu, Y., Liu, J., Huang, Y., Zhou, M., and Ma, W.-Y. (2017a).
\newblock Topic aware neural response generation.
\newblock In {\em AAAI}, pages 3351--3357.

\bibitem[Xing et~al., 2017b]{Xing:2017}
Xing, C., Wu, W., Wu, Y., Zhou, M., Huang, Y., and Ma, W.-Y. (2017b).
\newblock Hierarchical recurrent attention network for response generation.
\newblock {\em arXiv preprint arXiv:1701.07149}.

\bibitem[Yao et~al., 2016]{Yao:2016}
Yao, K., Peng, B., Zweig, G., and Wong, K.-F. (2016).
\newblock An attentional neural conversation model with improved specificity.
\newblock {\em arXiv preprint arXiv:1606.01292}.

\bibitem[Yao et~al., 2015]{Yao:2015}
Yao, K., Zweig, G., and Peng, B. (2015).
\newblock Attention with intention for a neural network conversation model.
\newblock {\em arXiv preprint arXiv:1510.08565}.

\bibitem[Yin et~al., 2017]{Yin:2017}
Yin, Z., Chang, K.-h., and Zhang, R. (2017).
\newblock Deepprobe: Information directed sequence understanding and chatbot
  design via recurrent neural networks.
\newblock In {\em Proceedings of the 23rd ACM SIGKDD International Conference
  on Knowledge Discovery and Data Mining}, pages 2131--2139. ACM.

\bibitem[Yu et~al., 2017]{Yu:2017}
Yu, Z., Black, A.~W., and Rudnicky, A.~I. (2017).
\newblock Learning conversational systems that interleave task and non-task
  content.
\newblock {\em arXiv preprint arXiv:1703.00099}.

\bibitem[Zhao et~al., 2017a]{Zhao:2017}
Zhao, T., Lu, A., Lee, K., and Eskenazi, M. (2017a).
\newblock Generative encoder-decoder models for task-oriented spoken dialog
  systems with chatting capability.
\newblock {\em arXiv preprint arXiv:1706.08476}.

\bibitem[Zhao et~al., 2017b]{Zhaob:2017}
Zhao, T., Zhao, R., and Eskenazi, M. (2017b).
\newblock Learning discourse-level diversity for neural dialog models using
  conditional variational autoencoders.
\newblock {\em arXiv preprint arXiv:1703.10960}.

\bibitem[Zhou et~al., 2017]{Zhou:2017}
Zhou, H., Huang, M., Zhang, T., Zhu, X., and Liu, B. (2017).
\newblock Emotional chatting machine: Emotional conversation generation with
  internal and external memory.
\newblock {\em arXiv preprint arXiv:1704.01074}.

\bibitem[Zhu et~al., 2017]{Zhu:2017}
Zhu, Q., Li, Y., and Li, X. (2017).
\newblock Character sequence-to-sequence model with global attention for
  universal morphological reinflection.
\newblock {\em Proceedings of the CoNLL SIGMORPHON 2017 Shared Task: Universal
  Morphological Reinflection}, pages 85--89.

\bibitem[Zoph and Le, 2016]{Zoph:2016}
Zoph, B. and Le, Q.~V. (2016).
\newblock Neural architecture search with reinforcement learning.
\newblock {\em arXiv preprint arXiv:1611.01578}.

\end{thebibliography}
\end{document}